\documentclass[final]{cvpr}

\usepackage{times}
\usepackage{epsfig}
\usepackage{graphicx}
\usepackage{amsmath}
\usepackage{amssymb}
\usepackage{bbm}
\usepackage{float}
\usepackage{blindtext}
\usepackage{caption}
\usepackage{subcaption}
\usepackage{bm}
\usepackage{booktabs}
\usepackage{makecell}
\usepackage{multirow}

\usepackage[pagebackref=true,breaklinks=true,colorlinks,bookmarks=false]{hyperref}

\def\cvprPaperID{6885} %
\def\confYear{CVPR 2021}

\title{High-Fidelity and Arbitrary Face Editing}
\author{Yue Gao$^{1}$, Fangyun Wei$^{2}$, Jianmin Bao$^{2}$, Shuyang Gu$^{3}$, Dong Chen$^{2}$, Fang Wen$^{2}$, Zhouhui Lian$^{1}$\thanks{Zhouhui Lian is the corresponding author. This work was supported by Beijing Nova Program of Science and Technology (Grant No.: Z191100001119077).} \vspace{6pt}\\
    $^1$Wangxuan Institute of Computer Technology, Peking University, China \\
    $^2$Microsoft Research Asia \\
    $^3$University of Science and Technology of China \\
    {\tt\small \{gerry, lianzhouhui\}@pku.edu.cn,  \{fawe, jianbao, doch, fangwen\}@microsoft.com}, \\
    {\tt\small gsy777@mail.ustc.edu.cn} \\
}

\begin{document}
\maketitle

\begin{abstract}

Cycle consistency is widely used for face editing. However, we observe that the generator tends to find a tricky way to hide information from the original image to satisfy the constraint of cycle consistency, making it impossible to maintain the rich details (\eg, wrinkles and moles) of non-editing areas.
In this work, we propose a simple yet effective method named HifaFace to address the above-mentioned problem from two perspectives.
First, we relieve the pressure of the generator to synthesize rich details by directly feeding the high-frequency information of the input image into the end of the generator. Second, we adopt an additional discriminator to encourage the generator to synthesize rich details.
Specifically, we apply wavelet transformation to transform the image into multi-frequency domains, among which the high-frequency parts can be used to recover the rich details.
We also notice that a fine-grained and wider-range control for the attribute is of great importance for face editing. To achieve this goal, we propose a novel attribute regression loss.
Powered by the proposed framework, we achieve high-fidelity and arbitrary face editing, outperforming other state-of-the-art approaches.
\end{abstract}

\vspace{-4mm}
\section{Introduction}

Face editing is a process of editing the specific attributes or regions of an input facial image while keeping the non-editing attributes/areas unchanged. With the rapid development of Generative Adversarial Networks (GANs)~\cite{Goodfellow2014GenerativeAN}, many recent works on face editing~\cite{Isola2017ImagetoImageTW, Choi2018StarGANUG,He2019AttGANFA, Liu2019STGANAU, Wu2019RelGANMI, chusscgan} leverage the advanced conditional GANs and achieve remarkable progress. Due to the lack of paired images during training, they typically use \emph{cycle consistency} to keep the non-editing attributes/areas unchanged. Namely, given an image $\bm{x}$, it requires $\bm{x} = G(G( \bm{x}, \bm{\Delta}), -\bm{\Delta})$, where $G$ represents the generator and $\bm{\Delta}$ indicates the attribute that needs to be changed. However, we find that even if the cycle consistency is satisfied, images generated by $G$ may still be blurry and lose rich details from input images.

\begin{figure}
    \captionsetup[subfigure]{aboveskip=1pt} %
    \captionsetup[subfigure]{font={small}}
    \centering
    \begin{subfigure}[t]{0.247\linewidth}
        \includegraphics[width=\linewidth]{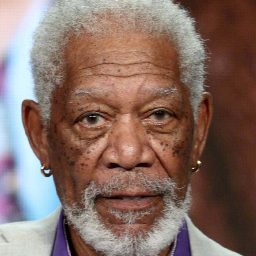}
        \caption{Input $\bm{x}$}
    \end{subfigure}\hfill
    \begin{subfigure}[t]{0.247\linewidth}
        \includegraphics[width=\linewidth]{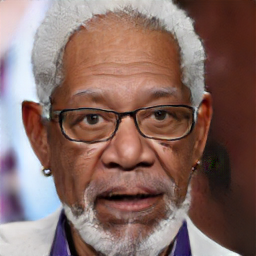}
        \caption{$\bm{\hat{y}} = \bm{x}\text{ + EG}$}
    \end{subfigure}\hfill
    \begin{subfigure}[t]{0.247\linewidth}
        \includegraphics[width=\linewidth]{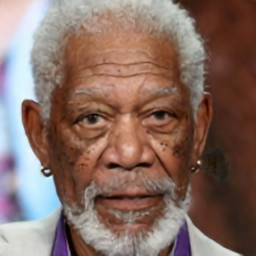}
        \caption{$\bm{\hat{x}} = \bm{\hat y}\text{ - EG}$}
    \end{subfigure}\hfill
    \begin{subfigure}[t]{0.247\linewidth}
        \includegraphics[width=\linewidth]{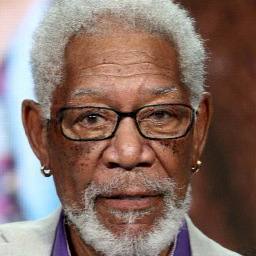}
        \caption{\textbf{HifaFace}}
    \end{subfigure}\hfill
    \vspace{-2mm}
    \caption{The results of a representative face editing method StarGAN~\cite{Choi2018StarGANUG}: (a) the input image; (b) the output image synthesized by editing (a) with the attribute eyeglasses (EG); (c) the reconstructed image with (b) as input. We observe that the rich details are all missing in $\bm{\hat y}$, but are almost restored in $\bm{\hat x}$. (d) The high-fidelity face image synthesized by our HifaFace.}  %
    \label{fig:steganography_recon}
    \vspace{-4.5mm}
\end{figure}

To demonstrate the above-mentioned problem, we take StarGAN~\cite{Choi2018StarGANUG} as an example. As shown in Figure~\ref{fig:steganography_recon}, we feed an input face image $\bm{x}$ to StarGAN and expect it to add eyeglasses on the face. Although the output $\bm{\hat{y}}$ does wear eyeglasses, the details (\eg, wrinkles and moles) are all missing. However, we observe an intriguing phenomenon as follows. When we feed $\bm{\hat{y}}$ into the StarGAN model and expect removing eyeglasses on the face, the reconstruction result $\bm{\hat x}$ surprisingly recovers almost all rich details, which satisfies the purpose of setting cycle consistency. This observation indicates that the generator encodes the rich details of the input image into the output image in the form of ``hidden'' signals, and then decodes the feature with these ``hidden'' signals to achieve reconstruction. The above-mentioned phenomenon is called steganography~\cite{Chu2017CycleGANAM} and is undesirable for face editing ~\cite{Snchez2020ARC}.

To prevent the generator from taking this trick route to satisfy cycle consistency, we propose a simple yet effective face editing method called HifaFace. We tackle this problem from two perspectives. First, we directly feed the high-frequency information of the input image to the end of the generator to alleviate the generator's struggles for synthesizing rich details so that it gives up encoding the hidden signals. Second, we adopt an additional discriminator to constrain the generator to synthesize rich details, thus further preventing the generator from finding a trivial solution for cycle consistency.

Specifically, we adopt wavelet transformation to transform an image into multiple frequency domains. We find that almost all rich details lie in the high-frequency domains. In order to feed the high-frequency information to the generator, we adopt an encoder-decoder-like structure and design a novel module named Wavelet-based Skip-Connection to replace the original Skip-Connection.

To achieve the goal of providing a fine-grained and wider-range control for each facial attribute, we also propose a novel loss, called the attribute regression loss, which requires the generated image to explicitly describe the change on selected attributes and thus enables wider-range and controllable face editing.
Furthermore, our method is able to effectively exploit large amounts of unlabeled face images for training, which can further improve the fidelity of synthesized faces in the wild. Powered by the proposed framework, we obtain high-fidelity and arbitrarily controllable face editing results.

In summary, our major contributions are threefold:

\begin{itemize}
\item We propose a novel wavelet-based face editing method, called \textbf{HifaFace}, for high-fidelity and arbitrary face editing.
\item We revisit cycle consistency in face editing and observe that the generator learns to apply a tricky way to satisfy the constraint of cycle consistency by hiding signals in the output image. We thoroughly analyze this phenomenon and provide an effective solution to handle the problem.
\item Both qualitative and quantitative results demonstrate the effectiveness of the proposed framework for improving the quality of edited face images.
\end{itemize}

\section{Related Work}
\noindent\textbf{Generative Adversarial Networks.} Generative Adversarial Networks (GANs)~\cite{Goodfellow2014GenerativeAN} have been widely used in the literature. A typical GAN-based model adopts a generator and a discriminator to implement adversarial training during the learning process. The capability of GANs enables a wide range of computer vision applications such as image generation~\cite{Karras2018ProgressiveGO, Karras2019ASG}, font generation~\cite{Gao2019ArtisticGI, Wang2020Attribute2Font}, low-level vision~\cite{ledig2017photo, yang2018low}, complex distribution modeling~\cite{brock2018large, nowozin2016f, zhang2019self}, and so on.

\noindent\textbf{Cycle Consistency.} Assessing the matches between two or more samples with cycle consistency is a commonly used technique in computer vision. It has been applied to many popular vision tasks such as image alignment~\cite{zhou2016learning, zhou2015multi}, depth estimation~\cite{zhou2017unsupervised, yin2018geonet}, correspondence learning~\cite{zhou2015flowweb, wang2019learning} and etc. How to exploit the cyclic relation to learn the bi-directional transformation functions between different domains attracts intensive attention in recent years, since it can be used in domain adaption~\cite{hoffman2018cycada} and unpaired image-to-image translation~\cite{Zhu2017UnpairedIT, Choi2018StarGANUG}. Also, some previous works~\cite{Chu2017CycleGANAM, porav2019reducing, Snchez2020ARC} notice the problem of cycle consistency in CycleGANs~\cite{Zhu2017UnpairedIT} that the cycle consistency tends to do steganography~\cite{Chu2017CycleGANAM} during the training process. In this paper, we concentrate on the process of steganography in the face editing framework and demonstrate its harmfulness for face editing results. Then, we propose a simple yet effective method to solve the problem of cycle consistency in the task of face editing.

\noindent\textbf{Face Generation and Editing.} The face is of great interest in computer vision and computer graphics circles. Thanks to the recent development of GANs, a number of methods have reported achieving the high-quality generation of face images~\cite{bao2018towards, Karras2018ProgressiveGO, Karras2019ASG} and the flexible manipulation of facial attributes for a given face~\cite{Choi2018StarGANUG, Wu2019RelGANMI, Liu2019STGANAU}. Generally speaking, these methods can be classified into two groups. The first group of methods~\cite{Shen2020InterFaceGANIT, abdal2019image2stylegan, Abdal2020StyleFlowAE} leverages pre-trained GANs to achieve the manipulation of faces. They first extract the latent code for a given face, then manipulate it to obtain the edited results. One major drawback of those methods is that they are not able to get the perfect latent code for a given image, resulting in the edited image that loses the rich details of the original face. The second group of methods~\cite{Gu2019MaskGuidedPE, Choi2018StarGANUG, Wu2019RelGANMI, Liu2019STGANAU, chusscgan} utilizes image-to-image translation techniques for face editing, where the original face image is also fed to the network. However, synthesis results of these methods are far from satisfactory and their qualities are lower compared to the first type of approaches. One possible reason is the existence of the above-mentioned problem of ``cycle consistency''. In this paper, we explicitly investigate the reasons why image-to-image translation methods cannot obtain satisfactory results
for the face editing task and propose an effective framework to solve this problem.

\section{Method Description}
\begin{figure*}
    \captionsetup[subfigure]{aboveskip=1pt} %
    \captionsetup[subfigure]{labelformat=empty}
    \captionsetup[subfigure]{font=footnotesize}
    \centering
    \begin{subfigure}[t]{0.136\linewidth}
        \includegraphics[width=\linewidth]
        {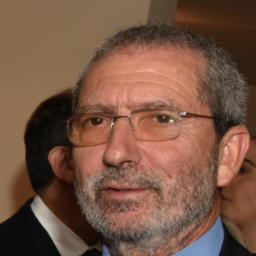} \\
        \vspace{-4mm}
        \includegraphics[width=\linewidth]
        {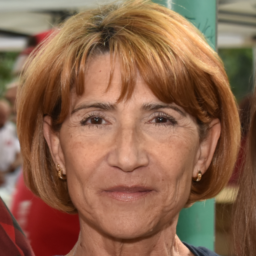}
        \caption{a) Input $\bm{x}$}
    \end{subfigure}\hspace{0.6mm}%
    \begin{subfigure}[t]{0.136\linewidth}
        \includegraphics[width=\linewidth]
        {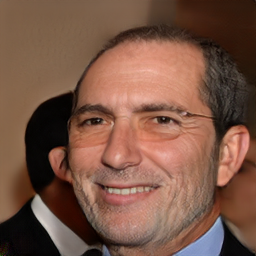} \\
        \vspace{-4mm}
        \includegraphics[width=\linewidth]
        {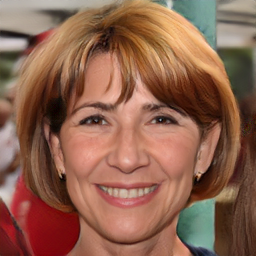}
        \caption{b) $\bm{\hat y} = G(\bm{x}, \bm{\Delta})$}
    \end{subfigure}\hspace{0.6mm}%
    \begin{subfigure}[t]{0.136\linewidth}
        \includegraphics[width=\linewidth]
        {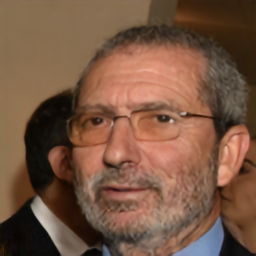} \\
        \vspace{-4mm}
        \includegraphics[width=\linewidth]
        {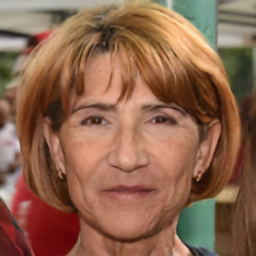}
        \caption{c) $ \bm{y'} = G(\bm{\hat y}, \bm{\Delta}')$}
    \end{subfigure}\hspace{0.6mm}%
    \begin{subfigure}[t]{0.136\linewidth}
        \includegraphics[width=\linewidth]
        {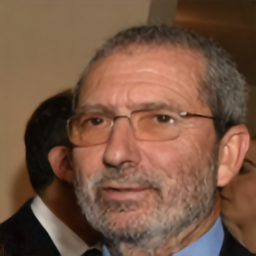} \\
        \vspace{-4mm}
        \includegraphics[width=\linewidth]
        {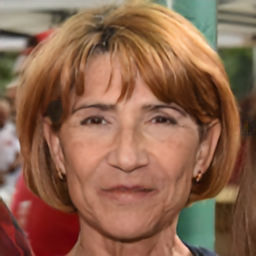}
        \caption{d) $\bm{\hat x} = G(\bm{\hat y}, -\bm{\Delta})$}
    \end{subfigure}\hspace{0.6mm}%
    \begin{subfigure}[t]{0.136\linewidth}
        \includegraphics[width=\linewidth]
        {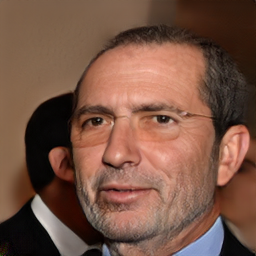} \\
        \vspace{-4mm}
        \includegraphics[width=\linewidth]
        {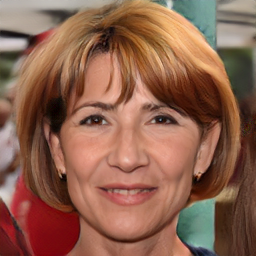}
        \caption{e) $ \bm{\overline{y}} = G(\bm{x}, \bm{0})$}
    \end{subfigure}\hspace{0.6mm}%
    \begin{subfigure}[t]{0.136\linewidth}
        \includegraphics[width=\linewidth]
        {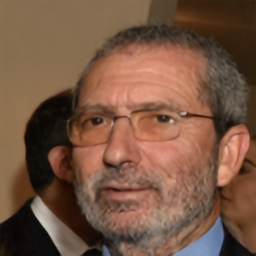} \\
        \vspace{-4mm}
        \includegraphics[width=\linewidth]
        {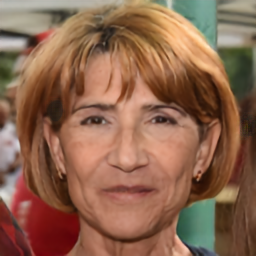}
        \caption{f) $\bm{\overline{x}} = G(\bm{\overline{y}}, \bm{0})$}
    \end{subfigure}\hspace{0.6mm}%
    \begin{subfigure}[t]{0.136\linewidth}
        \includegraphics[width=\linewidth]
        {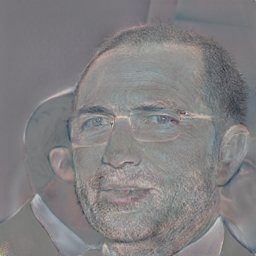} \\
        \vspace{-4mm}
        \includegraphics[width=\linewidth]
        {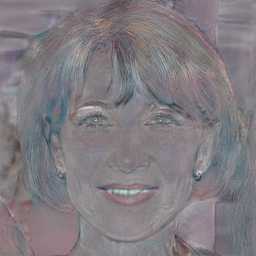}
        \caption{g) $\bm{h} = \bm{\overline{y}} - \bm{\overline{x}}$}
    \end{subfigure}
    \vspace{-3mm}
    \caption{Previous methods (\eg, StarGAN~\cite{Choi2018StarGANUG}) using cycle consistency tend to hide information in the output images.}
    \label{fig:analysis}
    \vspace{-3mm}
\end{figure*}

\subsection{Revisiting Cycle Consistency in Face Editing}  %
\label{sec:revisiting}
Suppose $G$ is a face editing model (\eg, StarGAN~\cite{Choi2018StarGANUG} or RelGAN~\cite{Wu2019RelGANMI}), which typically requires two inputs: a source face image and the difference between target attributes and source attributes (StarGAN directly uses the target attributes). Given an image $\bm{x}$ with the attributes $\bm{a}_x$, we aim to change its attributes to $\bm{a}_y$. Namely, the generator $G$ can take $\bm{x}$ and $\bm{\Delta} = \bm{a}_y - \bm{a}_x$ as input and generate the synthesis result $\bm{\hat y} = G(\bm{x}, \bm{\Delta})$. To guarantee that only the regions related to attribute changes in $\bm{\hat y}$ are different against $\bm{x}$, cycle consistency is usually applied between $G(\bm{\hat y}, -\bm{\Delta})$ and $\bm{x}$. However, we observe an intriguing phenomenon, that is, if we further input $\bm{\hat{y}}$ to the generator with any other attribute difference $\bm{\Delta'}$ ($\bm{\Delta'} \neq -\bm{\Delta}$) to get $\bm{y'} = G(\bm{\hat y}, \bm{\Delta'})$, the generated result $\bm{y}'$ will tend to be the same as input. This phenomenon is demonstrated in Figure \ref{fig:analysis}(a,b,c), indicating that the generator learns to apply a tricky way to achieve the cycle consistency by ``hiding'' information in the output image. If we feed an image with hidden information to the generator, it tends to ignore the input attributes and only leverage the hidden information to reconstruct the original image.

Thereby, it is crucial to figure out this hidden information. One possible way is to calculate the difference between the edited image and its ground truth~\cite{Chu2017CycleGANAM}. But, unfortunately, it is usually impossible to obtain the ground truth. To address this problem, we let the target attributes be identical to the original attributes by feeding the input image and $\bm{\Delta} = \bm{0}$ to the generator. Then we can get the synthesis result $\bm{\overline{y}} = G(\bm{x}, \bm{0})$, whose corresponding ground truth is the original input image. We further feed $\bm{\overline{y}}$ with $\bm{\Delta} = \bm{0}$ back to the generator to get the result $\bm{\overline{x}} = G(\bm{\overline{y}}, \bm{0})$. Examples of $\bm{\overline{y}}$ and its reconstructed image $\bm{\overline{x}}$ are shown in Figure~\ref{fig:analysis}(e, f). We find that there exist significant differences between $\bm{\overline{y}}$ and $\bm{x}$, especially the hair and wrinkles.  However, the reconstructed image  $\bm{\overline{x}}$ is almost perfect, which verifies the existence of hidden information in $\bm{\overline{y}}$. In this manner, we can get the hidden information $\bm{h} = \bm{\overline{y}} - \bm{\overline{x}}$ (see Figure~\ref{fig:analysis}(g)).%

Motivated by the analysis mentioned above, we can prevent the generator from hiding information by simply restricting the hidden information $\bm{h}$ to be $\bm{0}$, which is equivalent to restrict $G(\bm{x}, \bm{0})$ to be equal to $\bm{x}$. We notice that the above strategy has been adopted by several existing methods~\cite{Liu2019STGANAU, chusscgan}. But this restriction is harmful to the face editing model and makes the edited results remain unchanged during the editing process. Another possible way to prevent the generator from hiding information is to corrupt the hidden information during training. We have tried to add image transformations such as Gaussian blur, flip and rotation on the synthesis results to encourage the generator to give up hiding information. However, we find that the generator still struggles to synthesize rich details and learns to satisfy cycle consistency via a trivial solution.

This paper proposes addressing this problem with a novel framework. The key idea is to prevent the generator from encoding hidden information and encourage it to synthesize perceptible information. By inspecting the hidden information, we find that it is highly related to the high-frequency signals of the input image. Therefore, we choose to utilize the widely used wavelet transformation to decompose the image into domains with different frequencies and take the high-frequency parts to represent rich details.

More specifically, we propose solving the problem of cycle consistency from two perspectives. On the one hand, we directly feed the high-frequency signals to the end of the generator to mitigate the generator's struggles for synthesizing rich details so that it gives up to encode the hidden signals. On the other hand, we also employ an additional discriminator to encourage the generator to generate rich details, thus further preventing the generator from finding a trivial solution for cycle consistency.

\begin{figure*}
    \centering
    \includegraphics[width=0.9\textwidth]{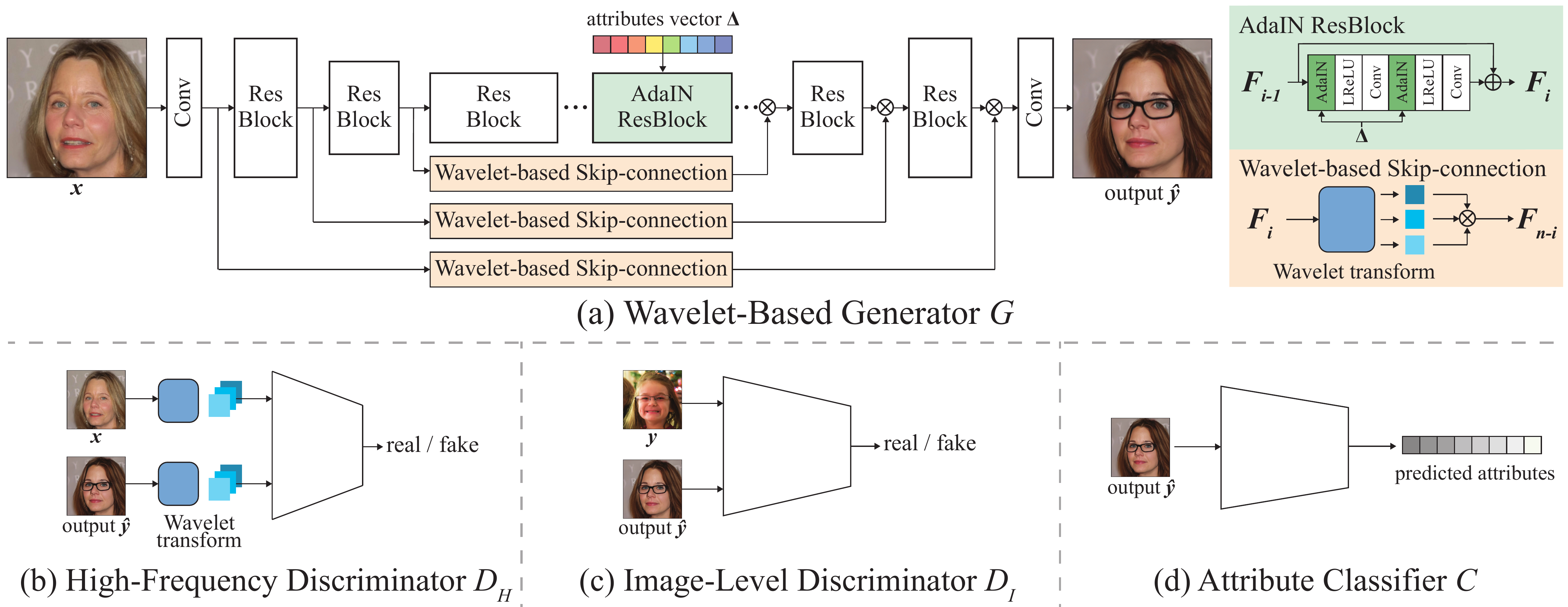}
    \vspace{-3mm}
    \caption{An overview of our proposed method, which contains four parts: a) the Wavelet-Based Generator $G$; b) the High-Frequency Discriminator $D_H$; c) the Image-Level Discriminator $D_I$ and; d) the Attribute Classifier $C$. $\bigoplus$ denotes element-wise plus and $\bigotimes$ denotes channel-wise concatenation.}
    \label{fig:framework}
    \vspace{-5mm}
\end{figure*}

\subsection{Wavelet Transformation}
\begin{figure}
    \captionsetup[subfigure]{aboveskip=1pt} %
    \captionsetup[subfigure]{labelformat=empty}
    \captionsetup[subfigure]{font={scriptsize}}
    \centering
    \centering
    \begin{subfigure}[t]{0.15\linewidth}
        \includegraphics[width=\linewidth]
        {images/motivation_179_not_in} \\
        \vspace{-4mm}
        \includegraphics[width=\linewidth]
        {images/motivation_5549_not_work_in}
        \caption{Input}
    \end{subfigure}
    \begin{subfigure}[t]{0.15\linewidth}
        \includegraphics[width=\linewidth]
        {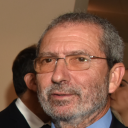} \\
        \vspace{-4mm}
        \includegraphics[width=\linewidth]
        {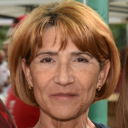}
        \caption{$\bm{\textbf{LL}}$}
    \end{subfigure}
    \begin{subfigure}[t]{0.15\linewidth}
        \includegraphics[width=\linewidth]
        {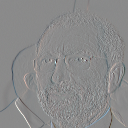} \\
        \vspace{-4mm}
        \includegraphics[width=\linewidth]
        {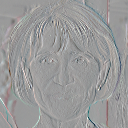}
        \caption{$\bm{\textbf{LH}}$}
    \end{subfigure}
    \begin{subfigure}[t]{0.15\linewidth}
        \includegraphics[width=\linewidth]
        {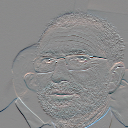} \\
        \vspace{-4mm}
        \includegraphics[width=\linewidth]
        {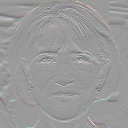}
        \caption{$\bm{\textbf{HL}}$}
    \end{subfigure}
    \begin{subfigure}[t]{0.15\linewidth}
        \includegraphics[width=\linewidth]
        {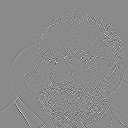} \\
        \vspace{-4mm}
        \includegraphics[width=\linewidth]
        {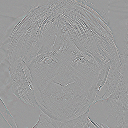}
        \caption{$\bm{\textbf{HH}}$}
    \end{subfigure}
    \begin{subfigure}[t]{0.15\linewidth}
        \includegraphics[width=\linewidth]
        {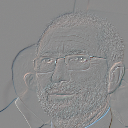} \\
        \vspace{-4mm}
        \includegraphics[width=\linewidth]
        {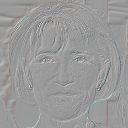}
        \caption{\resizebox{\linewidth}{!}{$\bm{\textbf{LH}}+\bm{\textbf{HL}}+\bm{\textbf{HH}}$}}
    \end{subfigure}
    \vspace{-3.5mm}
    \caption{An illustration of wavelet transformation.}
    \vspace{-8mm}
    \label{fig:wave_pool}
\end{figure}

Wavelet transformation has achieved remarkable success in applications that require decomposing images into domains with different frequencies. Following this idea, we adopt a classic wavelet transformation method, the Haar wavelet, which consists of two operations: wavelet pooling and unpooling. Wavelet pooling contains four kernels, $\{ LL^\top, LH^\top, HL^\top, HH^\top\}$, where the low (L) and high (H) pass filters are $L^\top = \frac{1}{\sqrt{2}}[1, 1]$ and $H^\top = \frac{1}{\sqrt{2}}[-1, 1]$, respectively. The low-pass filter concentrates on the smooth surface which is mostly related to low-frequency signals while the high-pass filter captures most high-frequency signals like vertical, horizontal and diagonal edges. Figure~\ref{fig:wave_pool} shows the information of four frequency domains (\ie, \textbf{LL}, \textbf{LH}, \textbf{HL} and
\textbf{HH}) decomposed from two images by implementing the Haar wavelet transformation. \textbf{LL} mainly consists of information in the low-frequency domain, depicting the overall appearance of an image, while \textbf{LH}, \textbf{HL} and \textbf{HH} contain information representing rich details. We find that the combination of \textbf{LH}, \textbf{HL} and \textbf{HH} can be considered a good approximation of the hidden information $\bm{h}$ (see Figure~\ref{fig:analysis}(g)). Besides, wavelet unpooling is employed to exactly reconstruct the original image from the signal components decomposed via wavelet pooling as follows. We first apply a component-wise transposed-convolution on the signal of each component and then sum all resulted features up to precisely reconstruct the image.

Wavelet transformation is usually applied at the image level, but here we implement it at the feature level. Specifically, we first adopt the above-mentioned wavelet pooling to extract features in the domains of different frequencies from different layers of the encoder (see Figure~\ref{fig:framework}). Then we ignore the information of \textbf{LL}, and apply wavelet unpooling to \textbf{LH}, \textbf{HL} and \textbf{HH} to reconstruct the information for high-frequency components of the original feature.

\subsection{HifaFace}
In this section, we introduce our proposed method called HifaFace. It requires two inputs: the input image $\bm{x}$ and the difference of attributes $\bm{\Delta}$, and outputs the result $\bm{\hat{y}}$ with the target attributes. Figure~\ref{fig:framework} gives an overview of our method which mainly contains the following four parts: 1) the Wavelet-based Generator, 2) the High-frequency Discriminator, 3) the Image-level Discriminator and 4) an attribute Classifier.%

\noindent\textbf{Wavelet-Based Generator.}~Our generator $G$ mainly follows the ``encoder-decoder'' structure, which contains the encoding part, bottleneck part and decoding part. The input image $\bm{x}$ is directly fed to the front of the network. We adopt the widely used AdaIN~\cite{Huang2017ArbitraryST} module in the bottleneck part to input the vector of condition attributes $\bm{\Delta}$. To alleviate the generator's pressure of synthesizing rich details, we propose to use a \emph{wavelet-based skip-connection} to feed the high-frequency information directly to the decoding part of the generator. Specifically, for the $i$-th layer in the encoding part, we adopt wavelet pooling to extract frequency features $\bm{E_{LL}^i}, \bm{E_{LH}^i}, \bm{E_{HL}^i}$ and $\bm{E_{HH}^i}$. Then we ignore the low-frequency feature $\bm{E_{LL}^i}$, and feed the remaining three high-frequency feature maps to the wavelet unpooling module which transforms them to the different frequency domains of the original feature. Finally, we use a skip-connection to feed them to the $(n-i)$-th layer in the decoding part, where $n$ is the number of all layers. This branch aims to maintain the high-frequency details of the input image.

\noindent\textbf{High-Frequency Discriminator.}~To ensure that synthesis results with rich details can be obtained by the generator, we also adopt a high-frequency discriminator $D_H$. For both real images and generated images, we first use wavelet pooling to extract their high-frequency features $\bm{E_{LH}}, \bm{E_{HL}}$ and $\bm{E_{HH}}$. Then we feed them into the high-frequency discriminator, thus encouraging the generator to synthesize images with high-frequency information.

\noindent\textbf{Image-Level Discriminator.}~To encourage the generated image to be realistic, we adopt an image-level discriminator $D_I$ to distinguish between real images and generated images.

\noindent\textbf{Attribute Classifier.}~To guarantee the consistency between synthesis results and their corresponding target attributes, we design an auxiliary attribute classifier $C$. Specifically, it consists of $K$ binary classifiers on top of the feature extractor, where $K$ denotes the number of attributes. To achieve a faster training procedure, we first train the classifier only on the labeled dataset. Then when training the whole HifaFace model, we apply the learned classifier to ensure that the generated image possesses the target attributes.

\subsection{Objective Function}
We train our model using the following losses.

\noindent\textbf{Image-Level Adversarial Loss.}
We adopt the adversarial loss to encourage the generated image to be realistic. Let $\bm{y}$ denote the sampled real images, the image-level adversarial loss is defined as:
\begin{equation}
\mathcal{L}_{GAN}^{I} = \mathbb{E} [ \log D_I(\bm{y}) + \log(1 - D_I(G(\bm{x}, \bm{\Delta}))) ].
\end{equation}
\noindent\textbf{High-Frequency Domain Adversarial Loss.}
To encourage the generator to maintain rich details, we apply the adversarial loss in the high-frequency domain. Here, we choose the combination of three domains (\ie, \textbf{LH}, \textbf{HL} and \textbf{HH}) as the high-frequency domain, and define the high-frequency domain adversarial loss as:
\begin{equation}
\mathcal{L}_{GAN}^{H} = \mathbb{E} [ \log D_H(\bm{x}) + \log(1 - D_H(G(\bm{x}, \bm{\Delta})))].
\end{equation}
\noindent\textbf{Cycle Reconstruction Loss.} In order to guarantee the generated image properly preserving the characteristics of the input image $\bm{x}$ that are invariant to the target attributes, we employ the cycle reconstruction loss which is defined as:
\begin{equation}
\mathcal{L}_{cyc} = \mathbb{E} [ \parallel \bm{x} - G(G(\bm{x}, \bm{\Delta}), -\bm{\Delta}) \parallel _1 ].
\end{equation}
\noindent\textbf{Attribute Classification Loss.} To ensure that the synthesis result $\bm{\hat{y}}$ possesses the desired attributes $\bm{a}_y$, where $a^k_y$ is the $k$-th element of $\bm{a}_y$, we introduce the attribute classification loss to constrain the generator $G$. The attribute classification loss is only applied on the attributes that are changed. Specifically, we use $\Delta^k$ to determine whether the $k$-th attribute has been changed (\ie,  $|\Delta^k| = 1$). Suppose $p^k$ is the probability value of the $k$-th attribute estimated by the classifier $C$, we have the attribute classification loss as:
\begin{equation}
\footnotesize
\begin{aligned}
\mathcal{L}_{ac} =  -\mathbb{E} [\sum_{k=1}^{K} \mathbbm{1}_{\{|\Delta^k| = 1 \}} (a_y^{k}\log {p^{k}}
                     +  (1 - a_y^{k}) \ \log {(1 - p^{k})} )],
\end{aligned}
\end{equation}
where $\mathbbm{1}$ denotes the indicator function, which is equal to $1$ when the condition is satisfied. The attribute classification loss restricts the generator to synthesize high-quality images with the corresponding target attributes.

\noindent\textbf{Attribute Regression Loss.} Most existing works only consider discrete editing operations~\cite{Choi2018StarGANUG, Liu2019STGANAU, chusscgan} or support a limited range of continuous editing~\cite{Wu2019RelGANMI}, making them less practical in real-world scenarios. We expect to precisely control the attributes with a scale factor $\alpha$, requiring that the generator is capable of synthesizing face images with different levels of attribute editing. Thus the output image can be denoted as $\bm{y}_{\alpha} = G(\bm{x}, \alpha \cdot \bm{\Delta})$, in which $\alpha \in [0,2]$. For example, we may want the people in an image to smile. If $\alpha=0.5$, we expect a smile, if $\alpha=2$, what we expect is laughing. We calculate the attribute regression loss by:
\begin{align}
\begin{split}
       \mathcal{L}_{ar} =& \ \mathbb{E} [ (d(\bm{f}_0, \bm{f}_{\alpha}) - d(\bm{f}_1, \bm{f}_0)) - (\alpha - 1)],
\end{split}
\end{align}
where $\bm{f}_{\alpha}$ means the $\ell_2$ normalized feature vector extracted by the attribute classifier $C$: $\bm{f}_{\alpha} = C(G(\bm{x}, \alpha \cdot \bm{\Delta}))$, in which $d(., .)$ denotes the $\ell_2$ distance of two feature vectors.

\begin{figure*}
    \captionsetup[subfigure]{labelformat=empty}
    \captionsetup[subfigure]{aboveskip=1pt} %
    \centering
    \begin{subfigure}[t]{\dimexpr0.11\textwidth+9pt\relax}
        \makebox[9pt]{\raisebox{26pt}{\rotatebox[origin=c]{90}{\small GANimation}}}%
        \includegraphics[width=\dimexpr\linewidth-9pt\relax]
        {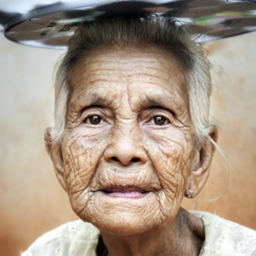}
        \makebox[9pt]{\raisebox{26pt}{\rotatebox[origin=c]{90}{STGAN}}}%
        \includegraphics[width=\dimexpr\linewidth-9pt\relax]
        {images/comp_12_in}
        \makebox[9pt]{\raisebox{26pt}{\rotatebox[origin=c]{90}{RelGAN}}}%
        \includegraphics[width=\dimexpr\linewidth-9pt\relax]
        {images/comp_12_in}
        \makebox[9pt]{\raisebox{26pt}{\rotatebox[origin=c]{90}{IFGAN}}}%
        \includegraphics[width=\dimexpr\linewidth-9pt\relax]
        {images/comp_12_in}
        \makebox[9pt]{\raisebox{26pt}{\rotatebox[origin=c]{90}{StyleFlow}}}%
        \includegraphics[width=\dimexpr\linewidth-9pt\relax]
        {images/comp_12_in}
        \makebox[9pt]{\raisebox{26pt}{\rotatebox[origin=c]{90}{\textbf{HifaFace}}}}%
        \includegraphics[width=\dimexpr\linewidth-9pt\relax]
        {images/comp_12_in}
        \caption{\quad Input}
    \end{subfigure}\hspace{0.4mm}%
    \begin{subfigure}[t]{0.11\textwidth}
        \includegraphics[width=\textwidth]
        {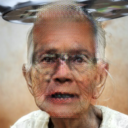}
        \includegraphics[width=\textwidth]
        {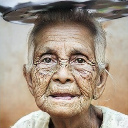}
        \includegraphics[width=\textwidth]
        {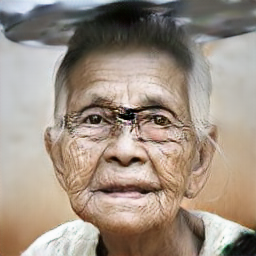}
        \includegraphics[width=\textwidth]
        {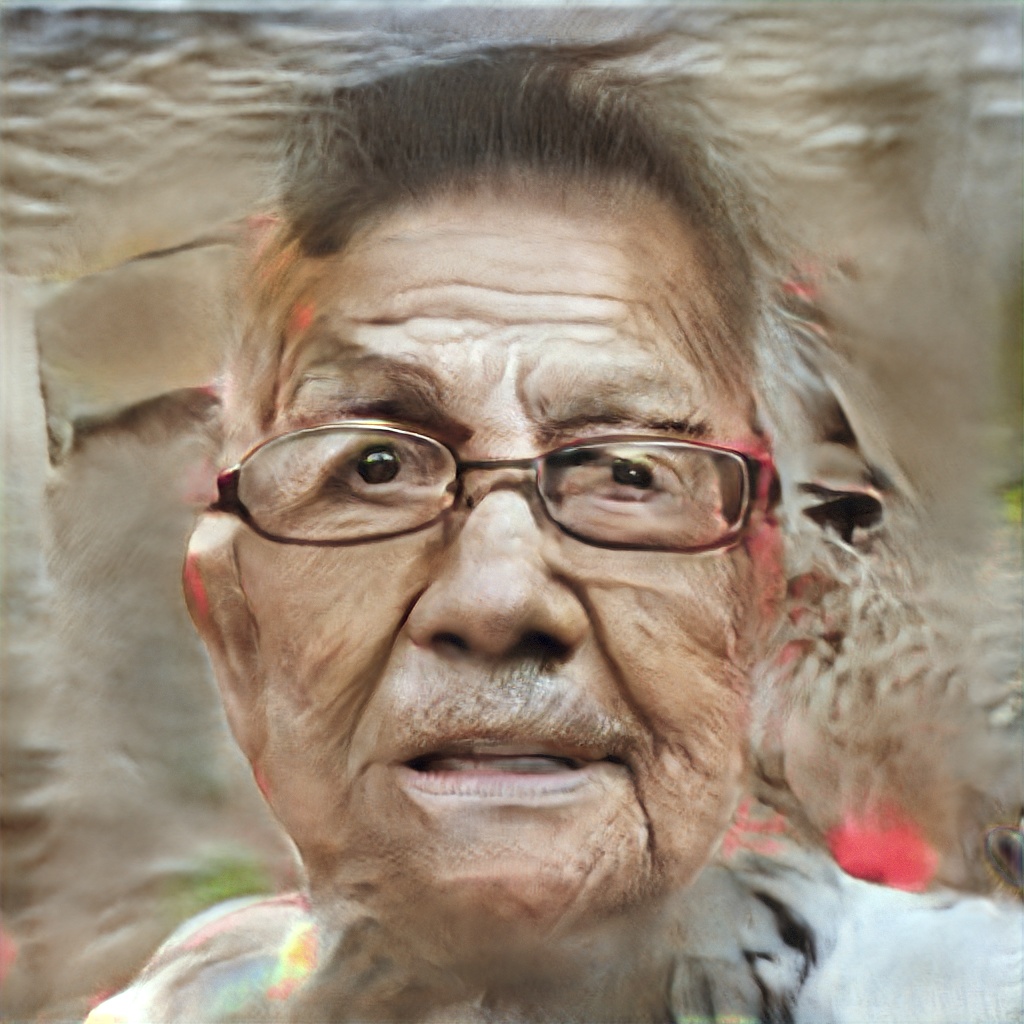}
        \includegraphics[width=\textwidth]
        {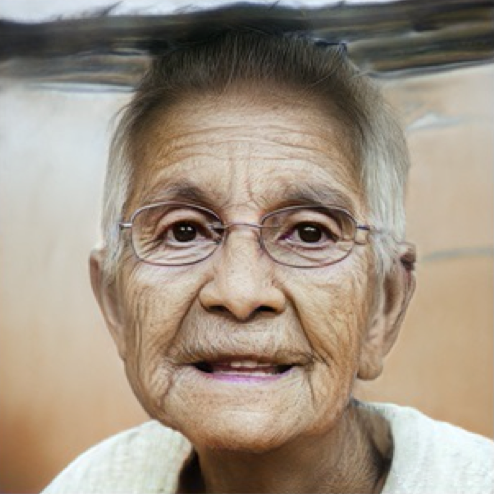}
        \includegraphics[width=\textwidth]
        {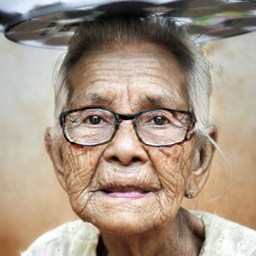}
        \caption{Eyeglasses}
    \end{subfigure}\hspace{0.4mm}%
    \begin{subfigure}[t]{0.11\textwidth}
        \includegraphics[width=\textwidth]
        {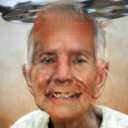}
        \includegraphics[width=\textwidth]
        {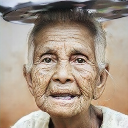}
        \includegraphics[width=\textwidth]
        {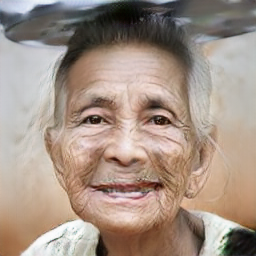}
        \includegraphics[width=\textwidth]
        {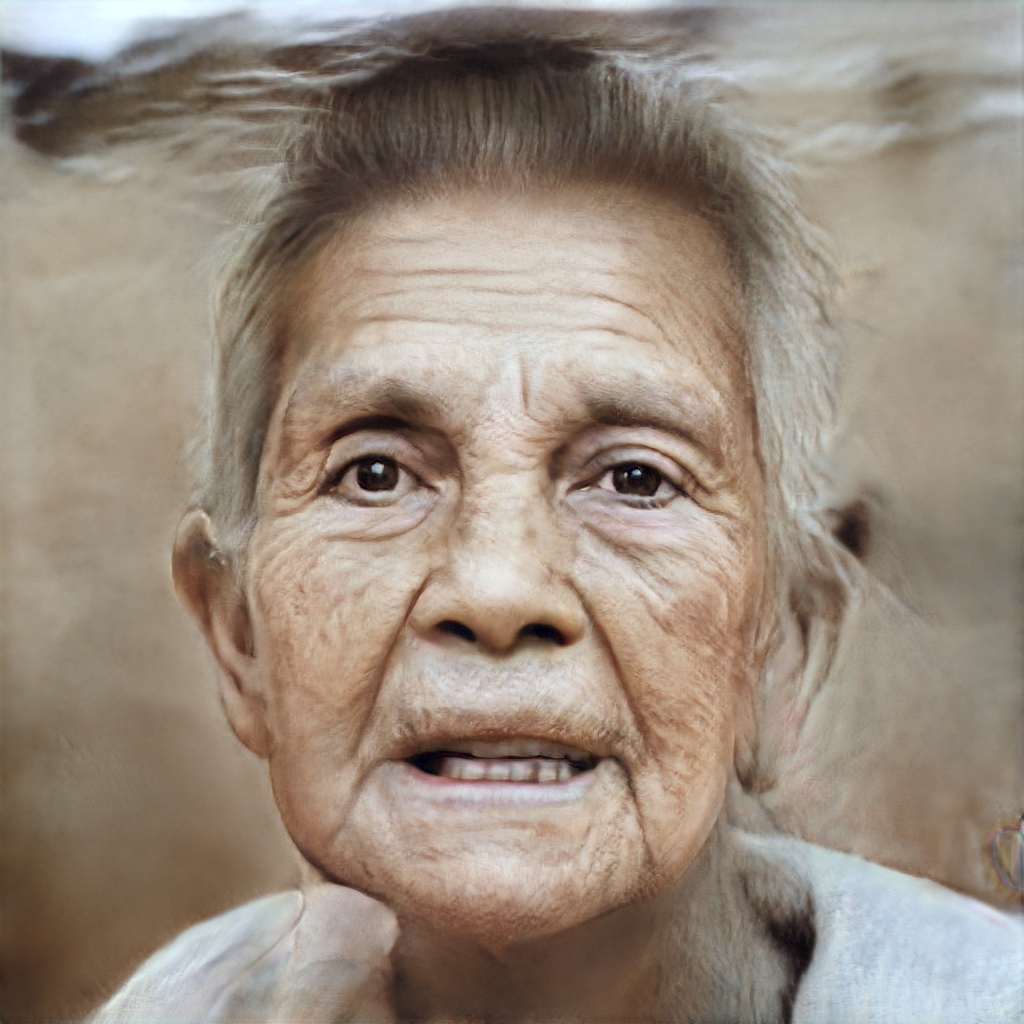}
        \includegraphics[width=\textwidth]
        {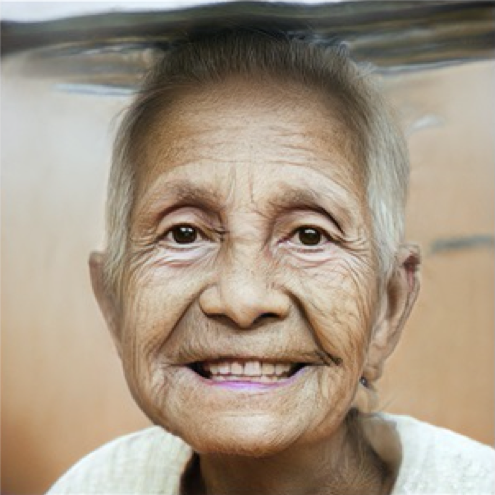}
        \includegraphics[width=\textwidth]
        {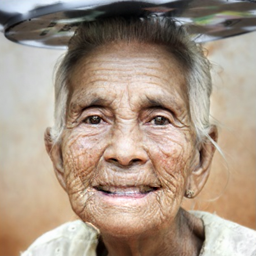}
        \caption{Smile}
    \end{subfigure}\hspace{0.4mm}%
    \begin{subfigure}[t]{0.11\textwidth}
        \includegraphics[width=\textwidth]
        {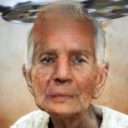}
        \includegraphics[width=\textwidth]
        {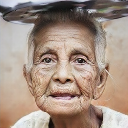}
        \includegraphics[width=\textwidth]
        {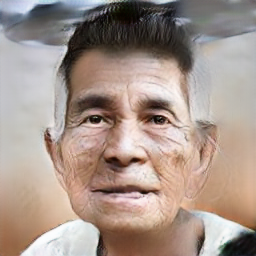}
        \includegraphics[width=\textwidth]
        {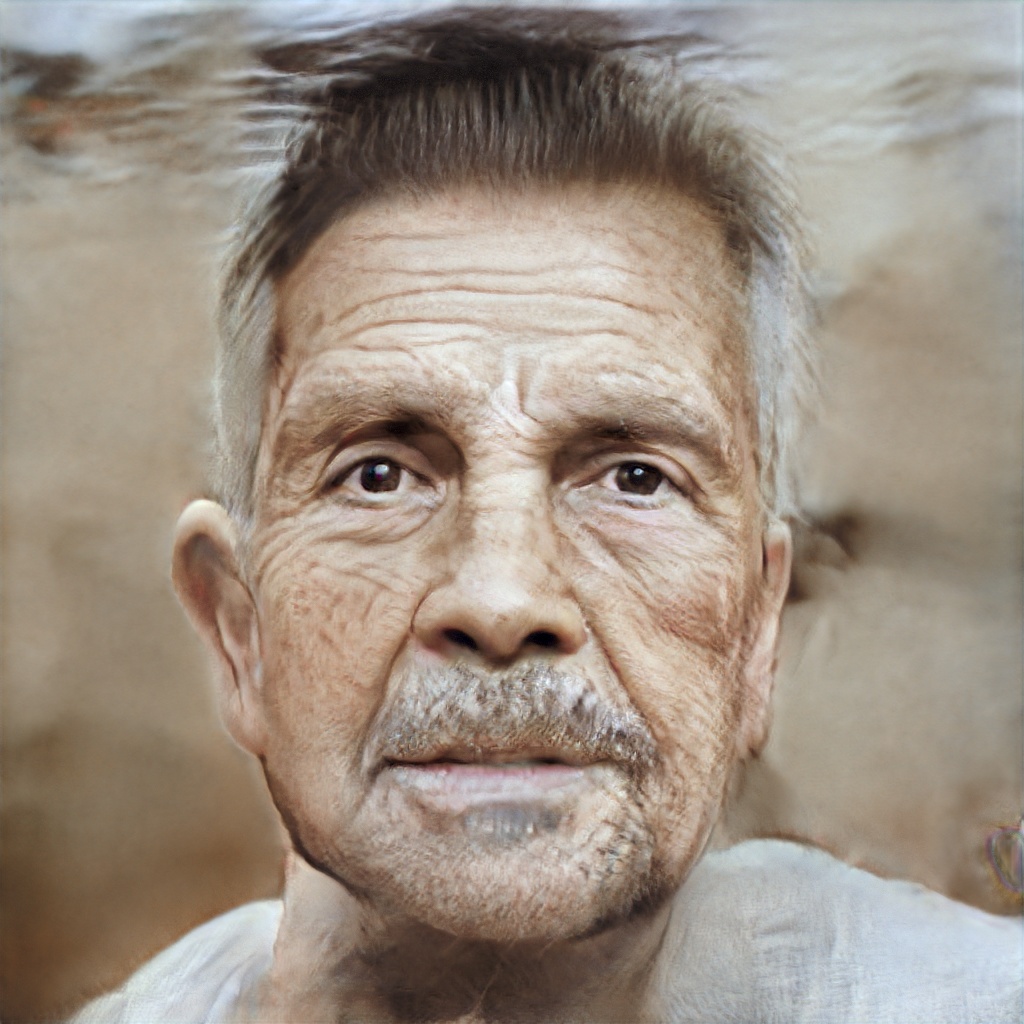}
        \includegraphics[width=\textwidth]
        {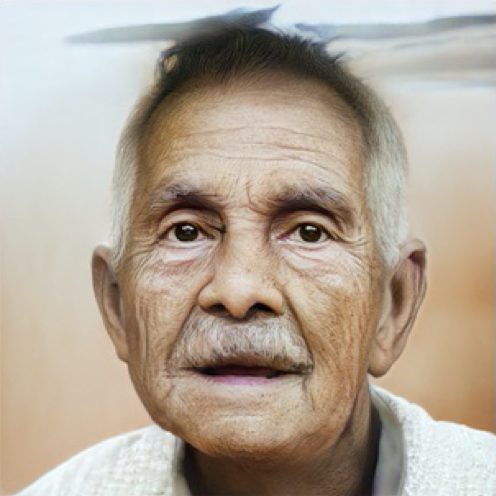}
        \includegraphics[width=\textwidth]
        {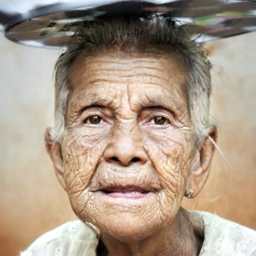}
        \caption{Gender}
    \end{subfigure}\hspace{0.4mm}%
    \begin{subfigure}[t]{0.11\textwidth}
        \includegraphics[width=\textwidth]
        {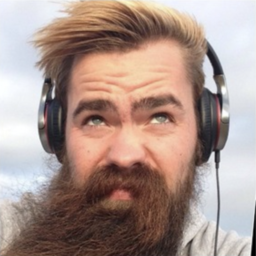}
        \includegraphics[width=\textwidth]
        {images/comp_13_in}
        \includegraphics[width=\textwidth]
        {images/comp_13_in}
        \includegraphics[width=\textwidth]
        {images/comp_13_in}
        \includegraphics[width=\textwidth]
        {images/comp_13_in}
        \includegraphics[width=\textwidth]
        {images/comp_13_in}
        \caption{Input}
    \end{subfigure}\hspace{0.4mm}%
    \begin{subfigure}[t]{0.11\textwidth}
        \includegraphics[width=\textwidth]
        {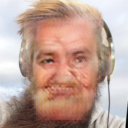}
        \includegraphics[width=\textwidth]
        {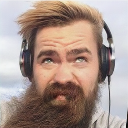}
        \includegraphics[width=\textwidth]
        {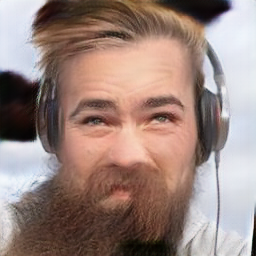}
        \includegraphics[width=\textwidth]
        {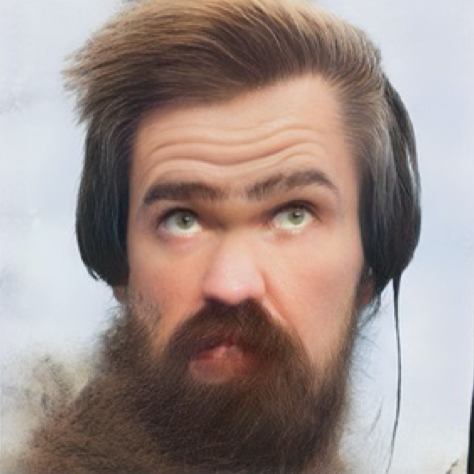}
        \includegraphics[width=\textwidth]
        {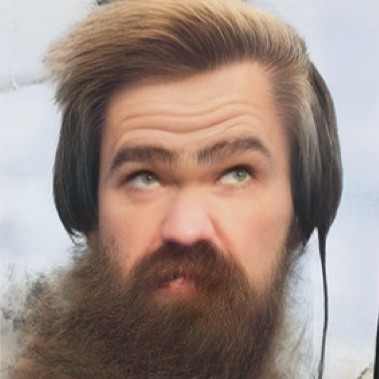}
        \includegraphics[width=\textwidth]
        {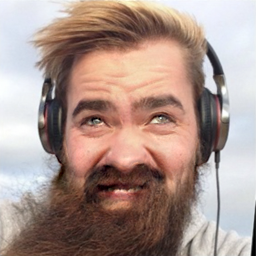}
        \caption{Smile}
    \end{subfigure}\hspace{0.4mm}%
    \begin{subfigure}[t]{0.11\textwidth}
        \includegraphics[width=\textwidth]
        {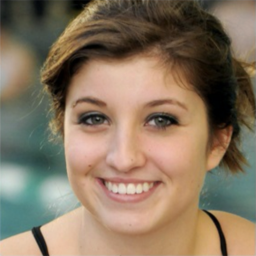}
        \includegraphics[width=\textwidth]
        {images/comp_14_in}
        \includegraphics[width=\textwidth]
        {images/comp_14_in}
        \includegraphics[width=\textwidth]
        {images/comp_14_in}
        \includegraphics[width=\textwidth]
        {images/comp_14_in}
        \includegraphics[width=\textwidth]
        {images/comp_14_in}
        \caption{Input}
    \end{subfigure}\hspace{0.4mm}%
    \begin{subfigure}[t]{0.11\textwidth}
        \includegraphics[width=\textwidth]
        {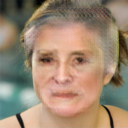}
        \includegraphics[width=\textwidth]
        {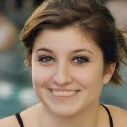}
        \includegraphics[width=\textwidth]
        {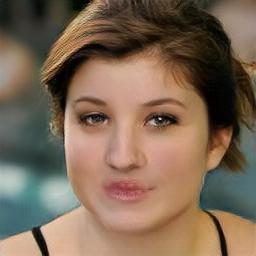}
        \includegraphics[width=\textwidth]
        {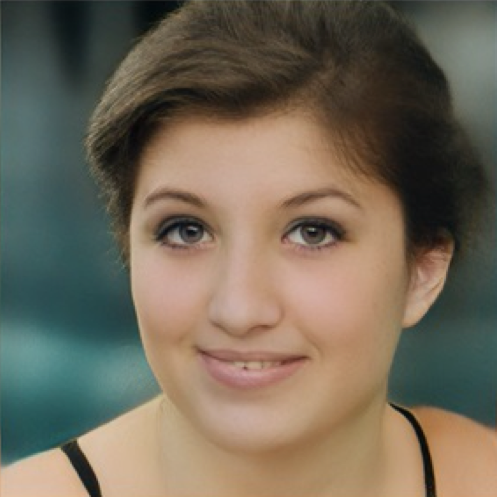}
        \includegraphics[width=\textwidth]
        {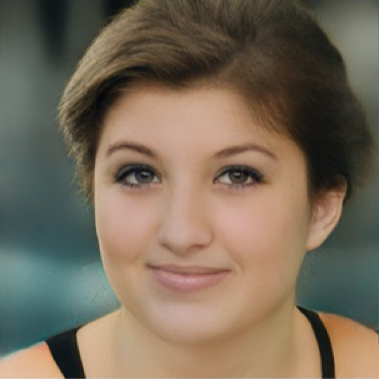}
        \includegraphics[width=\textwidth]
        {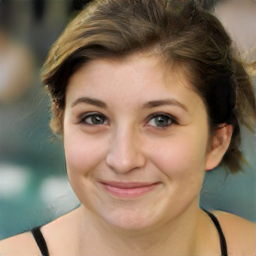}
        \caption{Close mouth}
    \end{subfigure}
    \vspace{-4mm}
    \caption{Comparison of attribute-based face editing results obtained by the proposed method and other state-of-the-art approaches. All test images are wild images used in~\cite{Abdal2020StyleFlowAE}. }
    \label{fig:comparison}
    \vspace{-6mm}
\end{figure*}

\noindent\textbf{Objective Function.} The overall loss function of our model is:
\begin{equation}
\footnotesize
\begin{aligned}
\mathcal{L} = \lambda_{ar} \mathcal{L}_{ar} + \lambda_{ac} \mathcal{L}_{ac} + \lambda_{GAN}^I \mathcal{L}_{GAN}^I + \lambda_{GAN}^H \mathcal{L}_{GAN}^H+ \lambda_{cyc} \mathcal{L}_{cyc},
\end{aligned}
\end{equation}
where $\lambda_{ar}$, $\lambda_{ac}$, $\lambda_{GAN}^I$, $\lambda_{GAN}^H$ and $\lambda_{cyc}$ denote the weights of corresponding loss terms, respectively.

\subsection{Semi-Supervised Learning} \label{semi}
Another important challenge for attribute editing is the limited size of the training dataset. Existing supervised learning based models~\cite{Choi2018StarGANUG, Liu2019STGANAU, Wu2019RelGANMI, chusscgan} are typically incapable of handling faces in the wild, especially for those with rich details, pose variance or complex background. However, manually annotating the attributes for a large number of face images is time-consuming, so we adopt a semi-supervised learning strategy to exploit large amounts of unlabeled data.
We use our attribute classifier $C$ to predict attributes for all images in the unlabeled dataset $\mathcal{D}_u$, and assign each image with the corresponding prediction result as the pseudo label.
Then, both the labeled dataset $\mathcal{D}_l$ and the pseudo-labeled dataset $\mathcal{D}_u$ are employed for training.

\section{Experiments}
\noindent\textbf{Datasets.} We evaluate our model on the CelebA-HQ~\cite{Karras2018ProgressiveGO} and FFHQ~\cite{Karras2019ASG} datasets. %
The classification model trained on CelebA-HQ is applied to get the pseudo labels for all images in FFHQ. The image resolution is chosen as $256 \times 256$ in our experiments.

\noindent\textbf{Implementation Details.} In the generator $G$, the wavelet-based skip-connection is employed between all the three encoding and decoding blocks. We apply the Spectral Normalization (SN)~\cite{Miyato2018SpectralNF} for both $D_H$ and $D_I$. For the attribute classifier $C$, we use the pre-trained ResNet-18~\cite{He2016DeepRL} as the feature extractor, and two non-linear classification layers are followed for each attribute. The classifier is fine-tuned on CelebA-HQ~\cite{Karras2018ProgressiveGO} with $92.8\%$ accuracy on the test set. We use the Adam optimizer~\cite{Kingma2015AdamAM} with TTUR~\cite{Heusel2017GANsTB} for training.%

\noindent\textbf{Baselines.} We compare our approach with two typical types of face editing models: 1) Latent space manipulation based models with pre-trained GANs: InterFaceGAN~\cite{Shen2020InterFaceGANIT} and StyleFlow~\cite{Abdal2020StyleFlowAE}; 2) Image-to-image translation-based methods: GANimation~\cite{pumarola2018ganimation}, STGAN~\cite{Liu2019STGANAU}, RelGAN~\cite{Wu2019RelGANMI}, CAFE-GAN~\cite{gicafe}, CooGAN~\cite{Chen2020CooGANAM} and SSCGAN~\cite{chusscgan}. %

\begin{figure*}[t]
    \captionsetup[subfigure]{labelformat=empty}
    \captionsetup[subfigure]{aboveskip=1pt} %
    \centering
    \begin{subfigure}[t]{\dimexpr0.09\textwidth+11pt\relax}
        \makebox[11pt]{\raisebox{18pt}{\rotatebox[origin=c]{90}{RelGAN}}}%
        \includegraphics[width=\dimexpr\linewidth-11pt\relax]{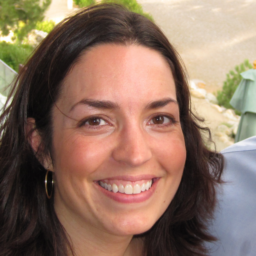}
        \vspace{-4mm}

        \makebox[11pt]{\raisebox{18pt}{\rotatebox[origin=c]{90}{IFGAN}}}%
        \includegraphics[width=\dimexpr\linewidth-11pt\relax]{images/triplet_wo_2062_glasses_in}
        \vspace{-4mm}

        \makebox[11pt]{\raisebox{18pt}{\rotatebox[origin=c]{90}{\textbf{HifaFace}}}}%
        \includegraphics[width=\dimexpr\linewidth-11pt\relax]{images/triplet_wo_2062_glasses_in}
        \caption{\quad (a) Input}
    \end{subfigure}\hspace{0.8mm}%
    \begin{subfigure}[t]{0.81\textwidth}
        \includegraphics[width=\textwidth]{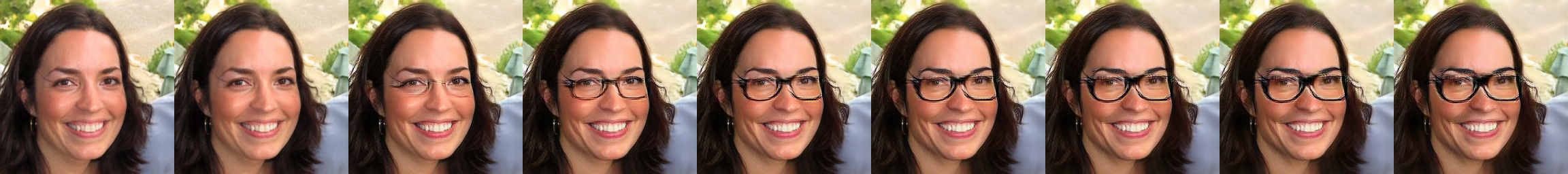}
        \vspace{-4mm}

        \includegraphics[width=\textwidth]{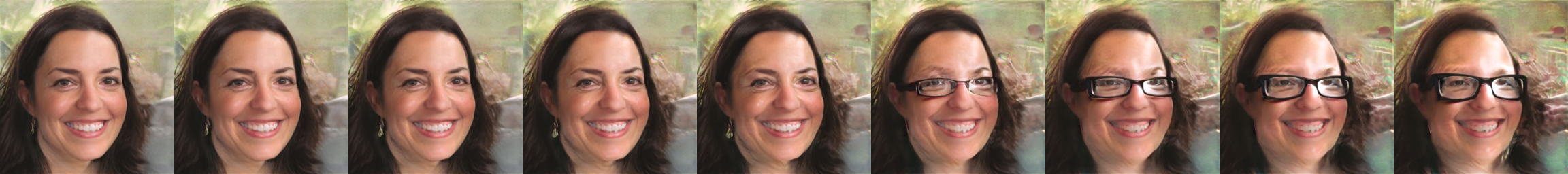}
        \vspace{-4mm}

        \includegraphics[width=\textwidth]{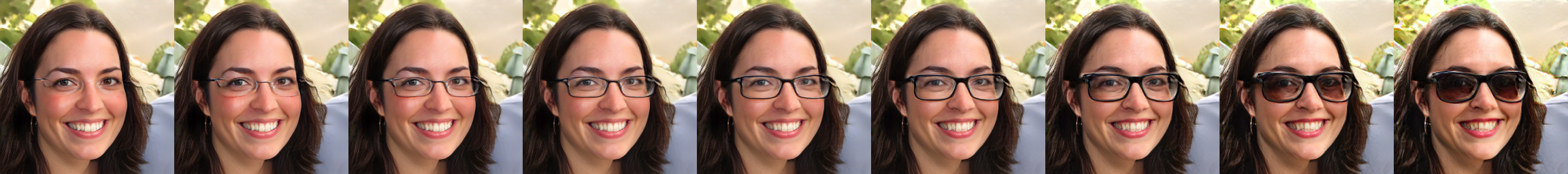}
        \caption{(b) Interpolation of the ``eyeglasses'' attribute, in range of $[0.4, 2.0]$ with an interval of $0.2$.}
    \end{subfigure}%
    \vspace{-3mm}
    \caption{Interpolation results obtained by RelGAN~\cite{Wu2019RelGANMI}, InterFaceGAN (IFGAN)~\cite{Shen2020InterFaceGANIT} and our HifaFace.}
    \label{fig:arbitrary_comparison}
    \vspace{-3mm}
\end{figure*}

\noindent\textbf{Evaluation Metrics.} We apply several quantitative metrics to evaluate the performance of different face editing methods: 1) Frechét inception distance (FID)~\cite{Heusel2017GANsTB}; 2) Quality Score (QS)~\cite{Gu2020GIQAGI}, which evaluates the quality of each sample; 3) Acc., which measures the classification accuracy of attributes for synthesized face images; 4) In addition to these commonly used metrics, we utilize the Self-Reconstruction Error (SRE) to quantify the models' capability of synthesizing rich details. Specifically, we compute the $\ell_1$ distance between the original image and its projection for the latent space based methods. For the image-to-image translation-based methods, we calculate the $\ell_1$ distance between the input image's self-reconstruction image $\bm{\overline{y}}$ and the reconstructed image $\bm{\overline{x}}$ of the self-reconstruction result.

\subsection{Comparison with State of the Art Methods}
In this section, we compare our face editing method with other existing approaches from two perspectives: attribute-based face editing and arbitrary face editing.%

\begin{table}[t]
	\centering
	\begin{tabular}{lcccc}
    \toprule
    Methods      &  FID $\downarrow$ &  Acc. $\uparrow$  & QS $\uparrow$ & SRE $\downarrow$ \\
    \midrule
    GANimation   &  15.72 &  64.7   & 0.710  & 0.145  \\
    STGAN        &  14.78 &  83.2   & 0.543  & 0.041  \\
    RelGAN       &  10.13 &  83.6   & 0.729  & 0.024  \\
    CAFE-GAN     &  -     &  81.9   & -      & -      \\
    CooGAN       &  -     &  83.8   & -      & -      \\
    SSCGAN       &  4.69  &  94.2   & -      & -      \\
    InterFaceGAN &  -     &  -      & -      & 0.163  \\
    \textbf{HifaFace}   &  \textbf{4.04}     &  \textbf{97.5}   & \textbf{0.803}  & \textbf{0.021}  \\    \bottomrule
	\end{tabular}
	\vspace{-3.5mm}
	\caption{Quantitative results of different methods.}
	\label{tab:comparison}
	\vspace{-6.5mm}
\end{table}

\begin{table*}[t]
    \scriptsize
	\centering
    \begin{tabular}{lcccccccccccccc}
    \toprule
    Methods & \makecell{Arched \\ Eyebrows} & \makecell{Black \\ Hair}  & \makecell{Blond \\ Hair} & \makecell{Brown \\ Hair} & \makecell{Eye-\\glasses} & \makecell{Gray \\ Hair} & \makecell{Heavy \\ Makeup} & Male & \makecell{Mouth \\ Open} & Mustache & \makecell{No \\ Beard} & Smiling & Young & Average \\
    \midrule
    GANimation & 69.2 & 74.0 & 52.8 & 54.1 & 87.2 & 77.1 & 75.9 & 65.9 & 57.2 & 54.8 & 44.6 & 67.7 & 57.5 & 64.7 \\
    STGAN     & 80.2 & 76.3 & 78.9 & 82.9 & 86.1 & 84.3 & 87.3 & 86.5 & 88.4 & 75.7 & 90.3 & 84.6 & 80.3 & 83.2 \\
    RelGAN    & 85.4 & 74.8 & 84.7 & 91.4 & 93.9 & 91.4 & 79.5 & 73.7 & 91.6 & 80.5 & 91.6 & 69.9 & 78.3 & 83.6 \\
    CAFE-GAN  & - & - & 88.1 & - & - & - & - & 95.2 & 97.2 & 40.1 & - & - & 88.6 & 81.9 \\
    CooGAN    & 89.1 & 80.1 & 84.2 & 64.1 & 99.8 & - & - & 85.0 & 94.9 & 59.4 & 96.3 & - & 85.2 & 83.8 \\
    SSCGAN    & 96.5 & \textbf{99.3} & - & - & \textbf{99.9} & - & - & \textbf{99.1} & \textbf{99.9} & 65.7 & - & - & \textbf{99.0} & 94.2 \\
    \textbf{HifaFace}      & \textbf{98.4} & 94.9 & \textbf{98.4} & \textbf{92.7} & 98.9 & \textbf{98.3} & \textbf{96.0} & 98.9 & 99.0 & \textbf{98.2} & \textbf{97.7} & \textbf{98.3} & 97.5 & \textbf{97.5} \\
    \bottomrule
	\end{tabular}
	\vspace{-3mm}
	\caption{The attribute editing accuracy of our method and other image-to-image translation-based face editing approaches.}
	\vspace{-5mm}
	\label{tab:comparison_all_attr}
\end{table*}

\noindent\textbf{Attribute-Based Face Editing}~We compare our method with some recent works: GANimation~\cite{pumarola2018ganimation} STGAN~\cite{Liu2019STGANAU}, RelGAN~\cite{Wu2019RelGANMI}, InterFaceGAN~\cite{Shen2020InterFaceGANIT} and StyleFlow~\cite{Abdal2020StyleFlowAE}, and show the qualitative results in Figure~\ref{fig:comparison}. GANimation utilizes a masking mechanism to force the generator to edit the regions that need to be edited. Unsatisfactory results indicate that the attention mask mechanism does not work well for the face editing task. STGAN is a typical method that adopts the reconstruction loss between the input image and its self-reconstruction instead of cycle consistency. We observe that their results keep the rich details but don't have the desired attributes. On the contrary, RelGAN is designed based on cycle consistency. We observe that their results lose rich details and have a lot of artifacts.
Meanwhile, for the latent space based methods InterFaceGAN~\cite{Shen2020InterFaceGANIT} and StyleFlow~\cite{Abdal2020StyleFlowAE}, they lose too many details of the original input image. %
Compared to those existing approaches, our method obtains impressive results with higher quality, which not only preserve the rich details of the input face but also perfectly satisfy the desired attributes.
Table~\ref{tab:comparison} compares the quantitative results of different methods, from which we can see that our method clearly outperforms others considering the values of FID and QS, and generates face images containing more rich details as well as low self-reconstruction errors. Furthermore, the high Acc. value indicates that our method has better capability of editing facial attributes. In Table~\ref{tab:comparison_all_attr}, we evaluate the attribute editing accuracy for each attribute. We can see that our method obtains the best results for most attributes.

\begin{figure*}[t]
    \captionsetup[subfigure]{labelformat=empty}
    \captionsetup[subfigure]{aboveskip=1pt} %
    \centering
    \begin{subfigure}[t]{\dimexpr0.096\textwidth+11pt\relax}
        \makebox[11pt]{\raisebox{18pt}{\rotatebox[origin=c]{90}{w/o $L_{ar}$}}}%
        \includegraphics[width=\dimexpr\linewidth-11pt\relax]{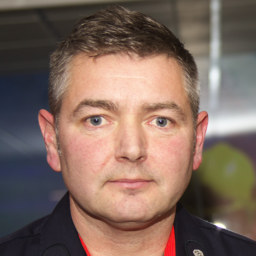}
        \vspace{-4mm}

        \makebox[11pt]{\raisebox{18pt}{\rotatebox[origin=c]{90}{\textbf{HifaFace}}}}%
        \includegraphics[width=\dimexpr\linewidth-11pt\relax]{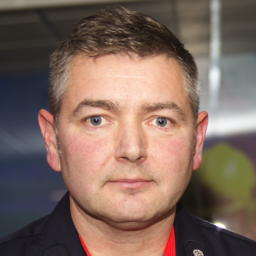}
        \caption{\quad (a) Input}
    \end{subfigure}\hspace{0.8mm}%
    \begin{subfigure}[t]{0.864\textwidth}
        \includegraphics[width=\textwidth]{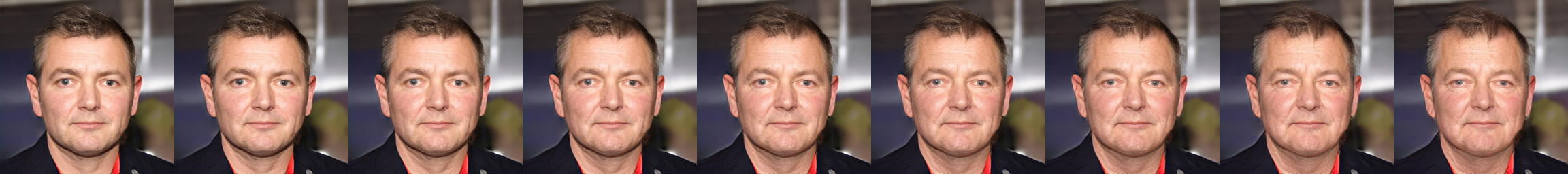}
        \vspace{-4mm}

        \includegraphics[width=\textwidth]{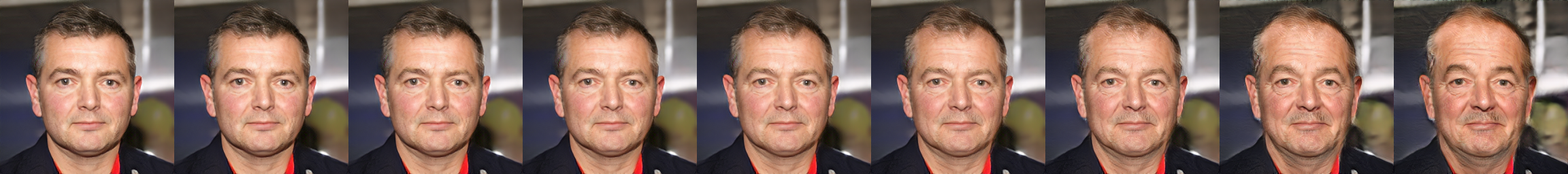}
        \caption{(b) Interpolation of the ``old'' attribute, in range of $[0.4, 2.0]$ with an interval of $0.2$.}
    \end{subfigure}%
    \vspace{-3mm}
    \caption{Comparison of interpolation results obtained using our models without and with the attribute regression loss.}
    \vspace{-5mm}
    \label{fig:lar_comparison}
\end{figure*}

\begin{figure}[t]
    \captionsetup[subfigure]{labelformat=empty}
    \captionsetup[subfigure]{aboveskip=1pt} %
    \centering
    \begin{subfigure}[t]{\dimexpr0.15\linewidth+9pt\relax}
        \makebox[9pt]{\raisebox{20pt}{\rotatebox[origin=c]{90}{\tiny{w/ VS in G}}}}%
        \includegraphics[width=\dimexpr\linewidth-9pt\relax]
        {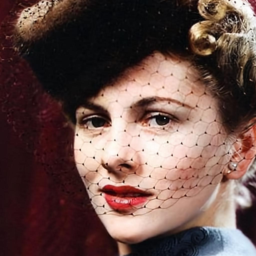}
        \makebox[9pt]{\raisebox{20pt}{\rotatebox[origin=c]{90}{\tiny{w/o WS in G }}}}%
        \includegraphics[width=\dimexpr\linewidth-9pt\relax]
        {images/wave_no_51_in}
        \makebox[9pt]{\raisebox{20pt}{\rotatebox[origin=c]{90}{\tiny{w/o $D_H$}}}}%
        \includegraphics[width=\dimexpr\linewidth-9pt\relax]
        {images/wave_no_51_in}
        \makebox[9pt]{\raisebox{20pt}{\rotatebox[origin=c]{90}{\tiny{\textbf{HifaFace}}}}}%
        \includegraphics[width=\dimexpr\linewidth-9pt\relax]
        {images/wave_no_51_in}
        \caption{\quad Input}
    \end{subfigure}
    \begin{subfigure}[t]{0.15\linewidth}
        \includegraphics[width=\linewidth]
        {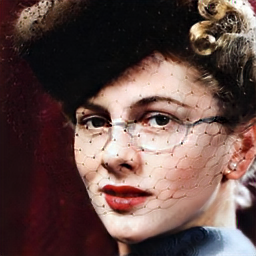}
        \includegraphics[width=\linewidth]
        {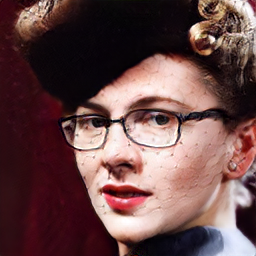}
        \includegraphics[width=\linewidth]
        {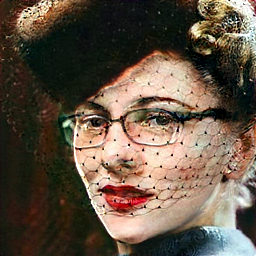}
        \includegraphics[width=\linewidth]
        {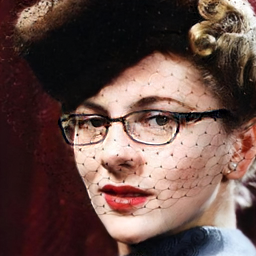}
        \caption{\scriptsize{Eyeglasses}}
    \end{subfigure}
    \begin{subfigure}[t]{0.15\linewidth}
        \includegraphics[width=\linewidth]
        {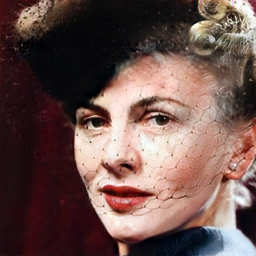}
        \includegraphics[width=\linewidth]
        {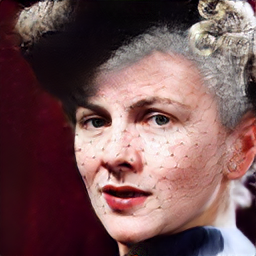}
        \includegraphics[width=\linewidth]
        {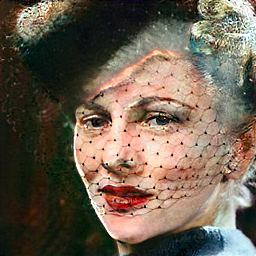}
        \includegraphics[width=\linewidth]
        {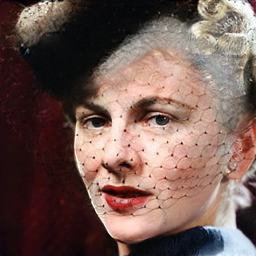}
        \caption{Gray hair}
    \end{subfigure}
    \begin{subfigure}[t]{0.15\linewidth}
        \includegraphics[width=\linewidth]
        {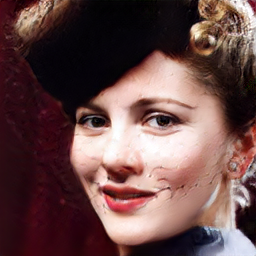}
        \includegraphics[width=\linewidth]
        {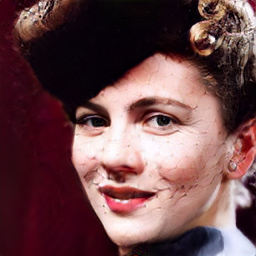}
        \includegraphics[width=\linewidth]
        {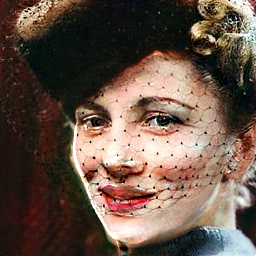}
        \includegraphics[width=\linewidth]
        {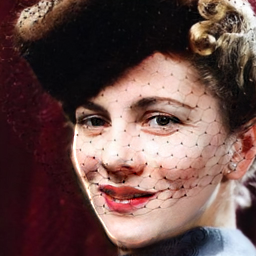}
        \caption{Smile}
    \end{subfigure}
    \begin{subfigure}[t]{0.15\linewidth}
        \includegraphics[width=\linewidth]
        {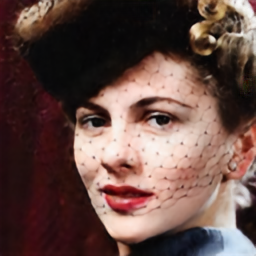}
        \includegraphics[width=\linewidth]
        {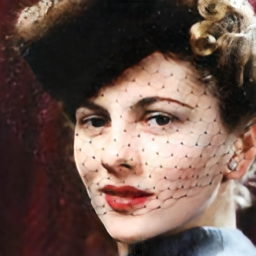}
        \includegraphics[width=\linewidth]
        {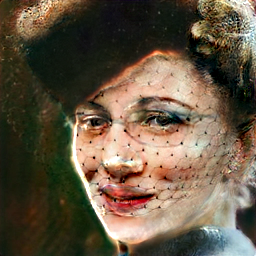}
        \includegraphics[width=\linewidth]
        {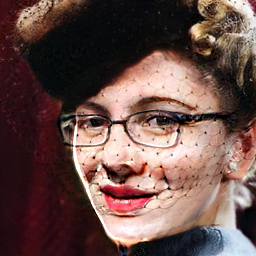}
        \caption{\scriptsize{+Eyeglasses}}
    \end{subfigure}
    \begin{subfigure}[t]{0.15\linewidth}
        \includegraphics[width=\linewidth]
        {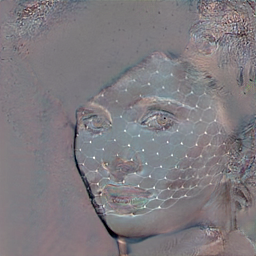}
        \includegraphics[width=\linewidth]
        {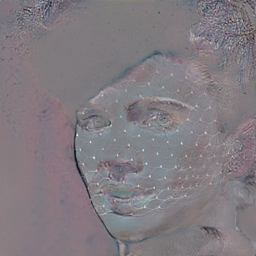}
        \includegraphics[width=\linewidth]
        {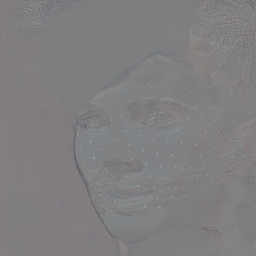}
        \includegraphics[width=\linewidth]
        {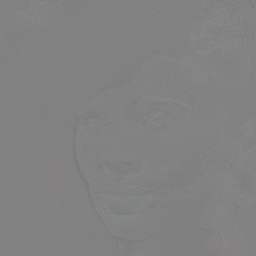}
        \caption{Hidden}
    \end{subfigure}
    \vspace{-3mm}
    \caption{Ablation studies of our proposed model.}
    \label{fig:ablation}
    \vspace{-4mm}
\end{figure}

\begin{table}[t]
	\centering
	\begin{tabular}{lcccc}
		\toprule
        Methods          & FID $\downarrow$  & Acc. $\uparrow$   & QS $\uparrow$  & SRE $\downarrow$  \\
        \midrule
        w/ VS in G       & 5.50              &  94.8  &  0.769     &  0.068    \\
        w/o WS in G      & 5.22              &  96.2  &  0.790     &  0.049    \\
        w/o $D_H$        & 5.34              &  96.4  &  0.762     &  0.057    \\
        \textbf{HifaFace}     & \textbf{4.04}    &  \textbf{97.5} &  \textbf{0.803} &  \textbf{0.021}    \\
        \bottomrule
	\end{tabular}
	\vspace{-3mm}
	\caption{Quantitative results for ablation studies.}
	\vspace{-6.5mm}
	\label{tab:ablation}
\end{table}

\begin{figure}[t]
    \captionsetup[subfigure]{aboveskip=1pt} %
    \captionsetup[subfigure]{labelformat=empty}
    \centering
    \begin{subfigure}[t]{\dimexpr0.1\textwidth+11pt\relax}
        \makebox[11pt]{\raisebox{18pt}{\rotatebox[origin=c]{90}{w/o SL}}}%
        \includegraphics[width=\dimexpr\linewidth-11pt\relax]{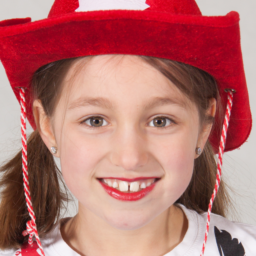}
        \vspace{-4mm}

        \makebox[11pt]{\raisebox{18pt}{\rotatebox[origin=c]{90}{\textbf{HifaFace}}}}%
        \includegraphics[width=\dimexpr\linewidth-11pt\relax]{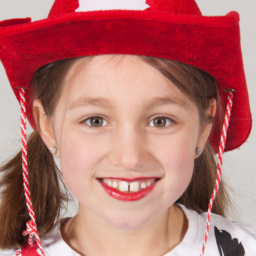}
        \caption{\quad Input}
    \end{subfigure}\hspace{0.6mm}%
    \begin{subfigure}[t]{0.1\textwidth}
        \includegraphics[width=\textwidth]{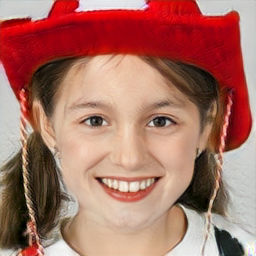}
        \vspace{-4mm}

        \includegraphics[width=\textwidth]{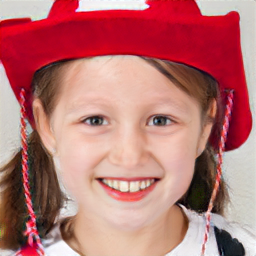}
        \caption{Eyebrows}
    \end{subfigure}\hspace{0.6mm}%
    \begin{subfigure}[t]{0.1\textwidth}
        \includegraphics[width=\textwidth]{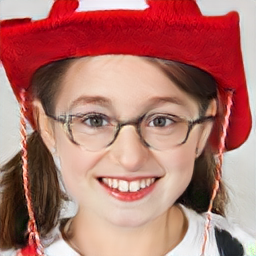}
        \vspace{-4mm}

        \includegraphics[width=\textwidth]{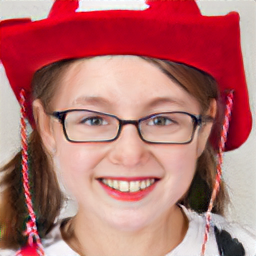}
        \caption{Eyeglasses}
    \end{subfigure}\hspace{0.6mm}%
    \begin{subfigure}[t]{0.1\textwidth}
        \includegraphics[width=\textwidth]{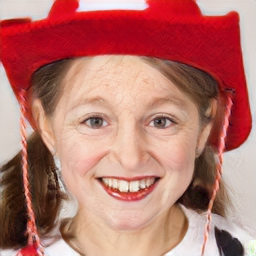}
        \vspace{-4mm}

        \includegraphics[width=\textwidth]{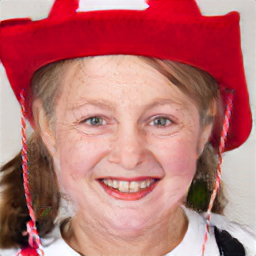}
        \caption{Old}
    \end{subfigure}
    \vspace{-3mm}
    \caption{Comparison of synthesis results obtained by our models without and with semi-supervised learning.}
    \label{fig:semi_comparison}
    \vspace{-6.5mm}
\end{figure}

\noindent\textbf{Arbitrary Face Editing.}
In real-world scenarios, a fine-grained and wider-range control for each attribute is very useful. Previous works such as RelGAN~\cite{Wu2019RelGANMI} and InterFaceGAN~\cite{Shen2020InterFaceGANIT} also provide continuous control for facial attribute editing. We perform attribute interpolation in the range of $\lambda \in [0.4, 2.0]$ with an interval of $0.2$, where $\lambda=0$ and $\lambda=1$ denote that the desired attributes are equal to the source and target attributes, respectively. We show the visualization results in Figure~\ref{fig:arbitrary_comparison}, which shows that our model generates smoother and higher-quality interpolation results compared to InterFaceGAN~\cite{Shen2020InterFaceGANIT}. When $\lambda$ is larger than $1$, interpolation results of RelGAN remain almost unchanged, while our method tends to strengthen the attribute of wearing eyeglasses. When $\lambda$ is set to $2$, the eyeglasses in the outputted face image of our method will be replaced by a pair of sunglasses. Both qualitative and quantitative results demonstrate that our method has a significantly stronger capability of implementing arbitrary face editing.

\subsection{Ablation Studies}
\label{sec:ablation}

In this section, we conduct experiments to validate the effectiveness of each component of the proposed HifaFace. We mainly study the following three important issues: 1) the wavelet generator and discriminator; 2) the attribute regression loss; 3) the semi-supervised learning strategy.

\noindent\textbf{Wavelet-Based Generator and Discriminator} To verify the effectiveness of our proposed wavelet-based generator and discriminator, we evaluate the performance of  several variants of our method. Specifically, we consider the following four variants: 1) our full model; 2) the proposed model without the wavelet-based skip-connection in the Generator (w/o WS in $G$); 3) the proposed model without the high-frequency Discriminator (w/o $D_H$); 4) the proposed model with the vanilla skip-connection (w/ VS in G). Qualitative and quantitative results of these methods are shown in Figure~\ref{fig:ablation} and Table~\ref{tab:ablation}, respectively, from which we can draw two conclusions as follows. 1) The wavelet-based generator and discriminator are extremely important for the overall framework, since it is impossible to obtain satisfactory face editing results without them. 2) Our generator does not encode hidden information in its output image. This is intuitively demonstrated in Figure~\ref{fig:ablation}, where we expect to add eyeglasses on the synthesized smile face images (the fifth column). We can see that only our full model is able to satisfactorily achieve this goal. The hidden information of each model is shown in the rightmost column of Figure~\ref{fig:ablation}.

\noindent\textbf{Attribute Regression Loss.} The attribute regression loss is introduced to achieve arbitrary face editing, which aims to obtain both high-quality interpolated results and extrapolated results, as shown in Figure~\ref{fig:lar_comparison}. With our proposed attribute regression loss, we can get a smooth and wider-range control for each attribute.

\noindent\textbf{Semi-Supervised Learning.}
To validate the effectiveness of our semi-supervised learning strategy, we conduct an ablation study and show the results in Figure~\ref{fig:semi_comparison}. It can be observed that benefiting from the utilization of large amounts of data, the quality of synthesis results obtained by the method with semi-supervised learning is significantly better, especially for some attributes such as ``eyeglasses'' and ``old'', where very few labeled data are available.

\begin{figure}[t]
    \captionsetup[subfigure]{aboveskip=1pt} %
    \captionsetup[subfigure]{labelformat=empty}
    \centering
    \begin{subfigure}[t]{0.1\textwidth}
        \includegraphics[width=\textwidth]{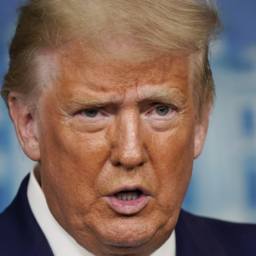}
        \vspace{-4mm}

        \includegraphics[width=\textwidth]{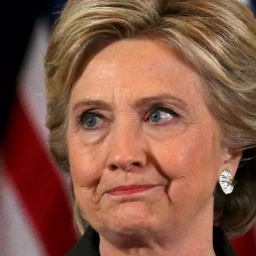}
        \caption{Input}
    \end{subfigure}\hspace{0.6mm}%
    \begin{subfigure}[t]{0.1\textwidth}
        \includegraphics[width=\textwidth]{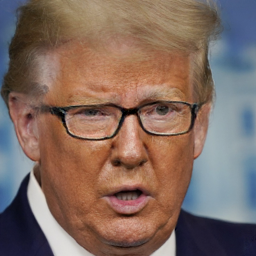}
        \vspace{-4mm}

        \includegraphics[width=\textwidth]{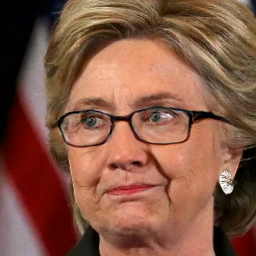}
        \caption{Eyeglasses}
    \end{subfigure}\hspace{0.6mm}%
    \begin{subfigure}[t]{0.1\textwidth}
        \includegraphics[width=\textwidth]{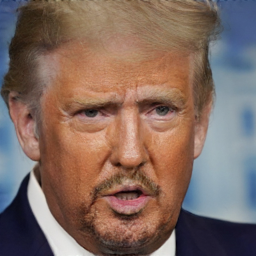}
        \vspace{-4mm}

        \includegraphics[width=\textwidth]{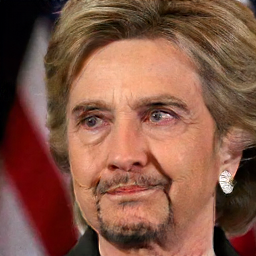}
        \caption{Beard}
    \end{subfigure}\hspace{0.6mm}%
    \begin{subfigure}[t]{0.1\textwidth}
        \includegraphics[width=\textwidth]{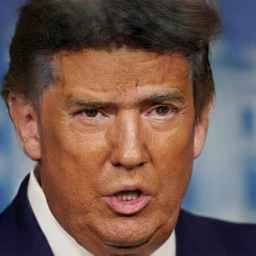}
        \vspace{-4mm}

        \includegraphics[width=\textwidth]{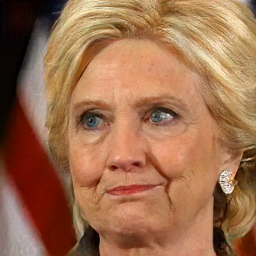}
        \caption{Hair color}
    \end{subfigure}
    \vspace{-3mm}
    \caption{Examples of face editing results for wild images obtained by our HifaFace.}
    \label{fig:editing_wild}
    \vspace{-6.5mm}
\end{figure}

\subsection{Editing Wild Faces}
Finally, we demonstrate the strong capability of HifaFace by editing wild face images downloaded from the Internet. As shown in Figure~\ref{fig:editing_wild}, our method works well for face images under various poses and expressions.
\section{Conclusion}
In this paper, we first revisit \emph{cycle consistency} in current face editing frameworks and observed an interesting phenomenon called steganography. We then propose a novel model named HifaFace to address this problem.
The key idea of our method is to adopt a wavelet-based generator and a high-frequency discriminator. Moreover, we also designed a novel attribute regression loss to achieve arbitrary face editing on a single attribute.
Extensive experiments demonstrate the superiority of our framework for high-fidelity and arbitrary face editing. Hopefully, this paper will be able to inspire researchers for solving the similar problems of cycle consistency in many other tasks.

{\small
\bibliographystyle{ieee_fullname}
\bibliography{HifaFace}
}

\clearpage
\appendix

\begin{table*}[h]
    \centering
    \begin{tabular}{lcc}
        \toprule
         Components  &  Input $\to$ Output Shape  &  Layer Information \\
         \midrule
        From RGB & (3, H, W) $\to$ (64, H, W)  & Conv(F64) \\
        \makecell[l]{Downsample \\ ResBlock} &  (64, H, W) $\to$ (128, H/2, W/2)  & IN-LReLU-Conv(F64)-Downsample-IN-LReLU-Conv(F128) \\
        \makecell[l]{Downsample \\ ResBlock} &  (128, H/2, W/2) $\to$ (256, H/4, W/4)  &  IN-LReLU-Conv(F64)-Downsample-IN-LReLU-Conv(F128) \\
        ResBlock &  (256, H/4, W/4) $\to$ (256, H/4, W/4)  & IN-LReLU-Conv(F256)-IN-LReLU-Conv(F256) \\
        ResBlock &  (256, H/4, W/4) $\to$ (256, H/4, W/4)  & IN-LReLU-Conv(F256)-IN-LReLU-Conv(F256) \\
        ResBlock &  (256, H/4, W/4) $\to$ (256, H/4, W/4)  & IN-LReLU-Conv(F256)-IN-LReLU-Conv(F256) \\
        AdaIN ResBlock &  (256, H/4, W/4) $\to$ (256, H/4, W/4)  & AdaIN-LReLU-Conv(F256)-AdaIN-LReLU-Conv(F256) \\
        ResBlock &  (256, H/4, W/4) $\to$ (256, H/4, W/4)  & IN-LReLU-Conv(F256)-IN-LReLU-Conv(F256) \\
        ResBlock &  (256, H/4, W/4) $\to$ (256, H/4, W/4)  & IN-LReLU-Conv(F256)-IN-LReLU-Conv(F256) \\
        Upsample ResBlock &  (256 $\times$ 4, H/4, W/4) $\to$ (128, H/2, W/2)  & IN-LReLU-Conv(F256)-Upsample-IN-LReLU-Conv(F128) \\
        Upsample ResBlock &  (128 $\times$ 4, H/2, W/2) $\to$ (64, H, W)  &  IN-LReLU-Conv(F64)-Upsample-IN-LReLU-Conv(F3) \\
        To RGB & (64 $\times$ 4, H, W) $\to$ (3, H, W)  & LReLU-Conv(F3) \\
        \bottomrule
    \end{tabular}
    \vspace{-2mm}
    \caption{The network architecture of the generator $G$. For all convolution (Conv) layers, the kernel size, stride and padding are $3, 1,\text{and }1$, respectively, F$x$ is the channel number of feature maps. ``IN" denotes the Instance Normalization~\cite{Ulyanov2016InstanceNT}, ``LReLU'' denotes the LeakyReLU activation function. ``AdaIN''~\cite{Huang2017ArbitraryST} is used to inject the attribute vector. Since we used the wavelet-base skip-connection in $G$, the number of input channels in decoding layers are multiplied by $4$.}
    \label{tab:sm_arch_g}
\end{table*}

\begin{table*}[h]
    \centering
    \begin{tabular}{lcc}
        \toprule
         Components  &  Input $\to$ Output Shape  &  Layer Information \\
         \midrule
         \multirow{7}{*}{$D_{I0}$} & (3, H, W) $\to$ (32, H/2, W/2)  & Conv(F32, K=4, S=2, P=1)-LReLU \\
        & (32, H/2, W/2) $\to$ (64, H/4, W/4)  & Conv(F64, K=4, S=2, P=1)-LReLU \\
        & (64, H/4, W/4) $\to$ (128, H/8, W/8)  & Conv(F128, K=4, S=2, P=1)-LReLU \\
        & (128, H/8, W/8) $\to$ (256, H/16, W/16)  & Conv(F256, K=4, S=2, P=1)-LReLU \\
        & (256, H/16, W/16) $\to$ (512, H/32, W/32)  & Conv(F512, K=4, S=2, P=1)-LReLU \\
        & (512, H/32, W/32) $\to$ (512, H/64, W/64)  & Conv(F512, K=4, S=2, P=1)-LReLU \\
        & (512, H/64, W/64) $\to$ (1, 1, 1)  & Conv(F1, K=4, S=1) \\
        \midrule
        \multirow{6}{*}{$D_{I1}$} & (3, H/2, W/2) $\to$ (32, H/4, W/4)  & Conv(F32, K=4, S=2, P=1)-LReLU \\
        & (32, H/4, W/4) $\to$ (64, H/8, W/8)  & Conv(F64, K=4, S=2, P=1)-LReLU \\
        & (64, H/8, W/8) $\to$ (128, H/16, W/16)  & Conv(F128, K=4, S=2, P=1)-LReLU \\
        & (128, H/16, W/16) $\to$ (256, H/32, W/32)  & Conv(F256, K=4, S=2, P=1)-LReLU \\
        & (256, H/32, W/32) $\to$ (512, H/64, W/64)  & Conv(F512, K=4, S=2, P=1)-LReLU \\
        & (512, H/64, W/64) $\to$ (1, 1, 1)  & Conv(F1, K=4, S=1) \\
        \bottomrule
    \end{tabular}
    \vspace{-2mm}
    \caption{The network architecture of the multi-scale image-level Discriminators: $D_{I0}$ and $D_{I1}$.}
    \label{tab:sm_arch_d}
    \vspace{-1mm}
\end{table*}

\begin{table*}[h]
    \centering
    \begin{tabular}{lcc}
        \toprule
         Components  &  Input $\to$ Output Shape  &  Layer Information \\
         \midrule
         \multirow{5}{*}{$D_{H0}$} & (3  $\times$ 3, H/2, W/2) $\to$ (32, H/4, W/4)  & Conv(F32, K=4, S=2, P=1)-LReLU \\
        & (32, H/4, W/4) $\to$ (64, H/8, W/8)  & Conv(F64, K=4, S=2, P=1)-LReLU \\
        & (64, H/8, W/8) $\to$ (128, H/16, W/16)  & Conv(F128, K=4, S=2, P=1)-LReLU \\
        & (128, H/16, W/16) $\to$ (256, H/32, W/32)  & Conv(F256, K=4, S=2, P=1)-LReLU \\
        & (256, H/32, W/32) $\to$ (512, H/64, W/64)  & Conv(F512, K=4, S=2, P=1)-LReLU \\
        & (512, H/64, W/64) $\to$ (1, 1, 1)  & Conv(F1, K=4, S=1) \\
        \midrule
        \multirow{4}{*}{$D_{H1}$} & (3  $\times$ 3, H/4, W/4) $\to$ (64, H/8, W/8)  & Conv(F64, K=4, S=2, P=1)-LReLU \\
        & (64, H/8, W/8) $\to$ (128, H/16, W/16)  & Conv(F128, K=4, S=2, P=1)-LReLU \\
        & (128, H/16, W/16) $\to$ (256, H/32, W/32)  & Conv(F256, K=4, S=2, P=1)-LReLU \\
        & (256, H/32, W/32) $\to$ (512, H/64, W/64)  & Conv(F512, K=4, S=2, P=1)-LReLU \\
        & (512, H/64, W/64) $\to$ (1, 1, 1)  & Conv(F1, K=4, S=1) \\
        \bottomrule
    \end{tabular}
    \caption{The network architecture of the multi-scale high-frequency Discriminators: $D_{H0}$ and $D_{H1}$.}
    \label{tab:sm_arch_dh}
\end{table*}

\section{Implementation Details}
\noindent\textbf{Dataset Details.} We use CelebA-HQ~\cite{Karras2018ProgressiveGO} as the labeled dataset, which contains $30,000$ images with $40$ binary attribute annotations for each image. We randomly select $28,000$ images as the training set to train the attribute classifier $C$, the remaining $2,000$ images are used as the testing set. For the unlabelled dataset FFHQ~\cite{Karras2019ASG}, we use the first $66,000$ images to train the $G$, $D_H$ and $D_I$ and the remaining $4,000$ images for testing. The image resolution is chosen as $256 \times 256$ in our experiments.

\noindent\textbf{Model Details.} The detailed architectures of the Generator, Discriminators are shown in Table~\ref{tab:sm_arch_g}, Table~\ref{tab:sm_arch_d} and Table~\ref{tab:sm_arch_dh} respectively.

\noindent\textbf{Hyper-Parameters Details.} The exponential moving average~\cite{Yazici2019TheUE} is applied to the Generator $G$. We use Adam optimizer~\cite{Kingma2015AdamAM} with $\beta_1 = 0.0$ and $\beta_2 = 0.999$, and utilize TTUR~\cite{Heusel2017GANsTB} with $lr_{G} = 5e-4, \ lr_{D_{I}} = 2e-3 $ and $ \ lr_{D_{H}} = 2e-3$. The loss weights are $\lambda_{GAN}^{I} = 1.0, \lambda_{GAN}^{H} = 1.0, \lambda_{ar} = 1.0, \lambda_{ac} = 1.0$ and $ \lambda_{cyc} = 10.0$. We train the model for $100$ epochs and another $100$ epochs training with learning rate decaying, where the decaying rate is set to $0.999$ for every $10$ epochs.

\begin{figure*}[h]
    \captionsetup[subfigure]{labelformat=empty}
    \captionsetup[subfigure]{aboveskip=1pt} %
    \centering
    \begin{subfigure}[t]{\dimexpr0.11\textwidth+9pt\relax}
        \makebox[9pt]{\raisebox{26pt}{\rotatebox[origin=c]{90}{H-Flip}}}%
        \includegraphics[width=\dimexpr\linewidth-9pt\relax]
        {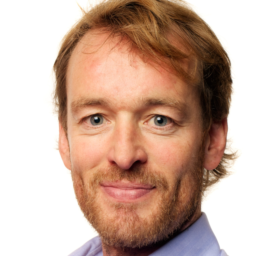}  \\
        \vspace{-3.6mm}
        \makebox[9pt]{\raisebox{26pt}{\rotatebox[origin=c]{90}{Noise}}}%
        \includegraphics[width=\dimexpr\linewidth-9pt\relax]
        {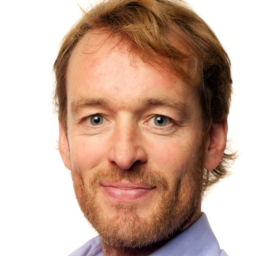}  \\
        \vspace{-3.6mm}
        \makebox[9pt]{\raisebox{26pt}{\rotatebox[origin=c]{90}{ColorJitter}}}%
        \includegraphics[width=\dimexpr\linewidth-9pt\relax]
        {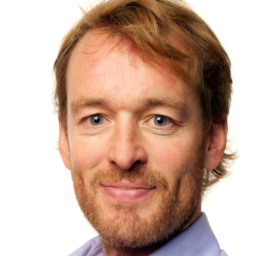}  \\
        \vspace{-3.6mm}
        \makebox[9pt]{\raisebox{26pt}{\rotatebox[origin=c]{90}{Affine}}}%
        \includegraphics[width=\dimexpr\linewidth-9pt\relax]
        {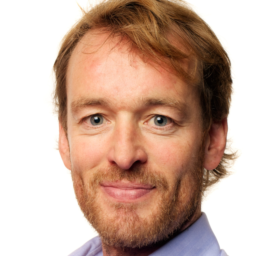}  \\
        \vspace{-3.6mm}
        \makebox[9pt]{\raisebox{26pt}{\rotatebox[origin=c]{90}{\textbf{HifaFace}}}}%
        \includegraphics[width=\dimexpr\linewidth-9pt\relax]
        {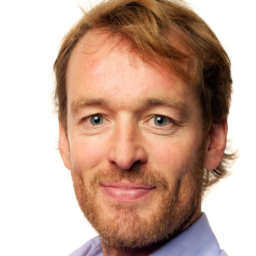}  \\
        \vspace{-3.6mm}
        \caption{\quad Input}
    \end{subfigure}\hspace{.3mm}
    \begin{subfigure}[t]{0.11\textwidth}
        \includegraphics[width=\textwidth]
        {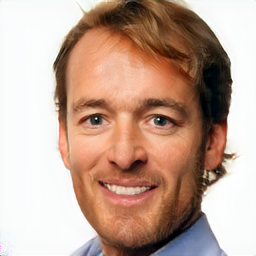}  \\
        \vspace{-3.6mm}
        \includegraphics[width=\textwidth]
        {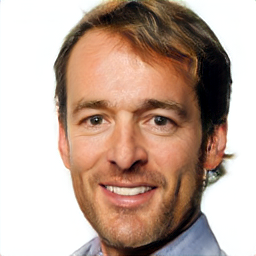}  \\
        \vspace{-3.6mm}
        \includegraphics[width=\textwidth]
        {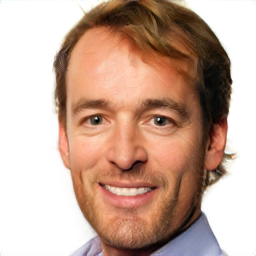}  \\
        \vspace{-3.6mm}
        \includegraphics[width=\textwidth]
        {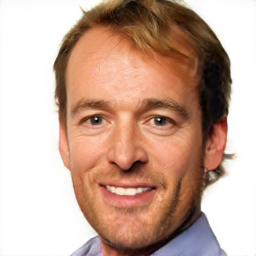}  \\
        \vspace{-3.6mm}
        \includegraphics[width=\textwidth]
        {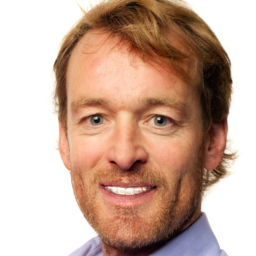}  \\
        \vspace{-3.6mm}
        \caption{+Smile}
    \end{subfigure}\hspace{.3mm}
    \begin{subfigure}[t]{0.11\textwidth}
        \includegraphics[width=\textwidth]
        {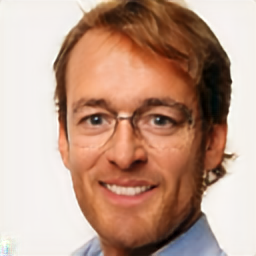}  \\
        \vspace{-3.6mm}
        \includegraphics[width=\textwidth]
        {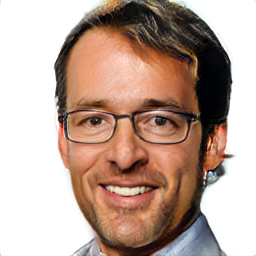}  \\
        \vspace{-3.6mm}
        \includegraphics[width=\textwidth]
        {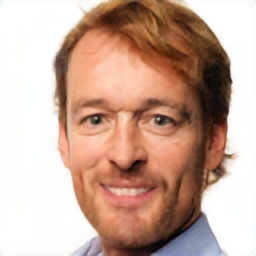}  \\
        \vspace{-3.6mm}
        \includegraphics[width=\textwidth]
        {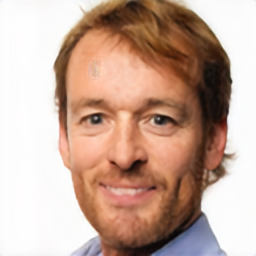}  \\
        \vspace{-3.6mm}
        \includegraphics[width=\textwidth]
        {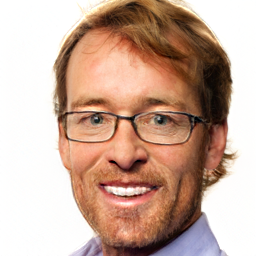}
        \caption{+Eyeglasses}
    \end{subfigure}\hspace{.3mm}
    \begin{subfigure}[t]{0.11\textwidth}
        \includegraphics[width=\textwidth]
        {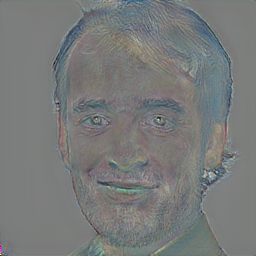}  \\
        \vspace{-3.6mm}
        \includegraphics[width=\textwidth]
        {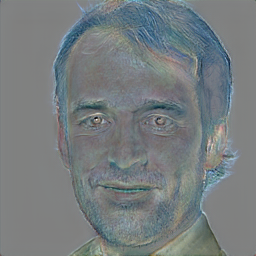}  \\
        \vspace{-3.6mm}
        \includegraphics[width=\textwidth]
        {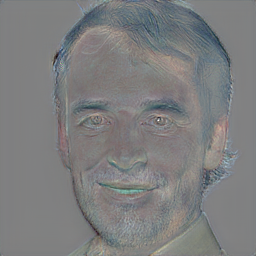}  \\
        \vspace{-3.6mm}
        \includegraphics[width=\textwidth]
        {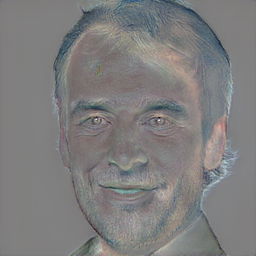}  \\
        \vspace{-3.6mm}
        \includegraphics[width=\textwidth]
        {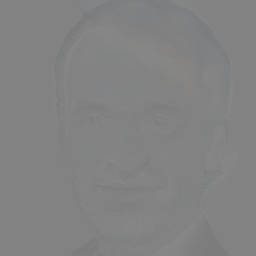}
        \caption{Hidden}
    \end{subfigure}\hspace{.3mm}
    \begin{subfigure}[t]{0.11\textwidth}
        \includegraphics[width=\textwidth]
        {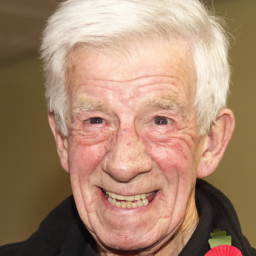}  \\
        \vspace{-3.6mm}
        \includegraphics[width=\textwidth]
        {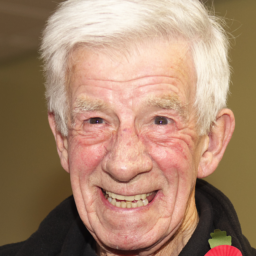}  \\
        \vspace{-3.6mm}
        \includegraphics[width=\textwidth]
        {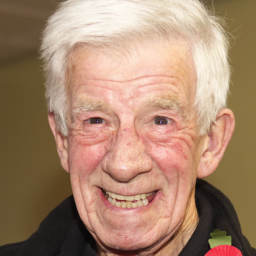}  \\
        \vspace{-3.6mm}
        \includegraphics[width=\textwidth]
        {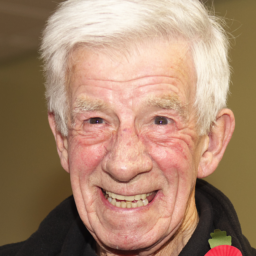}  \\
        \vspace{-3.6mm}
        \includegraphics[width=\textwidth]
        {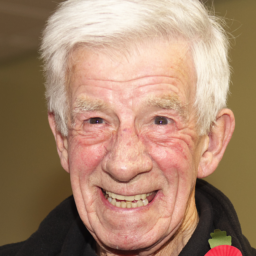}
        \caption{Input}
    \end{subfigure}\hspace{.3mm}
    \begin{subfigure}[t]{0.11\textwidth}
        \includegraphics[width=\textwidth]
        {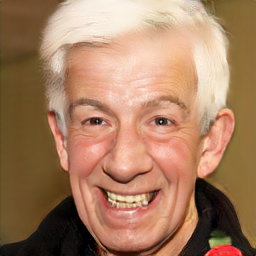}  \\
        \vspace{-3.6mm}
        \includegraphics[width=\textwidth]
        {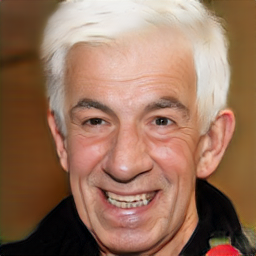}  \\
        \vspace{-3.6mm}
        \includegraphics[width=\textwidth]
        {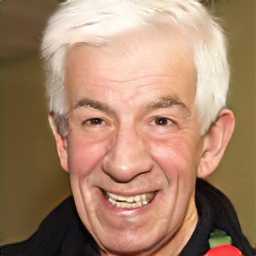}  \\
        \vspace{-3.6mm}
        \includegraphics[width=\textwidth]
        {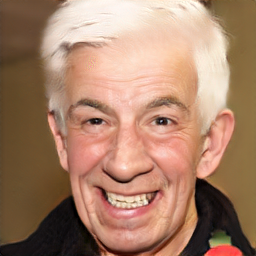}  \\
        \vspace{-3.6mm}
        \includegraphics[width=\textwidth]
        {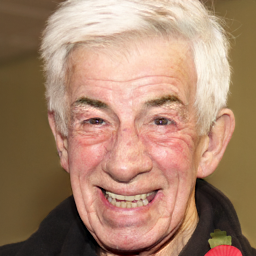}
        \caption{+Eyebrows}
    \end{subfigure}\hspace{.3mm}
    \begin{subfigure}[t]{0.11\textwidth}
        \includegraphics[width=\textwidth]
        {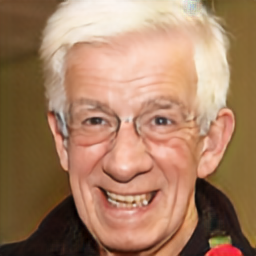}  \\
        \vspace{-3.6mm}
        \includegraphics[width=\textwidth]
        {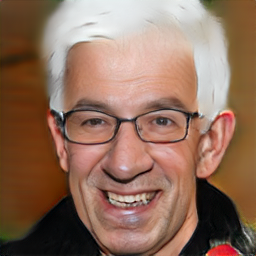}  \\
        \vspace{-3.6mm}
        \includegraphics[width=\textwidth]
        {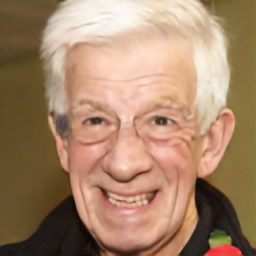}  \\
        \vspace{-3.6mm}
        \includegraphics[width=\textwidth]
        {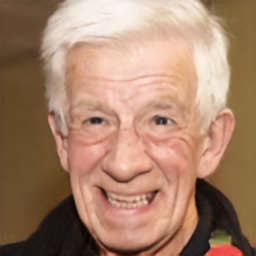}  \\
        \vspace{-3.6mm}
        \includegraphics[width=\textwidth]
        {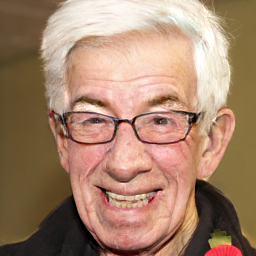}
        \caption{+Eyeglasses}
    \end{subfigure}\hspace{.3mm}
    \begin{subfigure}[t]{0.11\textwidth}
        \includegraphics[width=\textwidth]
        {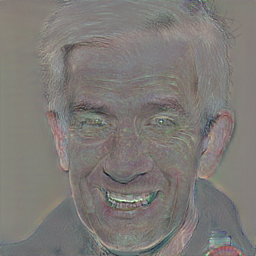}  \\
        \vspace{-3.6mm}
        \includegraphics[width=\textwidth]
        {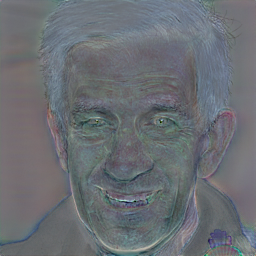}  \\
        \vspace{-3.6mm}
        \includegraphics[width=\textwidth]
        {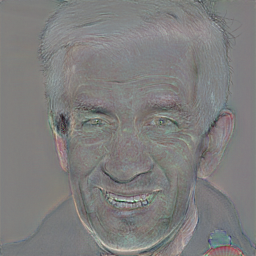}  \\
        \vspace{-3.6mm}
        \includegraphics[width=\textwidth]
        {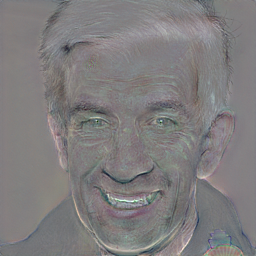}  \\
        \vspace{-3.6mm}
        \includegraphics[width=\textwidth]
        {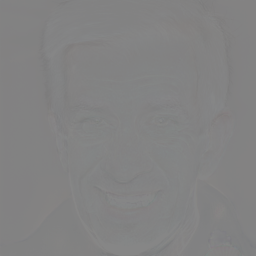}
        \caption{Hidden}
    \end{subfigure}
    \caption{Comparison of methods using different data augmentation techniques and our methods to solve the steganography problem in cycle consistency.}
    \label{fig:sm_comparison_aug}
\end{figure*}

\section{Attempts to Address the Steganography}
To alleviate the steganography problem caused by cycle consistency, we first tried a few data augmentation techniques to prevent the network from encoding hidden information to satisfy the cycle consistency. Specifically, we update the cycle consistency to
\begin{equation}
    \mathcal{L}_{cyc} = \mathbb{E} [ \parallel A(\bm{x}) - G(A(G(\bm{x}, \bm{\Delta})), -\bm{\Delta}) \parallel _1 ],
\end{equation}
where $A$ stands for data augmentation operations. Horizontal flip, random noise, color jitter (\ie, contrast, saturation, brightness) and affine transformation (\ie, rotation, translation, scaling) are investigated.

As shown in Figure~\ref{fig:sm_comparison_aug} and Table~\ref{tab:sm_comparison_aug}, even with data augmentations (\eg, horizontal flip, color jitter and affine transformation), the model can still find a way to hide the information, it still fails to synthesize rich details in the output image. Although adding noise can somehow alleviate the steganography problem, the quality of generated images is far from satisfactory, especially the rich details are missing. On the contrary, our results are high-fidelity keeping all the rich details from the input image. This validates that our proposed approach is effective to solve the steganography problem.

\begin{table}[H]
	\centering
	\begin{tabular}{lcccc}
		\toprule
             Methods          & FID $\downarrow$  & Acc. $\uparrow$   & QS $\uparrow$  & SRE $\downarrow$  \\
        \midrule
        H-flip         &   5.49       &  95.6 & 0.668    &  0.078 \\

        Noise          &   6.06       &  94.4 & 0.667    &  0.122 \\
        ColorJitter    &   5.15       &  95.9 & 0.703    &  0.059 \\
        Affine         &   5.34       &  95.8 & 0.681    &  0.071  \\
        \textbf{HifaFace}   &  \textbf{4.04}     &  \textbf{97.5}   & \textbf{0.803}  & \textbf{0.021}  \\
        \bottomrule
	\end{tabular}
	\caption{Quantitative comparison of using different data augmentation techniques and our method to solve the steganography problem in cycle consistency.}%
	\label{tab:sm_comparison_aug}
\end{table}

\begin{figure*}[h]
    \captionsetup[subfigure]{labelformat=empty}
    \captionsetup[subfigure]{aboveskip=1pt} %
    \centering
    \begin{subfigure}[t]{\dimexpr0.11\textwidth+9pt\relax}
        \makebox[9pt]{\raisebox{26pt}{\rotatebox[origin=c]{90}{Low-Freq}}}%
        \includegraphics[width=\dimexpr\linewidth-9pt\relax]
        {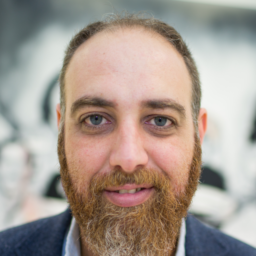}  \\
        \vspace{-3.6mm}
        \makebox[9pt]{\raisebox{26pt}{\rotatebox[origin=c]{90}{All-Freq}}}%
        \includegraphics[width=\dimexpr\linewidth-9pt\relax]
        {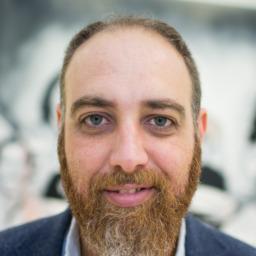}  \\
        \vspace{-3.6mm}
        \makebox[9pt]{\raisebox{26pt}{\rotatebox[origin=c]{90}{\textbf{HifaFace}}}}%
        \includegraphics[width=\dimexpr\linewidth-9pt\relax]
        {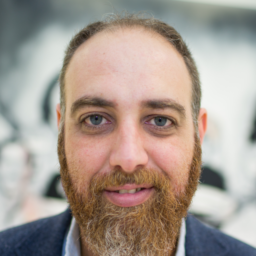}
        \caption{\quad Input}
    \end{subfigure}\hspace{.3mm}
    \begin{subfigure}[t]{0.11\textwidth}
        \includegraphics[width=\textwidth]
        {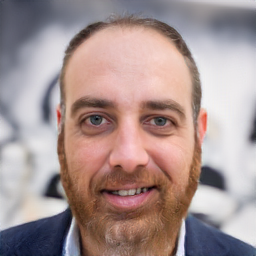}  \\
        \vspace{-3.6mm}
        \includegraphics[width=\textwidth]
        {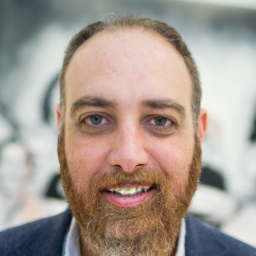}  \\
        \vspace{-3.6mm}
        \includegraphics[width=\textwidth]
        {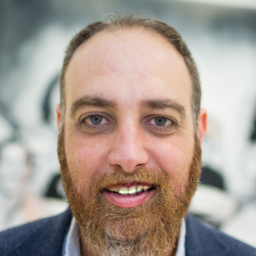}
        \caption{+ Open mouth}
    \end{subfigure}\hspace{.3mm}
    \begin{subfigure}[t]{0.11\textwidth}
        \includegraphics[width=\textwidth]
        {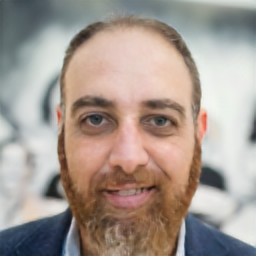}  \\
        \vspace{-3.6mm}
        \includegraphics[width=\textwidth]
        {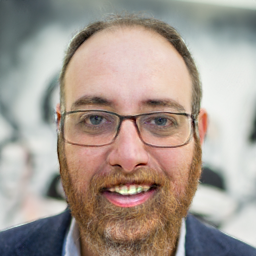}  \\
        \vspace{-3.6mm}
        \includegraphics[width=\textwidth]
        {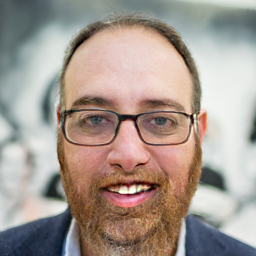}
        \caption{+Eyeglasses}
    \end{subfigure}\hspace{.3mm}
    \begin{subfigure}[t]{0.11\textwidth}
        \includegraphics[width=\textwidth]
        {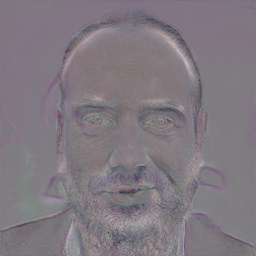}  \\
        \vspace{-3.6mm}
        \includegraphics[width=\textwidth]
        {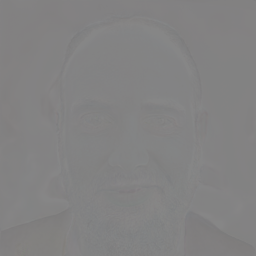}  \\
        \vspace{-3.6mm}
        \includegraphics[width=\textwidth]
        {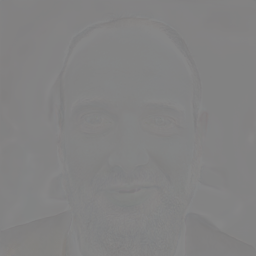}
        \caption{Hidden}
    \end{subfigure}\hspace{.3mm}
    \begin{subfigure}[t]{0.11\textwidth}
        \includegraphics[width=\textwidth]
        {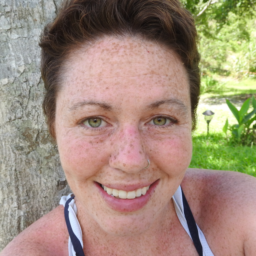}  \\
        \vspace{-3.6mm}
        \includegraphics[width=\textwidth]
        {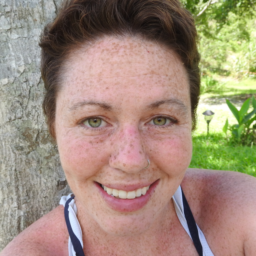}  \\
        \vspace{-3.6mm}
        \includegraphics[width=\textwidth]
        {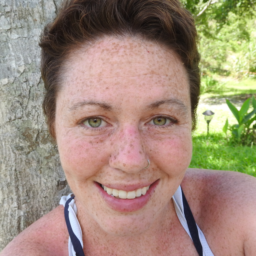}
        \caption{Input}
    \end{subfigure}\hspace{.3mm}
    \begin{subfigure}[t]{0.11\textwidth}
        \includegraphics[width=\textwidth]
        {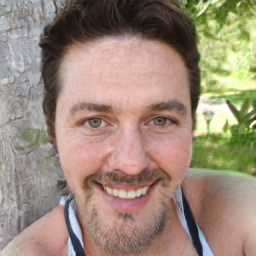}  \\
        \vspace{-3.6mm}
        \includegraphics[width=\textwidth]
        {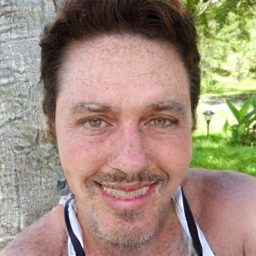}  \\
        \vspace{-3.6mm}
        \includegraphics[width=\textwidth]
        {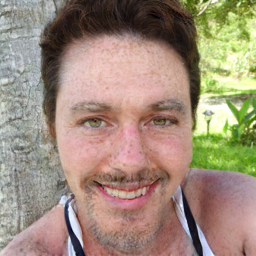}
        \caption{+Mustache}
    \end{subfigure}\hspace{.3mm}
    \begin{subfigure}[t]{0.11\textwidth}
        \includegraphics[width=\textwidth]
        {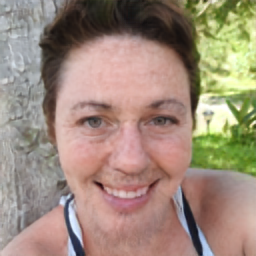}  \\
        \vspace{-3.6mm}
        \includegraphics[width=\textwidth]
        {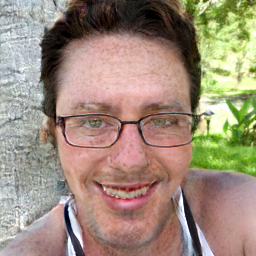}  \\
        \vspace{-3.6mm}
        \includegraphics[width=\textwidth]
        {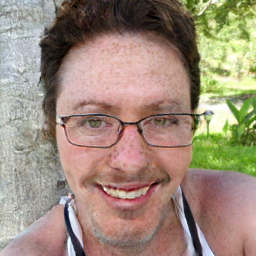}
        \caption{+Eyeglasses}
    \end{subfigure}\hspace{.3mm}
    \begin{subfigure}[t]{0.11\textwidth}
        \includegraphics[width=\textwidth]
        {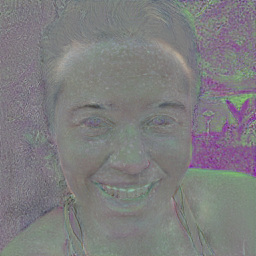}  \\
        \vspace{-3.6mm}
        \includegraphics[width=\textwidth]
        {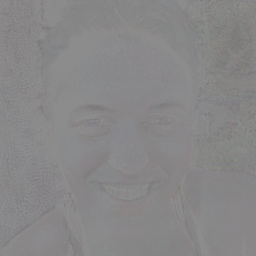}  \\
        \vspace{-3.6mm}
        \includegraphics[width=\textwidth]
        {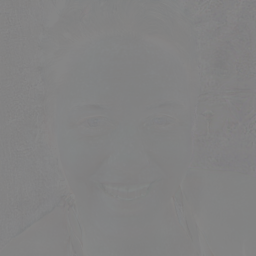}
        \caption{Hidden}
    \end{subfigure}
    \vspace{-3mm}
    \caption{Comparison of methods with different combinations of frequency components in the wavelet-based skip-connection.}
    \label{fig:sm_comparison_skip}
    \vspace{-3mm}
\end{figure*}

\section{Ablation Studies for the Generator}
To validate that the combination of \textbf{LH}, \textbf{HL} and \textbf{HH} frequency components are essential for the \emph{wavelet-based skip-connection}, we perform a few variants of different combinations of frequency components in \emph{wavelet-based skip-connection}. For concreteness, we qualitatively and quantitatively compare the following three variants with different choices of frequency components in the \emph{wavelet-based skip-connection}, we have three variants: (1) the \textbf{HifaFace}, skip-connecting \textbf{LH}, \textbf{HL} and \textbf{HH}; (2) the Low-Freq, which use the low-frequency \textbf{LL} in the skip-connection; (3) the All-Freq, skip-connecting all the four frequency components \textbf{LL}, \textbf{LH}, \textbf{HL} and \textbf{HH}. As shown in Figure~\ref{fig:sm_comparison_skip} and Table~\ref{tab:sm_comparison_skip}, we observe that the model can not synthesize rich details well without explicitly knowing high-frequency domain information. And if we skip-connecting all the low and high-frequency information, the model can produce rich details. The overall performance is slightly worse than our proposed HifaFace.

\begin{table}[H]
	\centering
	\begin{tabular}{lcccc}
		\toprule
             Methods       &  FID $\downarrow$  & Acc. $\uparrow$   & QS $\uparrow$  & SRE $\downarrow$  \\
        \midrule
        Low-Freq            &  5.37        & 95.9 &  0.707    & 0.060 \\
        All-Freq            &  4.18        & 97.4 &  0.792    & 0.022 \\
        \textbf{HifaFace}  &  \textbf{4.04}     &  \textbf{97.5}   & \textbf{0.803}  & \textbf{0.021}  \\
        \bottomrule
	\end{tabular}
	\caption{Quantitative comparison of results of using different frequency components in \emph{wavelet-based skip-connection}.}
	\label{tab:sm_comparison_skip}
\end{table}

\begin{figure*}[h]
    \captionsetup[subfigure]{labelformat=empty}
    \captionsetup[subfigure]{aboveskip=1pt} %
    \centering
    \begin{subfigure}[t]{\dimexpr0.1\textwidth+9pt\relax}
        \makebox[9pt]{\raisebox{23pt}{\rotatebox[origin=c]{90}{GANimation}}}%
        \includegraphics[width=\dimexpr\linewidth-9pt\relax]
        {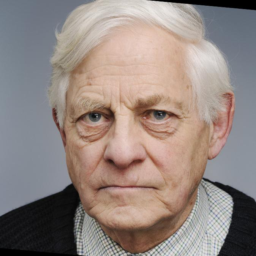}  \\
        \vspace{-3.6mm}
        \makebox[9pt]{\raisebox{24pt}{\rotatebox[origin=c]{90}{STGAN}}}%
        \includegraphics[width=\dimexpr\linewidth-9pt\relax]
        {images_sm/sota_65_in}  \\
        \vspace{-3.6mm}
        \makebox[9pt]{\raisebox{24pt}{\rotatebox[origin=c]{90}{RelGAN}}}%
        \includegraphics[width=\dimexpr\linewidth-9pt\relax]
        {images_sm/sota_65_in}  \\
        \vspace{-3.6mm}
        \makebox[9pt]{\raisebox{24pt}{\rotatebox[origin=c]{90}{IFGAN}}}%
        \includegraphics[width=\dimexpr\linewidth-9pt\relax]
        {images_sm/sota_65_in}  \\
        \vspace{-3.6mm}
        \makebox[9pt]{\raisebox{24pt}{\rotatebox[origin=c]{90}{StyleFlow}}}%
        \includegraphics[width=\dimexpr\linewidth-9pt\relax]
        {images_sm/sota_65_in}  \\
        \vspace{-3.6mm}
        \makebox[9pt]{\raisebox{24pt}{\rotatebox[origin=c]{90}{FaceApp}}}%
        \includegraphics[width=\dimexpr\linewidth-9pt\relax]
        {images_sm/sota_65_in}  \\
        \vspace{-3.6mm}
        \makebox[9pt]{\raisebox{24pt}{\rotatebox[origin=c]{90}{\textbf{HifaFace}}}}%
        \includegraphics[width=\dimexpr\linewidth-9pt\relax]
        {images_sm/sota_65_in}
        \caption{\quad Input}
    \end{subfigure}\hspace{0.3mm}
    \begin{subfigure}[t]{0.1\textwidth}
        \includegraphics[width=\textwidth]
        {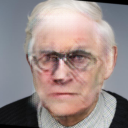}  \\
        \vspace{-3.6mm}
        \includegraphics[width=\textwidth]
        {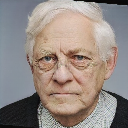}  \\
        \vspace{-3.6mm}
        \includegraphics[width=\textwidth]
        {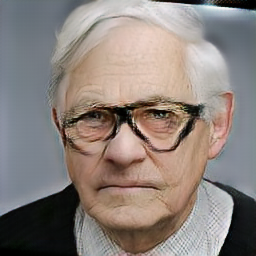}  \\
        \vspace{-3.6mm}
        \includegraphics[width=\textwidth]
        {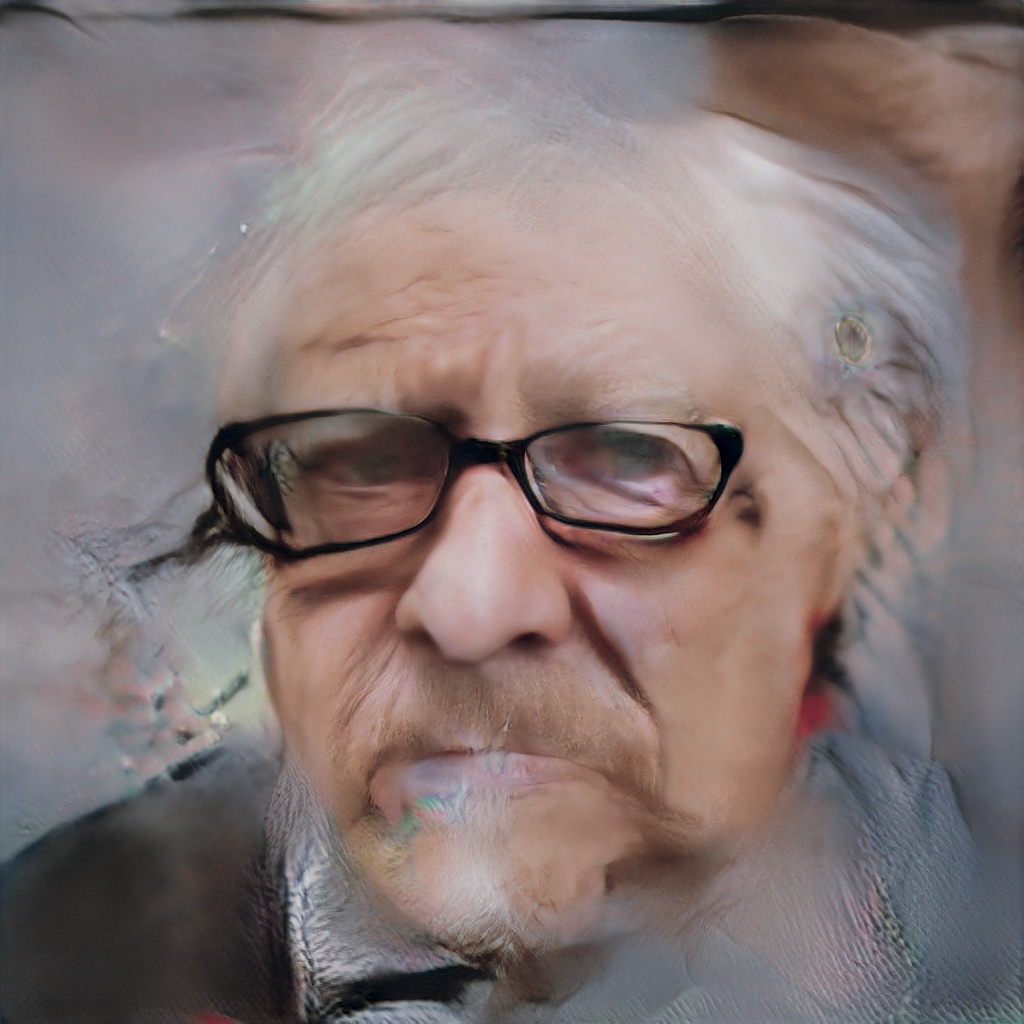}  \\
        \vspace{-3.6mm}
        \includegraphics[width=\textwidth]
        {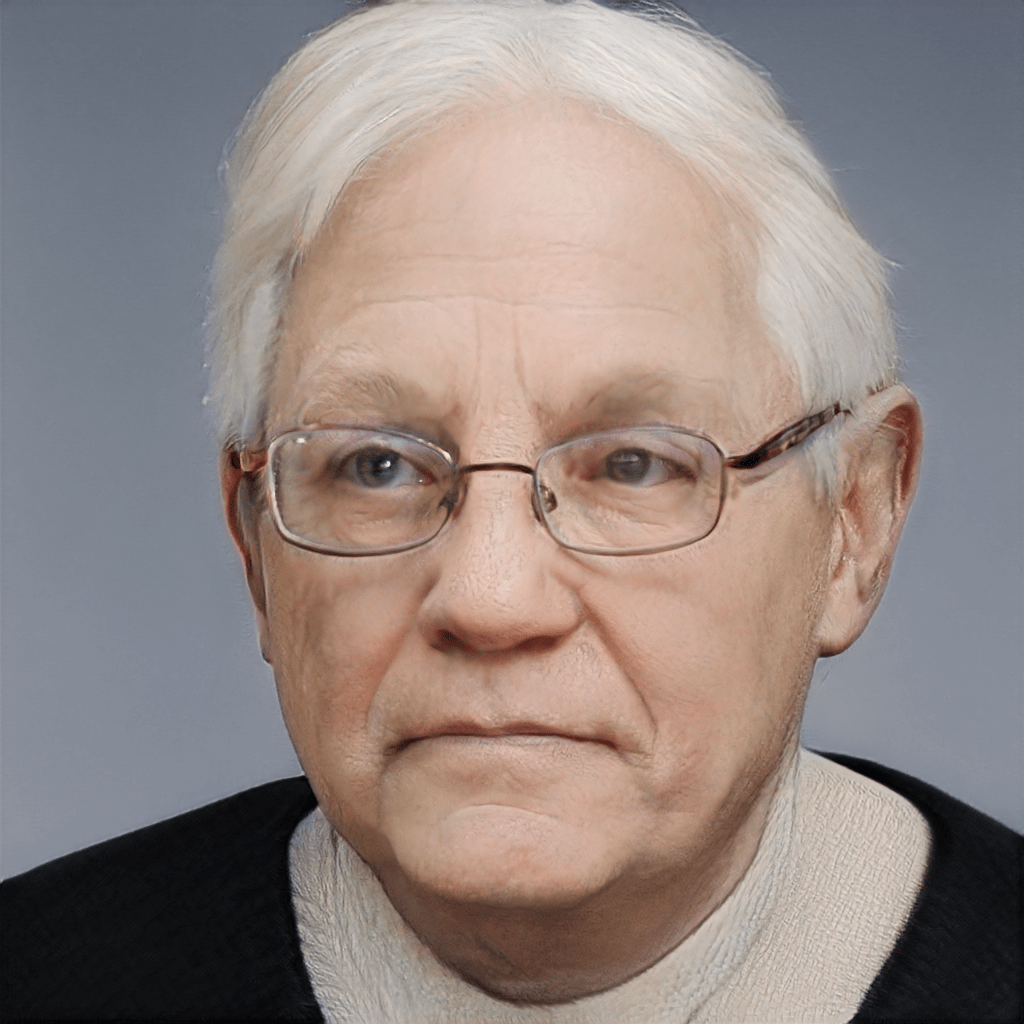}  \\
        \vspace{-3.6mm}
        \includegraphics[width=\textwidth]
        {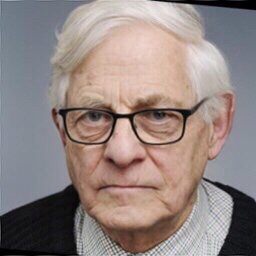}  \\
        \vspace{-3.6mm}
        \includegraphics[width=\textwidth]
        {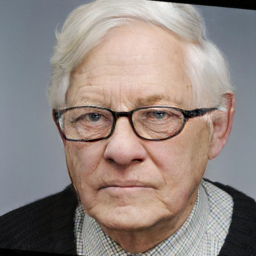}
        \caption{Eyeglasses}
    \end{subfigure}\hspace{0.3mm}
    \begin{subfigure}[t]{0.1\textwidth}
        \includegraphics[width=\textwidth]
        {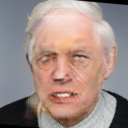}  \\
        \vspace{-3.6mm}
        \includegraphics[width=\textwidth]
        {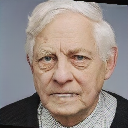}  \\
        \vspace{-3.6mm}
        \includegraphics[width=\textwidth]
        {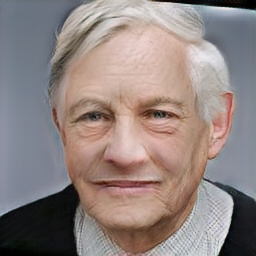}  \\
        \vspace{-3.6mm}
        \includegraphics[width=\textwidth]
        {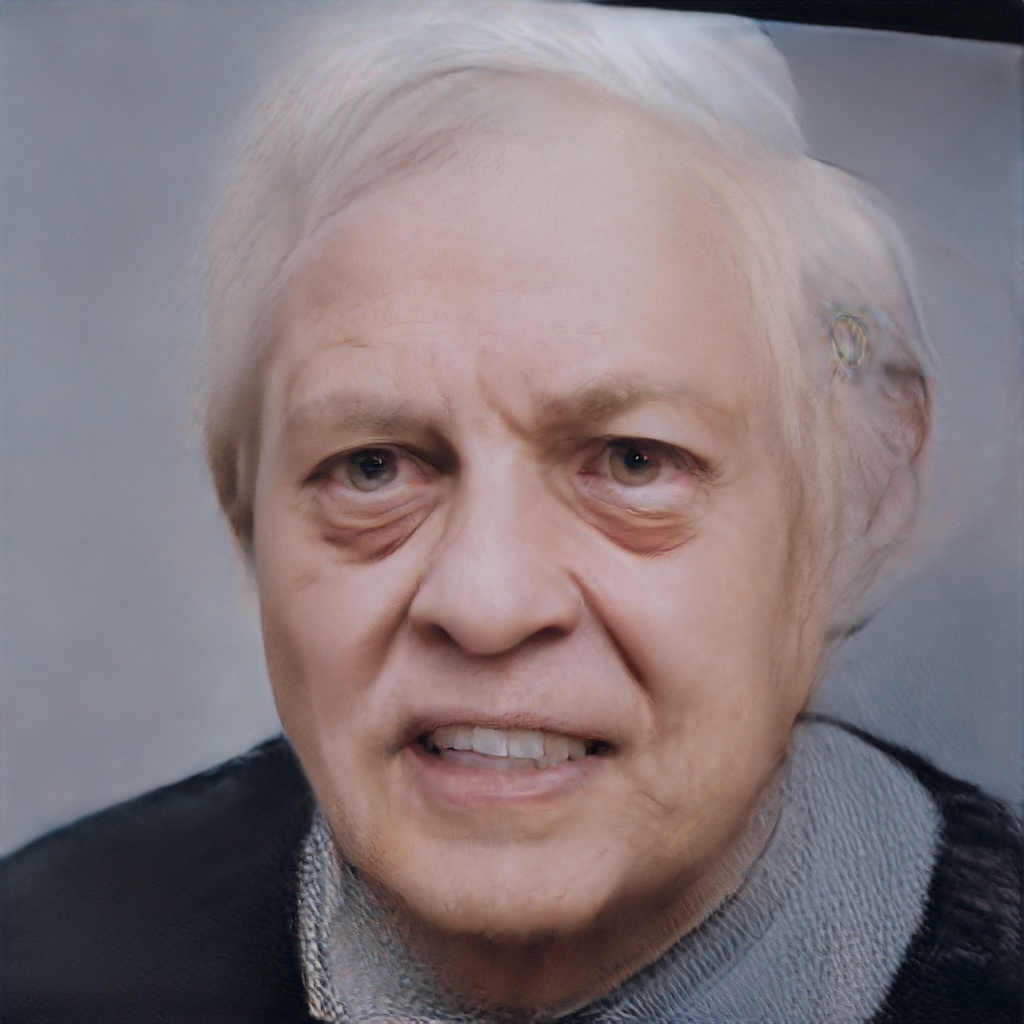}  \\
        \vspace{-3.6mm}
        \includegraphics[width=\textwidth]
        {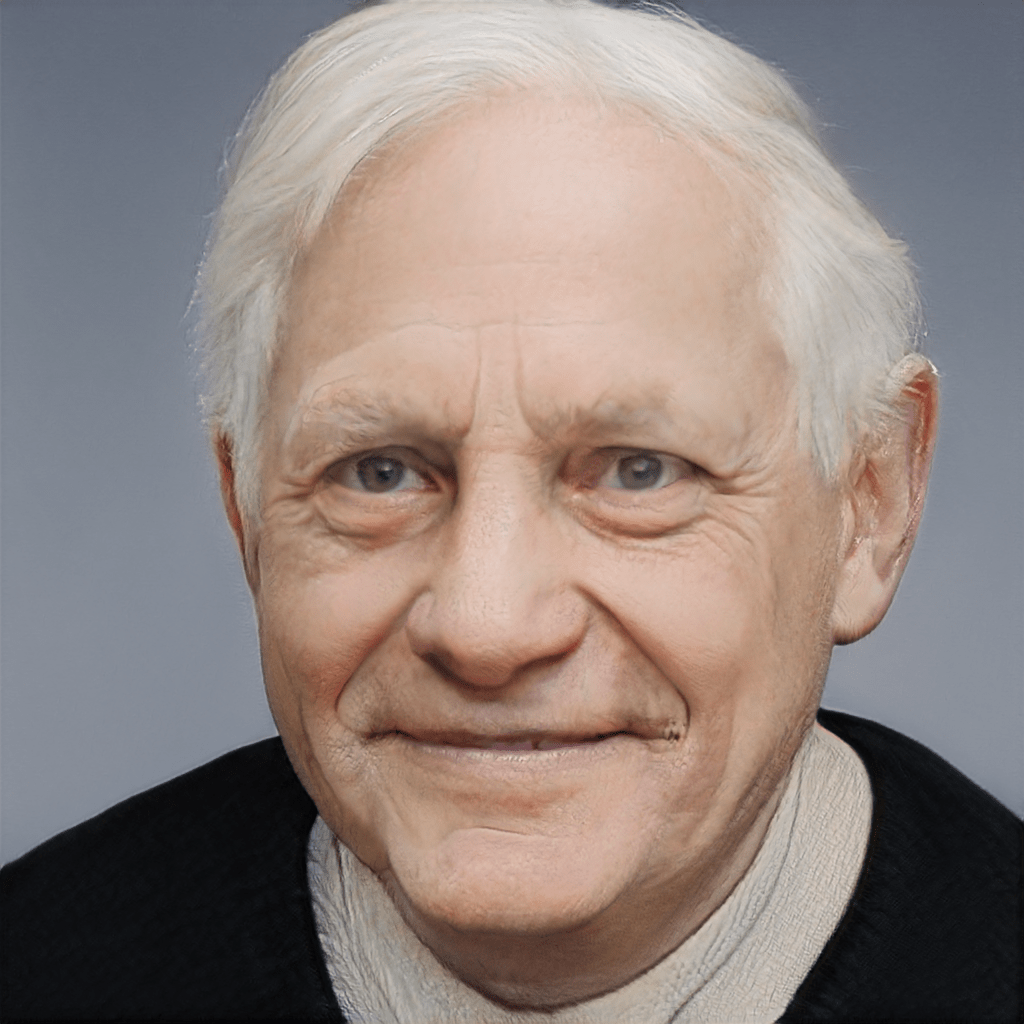}  \\
        \vspace{-3.6mm}
        \includegraphics[width=\textwidth]
        {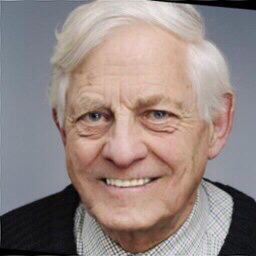}  \\
        \vspace{-3.6mm}
        \includegraphics[width=\textwidth]
        {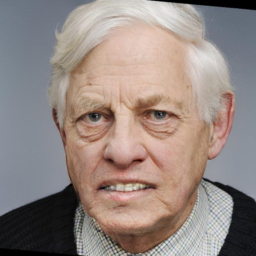}
        \caption{Expression}
    \end{subfigure}\hspace{0.3mm}
    \begin{subfigure}[t]{0.1\textwidth}
        \includegraphics[width=\textwidth]
        {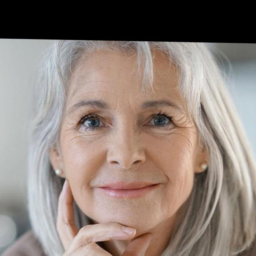}  \\
        \vspace{-3.6mm}
        \includegraphics[width=\textwidth]
        {images_sm/sota_72_in}  \\
        \vspace{-3.6mm}
        \includegraphics[width=\textwidth]
        {images_sm/sota_72_in}  \\
        \vspace{-3.6mm}
        \includegraphics[width=\textwidth]
        {images_sm/sota_72_in}  \\
        \vspace{-3.6mm}
        \includegraphics[width=\textwidth]
        {images_sm/sota_72_in}  \\
        \vspace{-3.6mm}
        \includegraphics[width=\textwidth]
        {images_sm/sota_72_in}  \\
        \vspace{-3.6mm}
        \includegraphics[width=\textwidth]
        {images_sm/sota_72_in}
        \caption{Input}
    \end{subfigure}\hspace{0.3mm}
    \begin{subfigure}[t]{0.1\textwidth}
        \includegraphics[width=\textwidth]
        {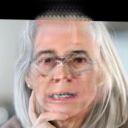}  \\
        \vspace{-3.6mm}
        \includegraphics[width=\textwidth]
        {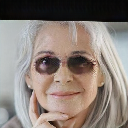}  \\
        \vspace{-3.6mm}
        \includegraphics[width=\textwidth]
        {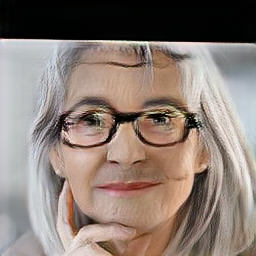}  \\
        \vspace{-3.6mm}
        \includegraphics[width=\textwidth]
        {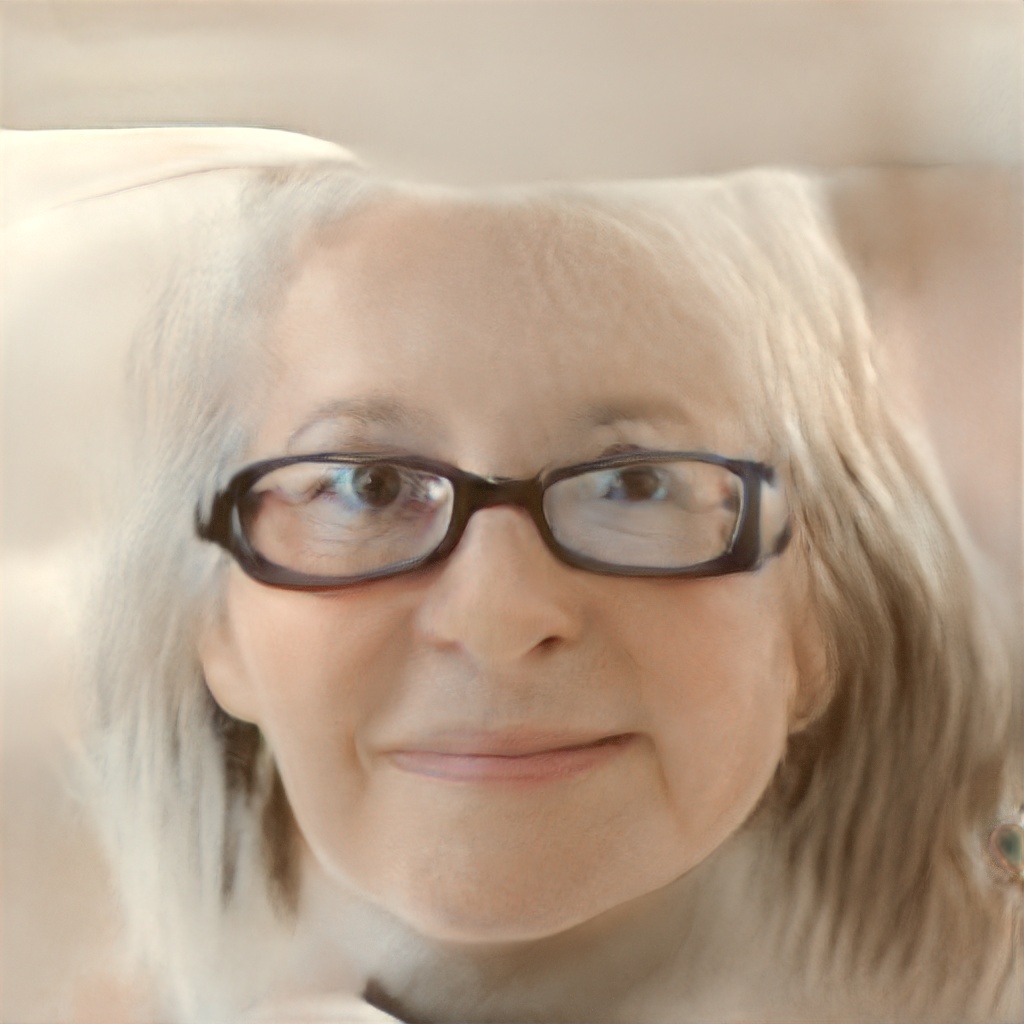}  \\
        \vspace{-3.6mm}
        \includegraphics[width=\textwidth]
        {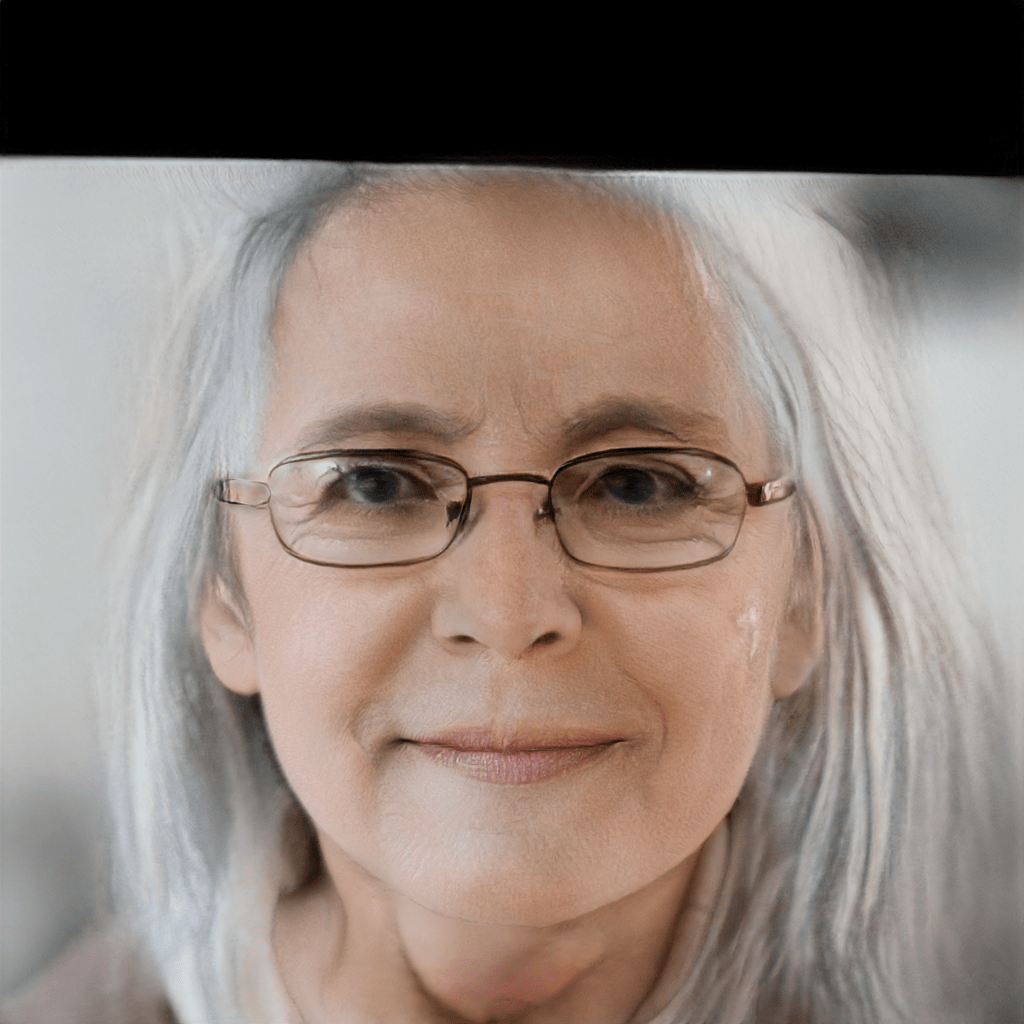}  \\
        \vspace{-3.6mm}
        \includegraphics[width=\textwidth]
        {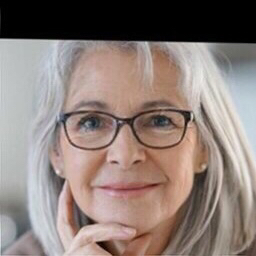}  \\
        \vspace{-3.6mm}
        \includegraphics[width=\textwidth]
        {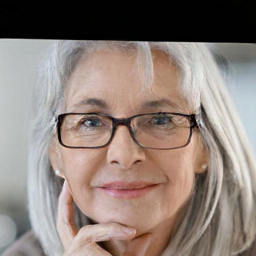}
        \caption{Eyeglasses}
    \end{subfigure}\hspace{0.3mm}
    \begin{subfigure}[t]{0.1\textwidth}
        \includegraphics[width=\textwidth]
        {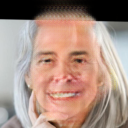}  \\
        \vspace{-3.6mm}
        \includegraphics[width=\textwidth]
        {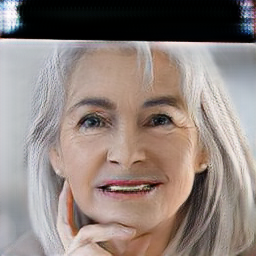}  \\
        \vspace{-3.6mm}
        \includegraphics[width=\textwidth]
        {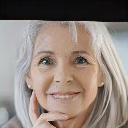}  \\
        \vspace{-3.6mm}
        \includegraphics[width=\textwidth]
        {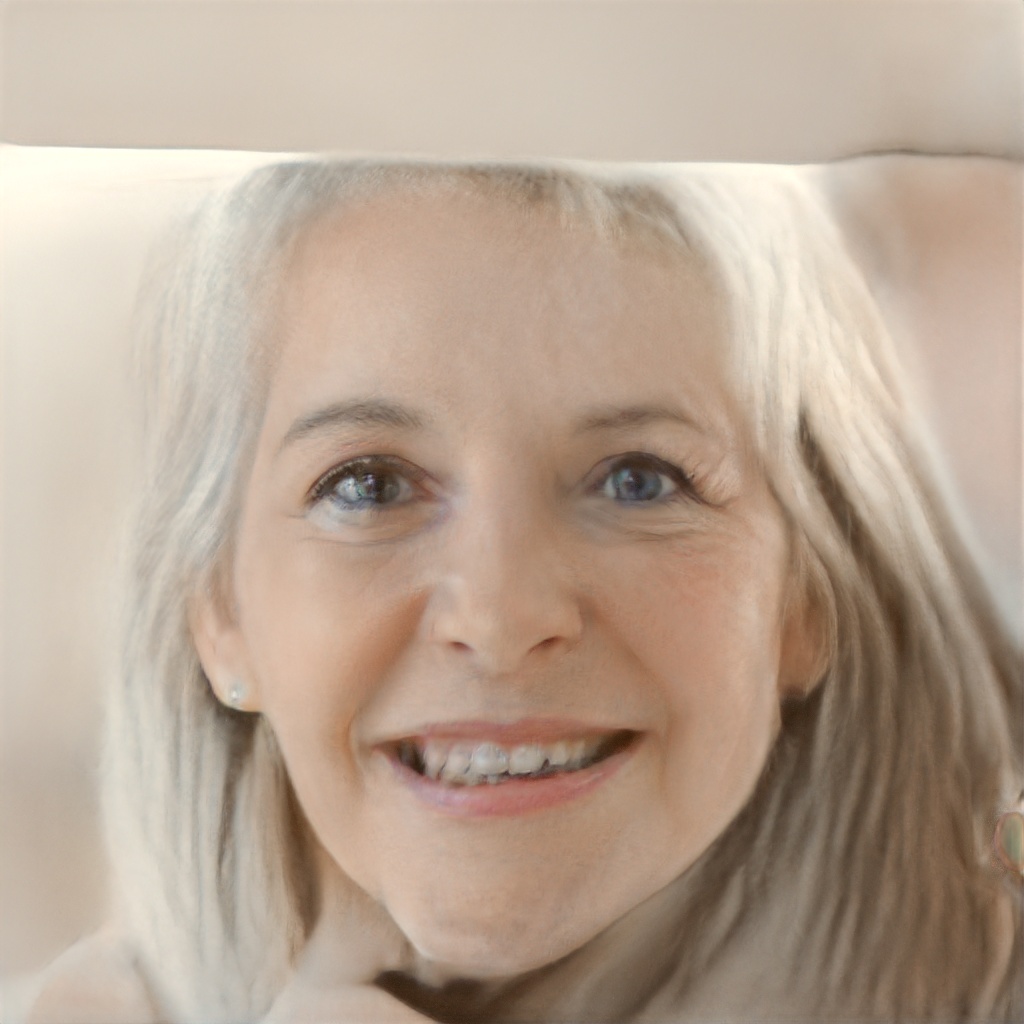}  \\
        \vspace{-3.6mm}
        \includegraphics[width=\textwidth]
        {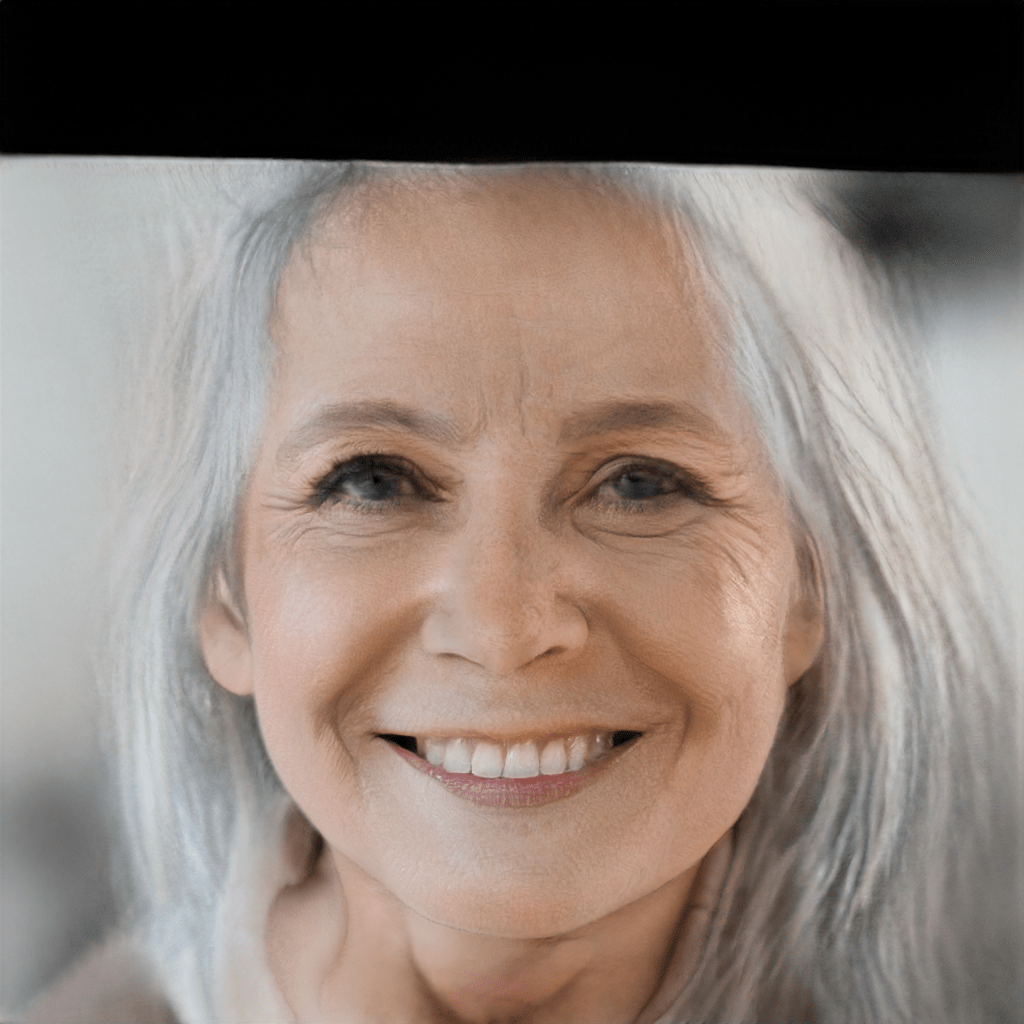}  \\
        \vspace{-3.6mm}
        \includegraphics[width=\textwidth]
        {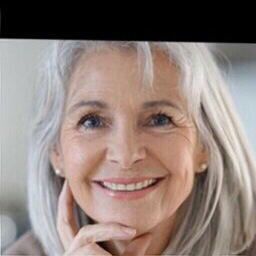}  \\
        \vspace{-3.6mm}
        \includegraphics[width=\textwidth]
        {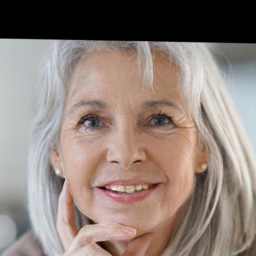}
        \caption{Expression}
    \end{subfigure}\hspace{0.3mm}
    \begin{subfigure}[t]{0.1\textwidth}
        \includegraphics[width=\textwidth]
        {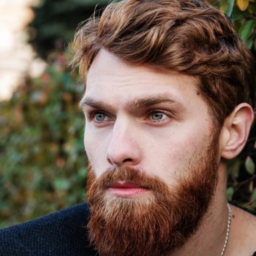}  \\
        \vspace{-3.6mm}
        \includegraphics[width=\textwidth]
        {images_sm/sota_37_in}  \\
        \vspace{-3.6mm}
        \includegraphics[width=\textwidth]
        {images_sm/sota_37_in}  \\
        \vspace{-3.6mm}
        \includegraphics[width=\textwidth]
        {images_sm/sota_37_in}  \\
        \vspace{-3.6mm}
        \includegraphics[width=\textwidth]
        {images_sm/sota_37_in}  \\
        \vspace{-3.6mm}
        \includegraphics[width=\textwidth]
        {images_sm/sota_37_in}  \\
        \vspace{-3.6mm}
        \includegraphics[width=\textwidth]
        {images_sm/sota_37_in}
        \caption{Input}
    \end{subfigure}\hspace{0.3mm}
    \begin{subfigure}[t]{0.1\textwidth}
        \includegraphics[width=\textwidth]
        {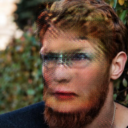}  \\
        \vspace{-3.6mm}
        \includegraphics[width=\textwidth]
        {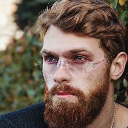}  \\
        \vspace{-3.6mm}
        \includegraphics[width=\textwidth]
        {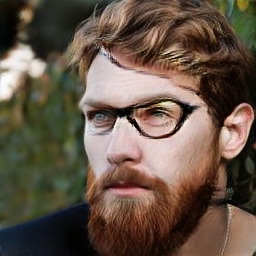}  \\
        \vspace{-3.6mm}
        \includegraphics[width=\textwidth]
        {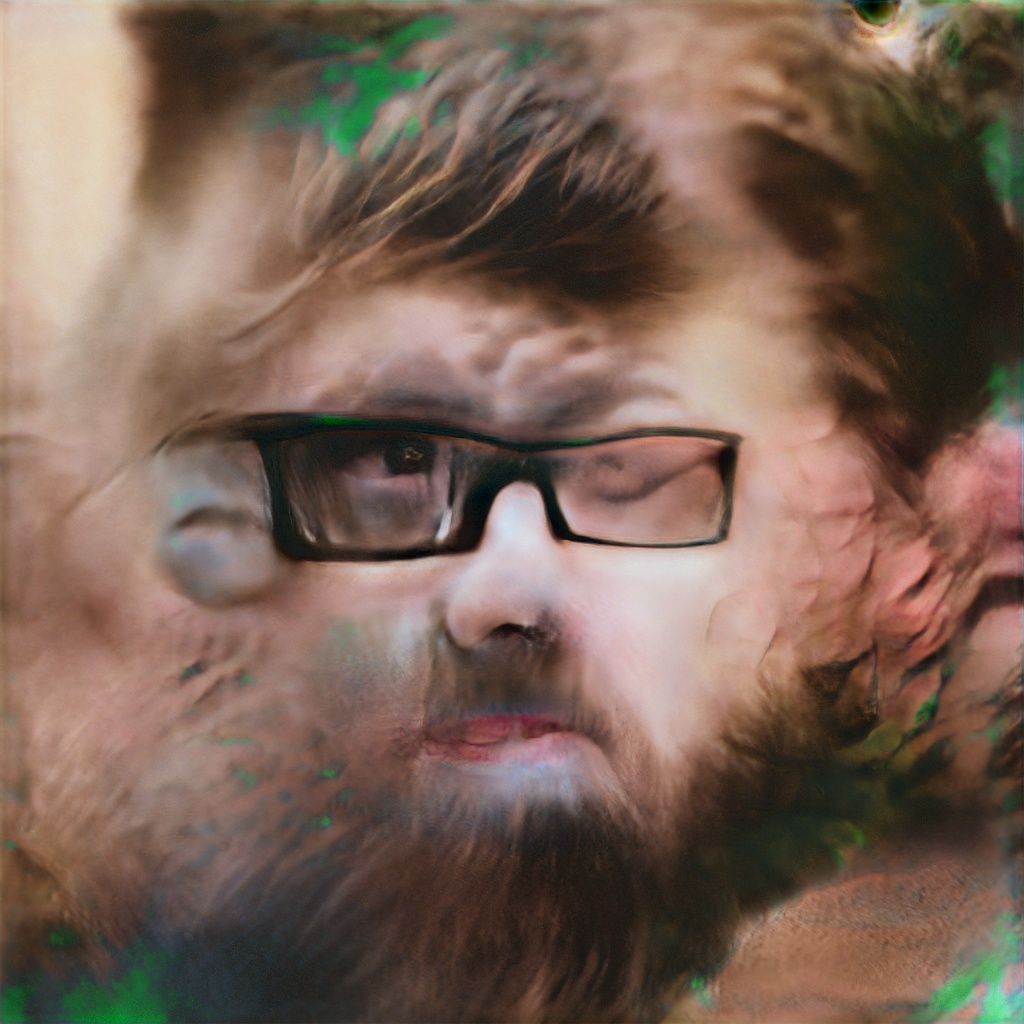}  \\
        \vspace{-3.6mm}
        \includegraphics[width=\textwidth]
        {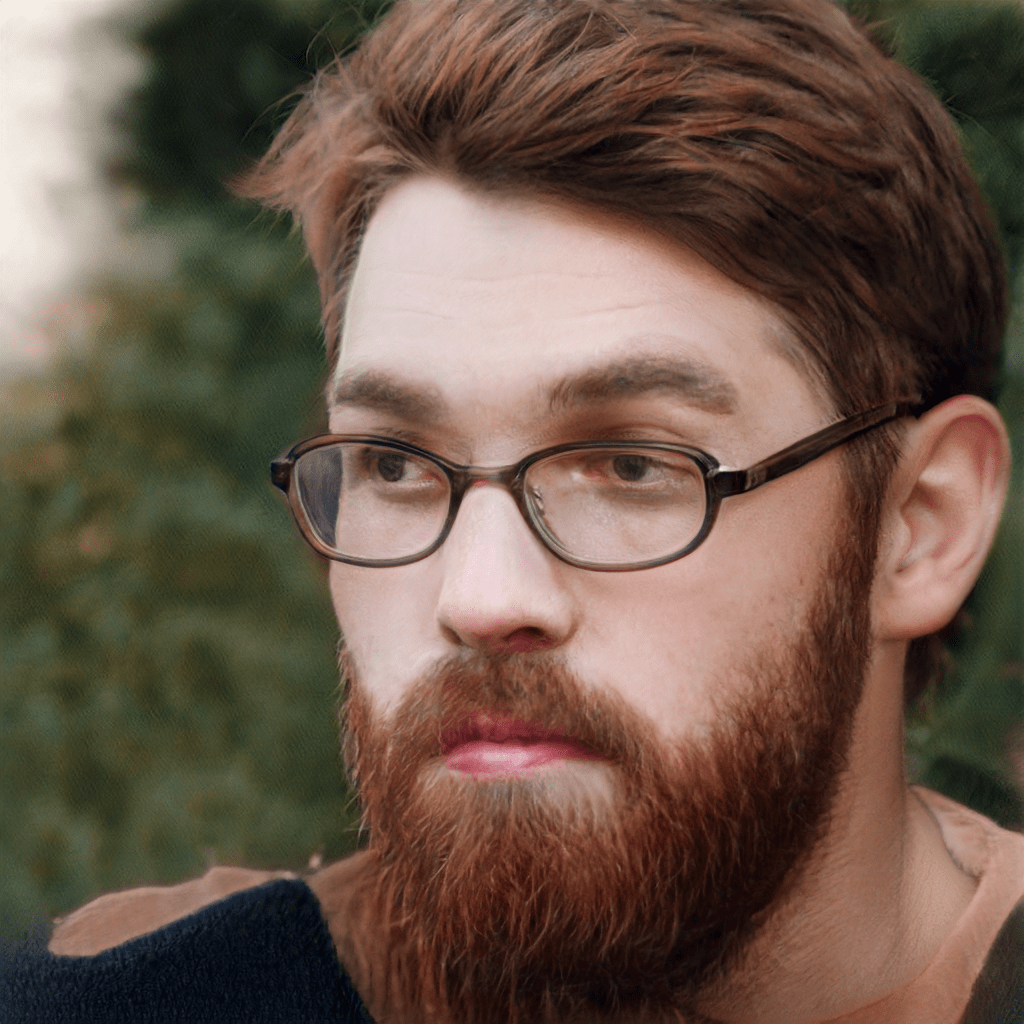}  \\
        \vspace{-3.6mm}
        \includegraphics[width=\textwidth]
        {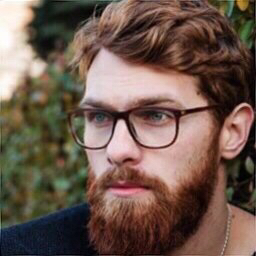}  \\
        \vspace{-3.6mm}
        \includegraphics[width=\textwidth]
        {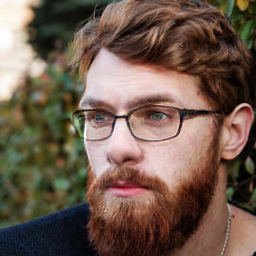}
        \caption{Eyeglasses}
    \end{subfigure}\hspace{0.3mm}
    \begin{subfigure}[t]{0.1\textwidth}
        \includegraphics[width=\textwidth]
        {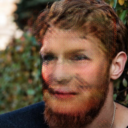}  \\
        \vspace{-3.6mm}
        \includegraphics[width=\textwidth]
        {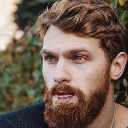}  \\
        \vspace{-3.6mm}
        \includegraphics[width=\textwidth]
        {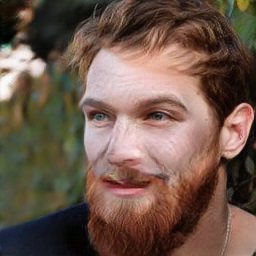}  \\
        \vspace{-3.6mm}
        \includegraphics[width=\textwidth]
        {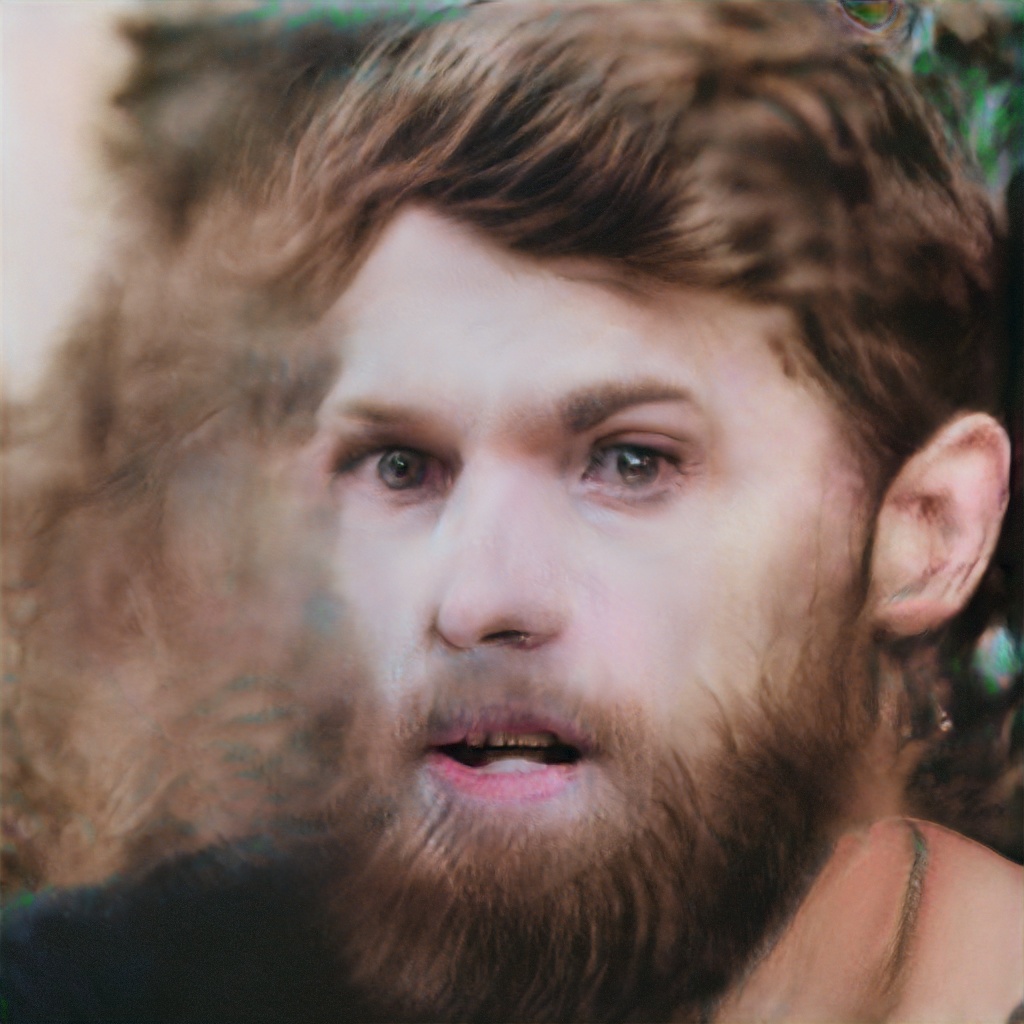}  \\
        \vspace{-3.6mm}
        \includegraphics[width=\textwidth]
        {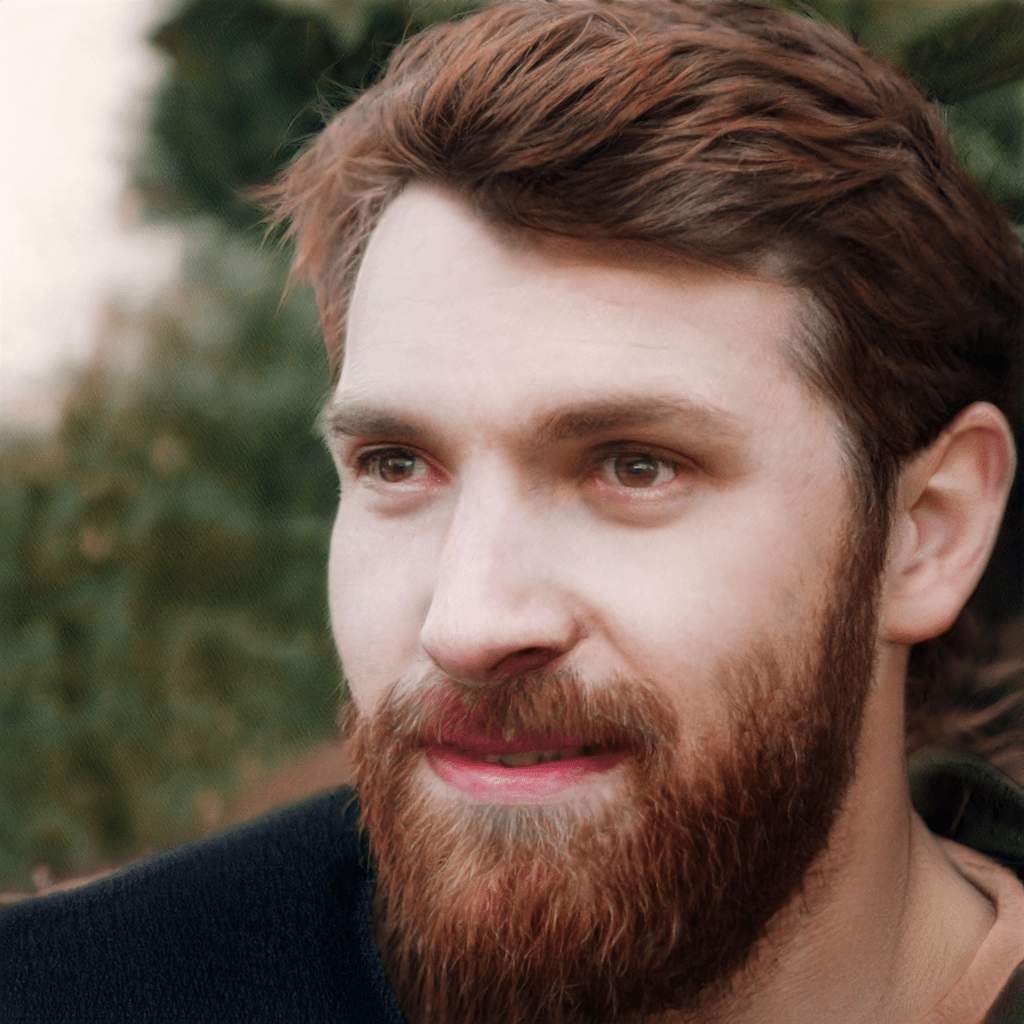}  \\
        \vspace{-3.6mm}
        \includegraphics[width=\textwidth]
        {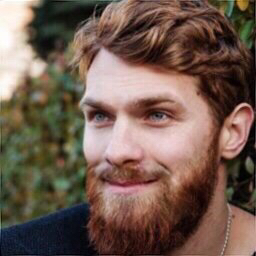}  \\
        \vspace{-3.6mm}
        \includegraphics[width=\textwidth]
        {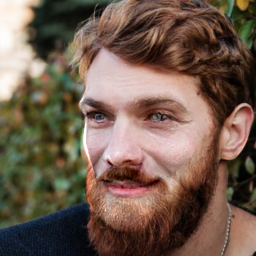}
        \caption{Expression}
    \end{subfigure}\hspace{0.3mm}
    \vspace{-3mm}
    \caption{Comparison of results obtained by our HifaFace and other state-of-the-art methods.}
    \label{fig:sm_comparison_sota}
    \vspace{-3mm}
\end{figure*}

\begin{figure*}[h]
    \captionsetup[subfigure]{labelformat=empty}
    \captionsetup[subfigure]{aboveskip=1pt} %
    \centering
    \begin{subfigure}[t]{\dimexpr0.092\textwidth+11pt\relax}
        \makebox[11pt]{\raisebox{18pt}{\rotatebox[origin=c]{90}{RelGAN}}}%
        \includegraphics[width=\dimexpr\linewidth-11pt\relax]{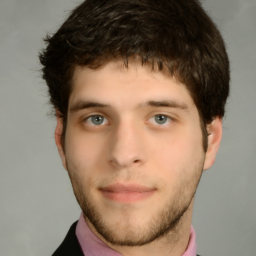}
        \vspace{-3.8mm}

        \makebox[11pt]{\raisebox{18pt}{\rotatebox[origin=c]{90}{IFGAN}}}%
        \includegraphics[width=\dimexpr\linewidth-11pt\relax]{images_sm/interp_wo_1403_in}
        \vspace{-3.8mm}

        \makebox[11pt]{\raisebox{18pt}{\rotatebox[origin=c]{90}{w/o $\mathcal{L}_{ar}$}}}%
        \includegraphics[width=\dimexpr\linewidth-11pt\relax]{images_sm/interp_wo_1403_in}
        \vspace{-3.8mm}

        \makebox[11pt]{\raisebox{18pt}{\rotatebox[origin=c]{90}{\textbf{HifaFace}}}}%
        \includegraphics[width=\dimexpr\linewidth-11pt\relax]{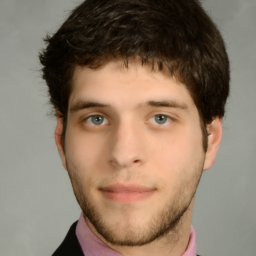}
        \caption{\quad Input}
    \end{subfigure}\hspace{0.4mm}
    \begin{subfigure}[t]{0.092\textwidth}
        \includegraphics[width=\textwidth]{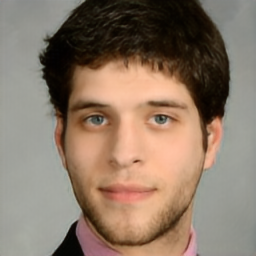}
        \vspace{-3.8mm}

        \includegraphics[width=\textwidth]{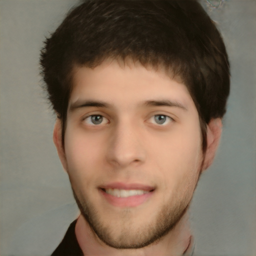}
        \vspace{-3.8mm}

        \includegraphics[width=\textwidth]{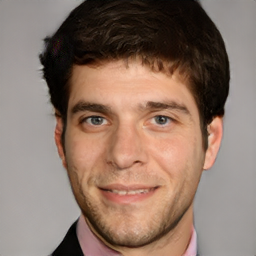}
        \vspace{-3.8mm}

        \includegraphics[width=\textwidth]{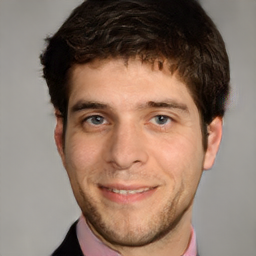}
        \caption{$0.4$}
    \end{subfigure}
    \begin{subfigure}[t]{0.092\textwidth}
        \includegraphics[width=\textwidth]{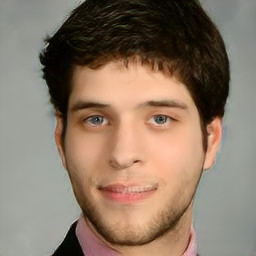}
        \vspace{-3.8mm}

        \includegraphics[width=\textwidth]{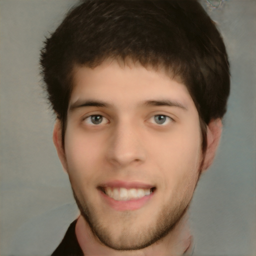}
        \vspace{-3.8mm}

        \includegraphics[width=\textwidth]{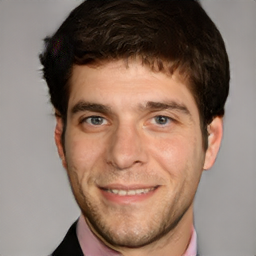}
        \vspace{-3.8mm}

        \includegraphics[width=\textwidth]{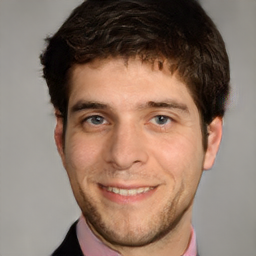}
        \caption{$0.6$}
    \end{subfigure}
    \begin{subfigure}[t]{0.092\textwidth}
        \includegraphics[width=\textwidth]{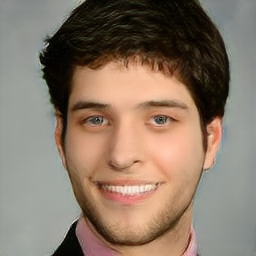}
        \vspace{-3.8mm}

        \includegraphics[width=\textwidth]{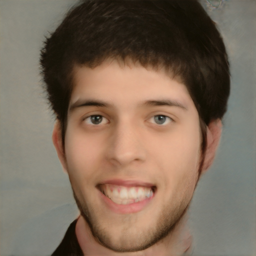}
        \vspace{-3.8mm}

        \includegraphics[width=\textwidth]{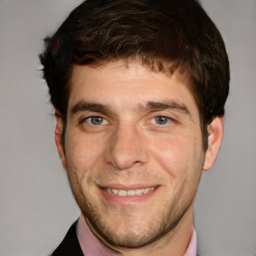}
        \vspace{-3.8mm}

        \includegraphics[width=\textwidth]{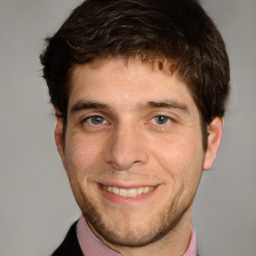}
        \caption{$0.8$}
    \end{subfigure}
    \begin{subfigure}[t]{0.092\textwidth}
        \includegraphics[width=\textwidth]{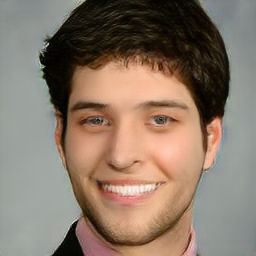}
        \vspace{-3.8mm}

        \includegraphics[width=\textwidth]{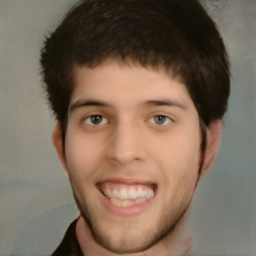}
        \vspace{-3.8mm}

        \includegraphics[width=\textwidth]{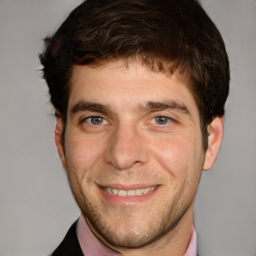}
        \vspace{-3.8mm}

        \includegraphics[width=\textwidth]{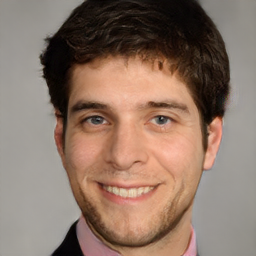}
        \caption{$1.0$}
    \end{subfigure}
    \begin{subfigure}[t]{0.092\textwidth}
        \includegraphics[width=\textwidth]{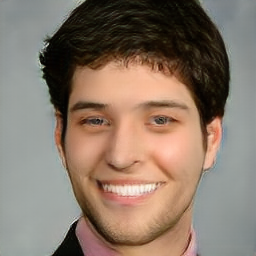}
        \vspace{-3.8mm}

        \includegraphics[width=\textwidth]{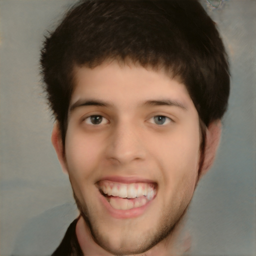}
        \vspace{-3.8mm}

        \includegraphics[width=\textwidth]{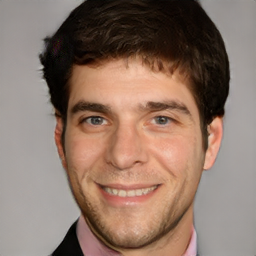}
        \vspace{-3.8mm}

        \includegraphics[width=\textwidth]{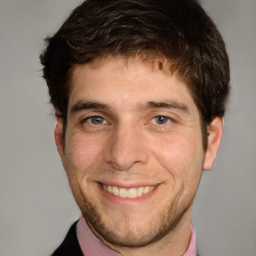}
        \caption{$1.2$}
    \end{subfigure}
    \begin{subfigure}[t]{0.092\textwidth}
        \includegraphics[width=\textwidth]{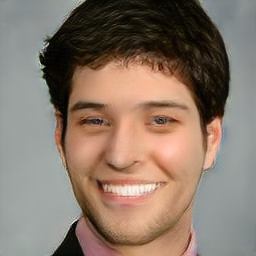}
        \vspace{-3.8mm}

        \includegraphics[width=\textwidth]{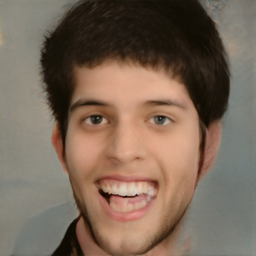}
        \vspace{-3.8mm}

        \includegraphics[width=\textwidth]{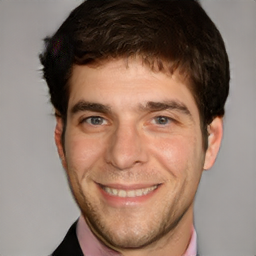}
        \vspace{-3.8mm}

        \includegraphics[width=\textwidth]{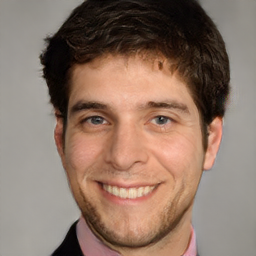}
        \caption{$1.4$}
    \end{subfigure}
    \begin{subfigure}[t]{0.092\textwidth}
        \includegraphics[width=\textwidth]{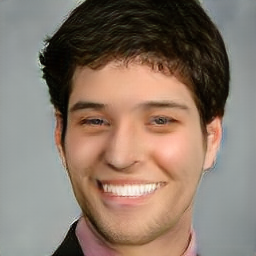}
        \vspace{-3.8mm}

        \includegraphics[width=\textwidth]{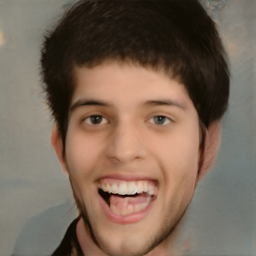}
        \vspace{-3.8mm}

        \includegraphics[width=\textwidth]{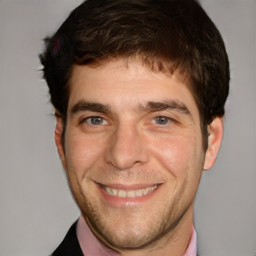}
        \vspace{-3.8mm}

        \includegraphics[width=\textwidth]{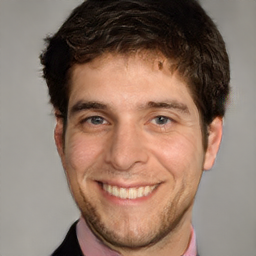}
        \caption{$1.6$}
    \end{subfigure}
    \begin{subfigure}[t]{0.092\textwidth}
        \includegraphics[width=\textwidth]{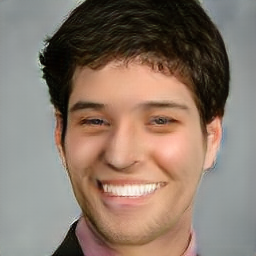}
        \vspace{-3.8mm}

        \includegraphics[width=\textwidth]{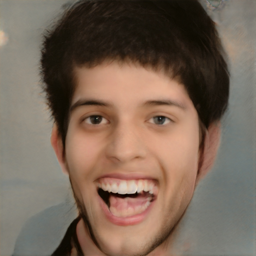}
        \vspace{-3.8mm}

        \includegraphics[width=\textwidth]{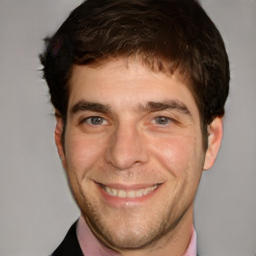}
        \vspace{-3.8mm}

        \includegraphics[width=\textwidth]{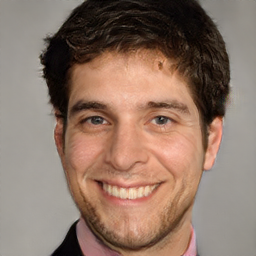}
        \caption{$1.8$}
    \end{subfigure}
    \begin{subfigure}[t]{0.092\textwidth}
        \includegraphics[width=\textwidth]{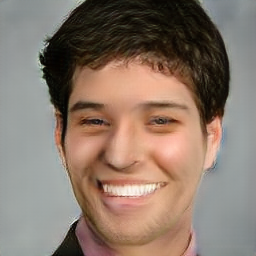}
        \vspace{-3.8mm}

        \includegraphics[width=\textwidth]{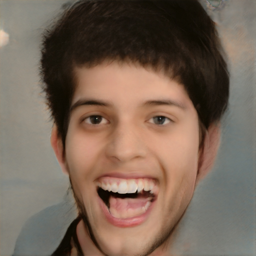}
        \vspace{-3.8mm}

        \includegraphics[width=\textwidth]{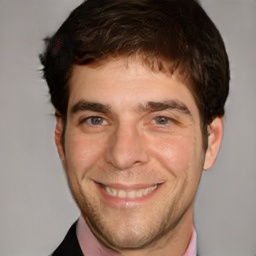}
        \vspace{-3.8mm}

        \includegraphics[width=\textwidth]{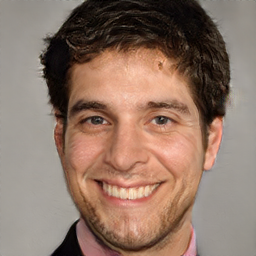}
        \caption{$2.0$}
    \end{subfigure}\hspace{0.4mm}
    \vspace{-3mm}
    \caption{Interpolation results on attribute ``smile'' obtained by RelGAN~\cite{Wu2019RelGANMI}, InterFaceGAN(IFGAN)~\cite{Shen2020InterFaceGANIT}, HifaFace without the $\mathcal{L}_{ar}$ and our HifaFace.}
    \label{fig:sm_arbitrary_comparison1}
    \vspace{-3mm}
\end{figure*}

\begin{figure*}[h]
    \captionsetup[subfigure]{labelformat=empty}
    \captionsetup[subfigure]{aboveskip=1pt} %
    \centering
    \begin{subfigure}[t]{\dimexpr0.092\textwidth+11pt\relax}
        \makebox[11pt]{\raisebox{18pt}{\rotatebox[origin=c]{90}{RelGAN}}}%
        \includegraphics[width=\dimexpr\linewidth-11pt\relax]{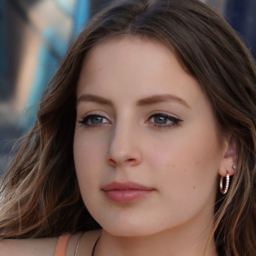}
        \vspace{-3.8mm}

        \makebox[11pt]{\raisebox{18pt}{\rotatebox[origin=c]{90}{IFGAN}}}%
        \includegraphics[width=\dimexpr\linewidth-11pt\relax]{images_sm/interp_wo_2054_in}
        \vspace{-3.8mm}

        \makebox[11pt]{\raisebox{18pt}{\rotatebox[origin=c]{90}{w/o $\mathcal{L}_{ar}$}}}%
        \includegraphics[width=\dimexpr\linewidth-11pt\relax]{images_sm/interp_wo_2054_in}
        \vspace{-3.8mm}

        \makebox[11pt]{\raisebox{18pt}{\rotatebox[origin=c]{90}{\textbf{HifaFace}}}}%
        \includegraphics[width=\dimexpr\linewidth-11pt\relax]{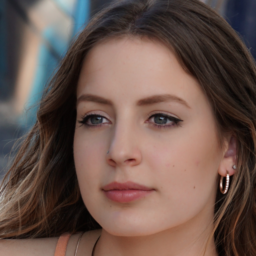}
        \caption{\quad Input}
    \end{subfigure}\hspace{0.4mm}
    \begin{subfigure}[t]{0.092\textwidth}
        \includegraphics[width=\textwidth]{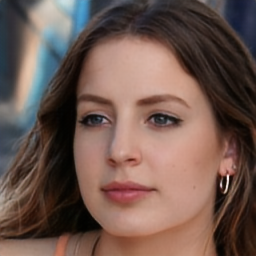}
        \vspace{-3.8mm}

        \includegraphics[width=\textwidth]{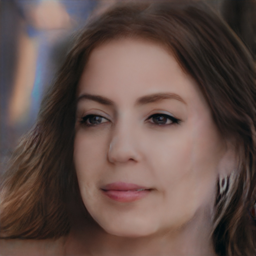}
        \vspace{-3.8mm}

        \includegraphics[width=\textwidth]{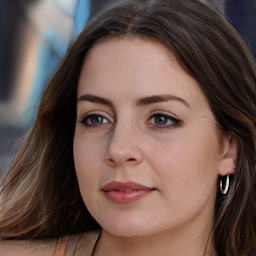}
        \vspace{-3.8mm}

        \includegraphics[width=\textwidth]{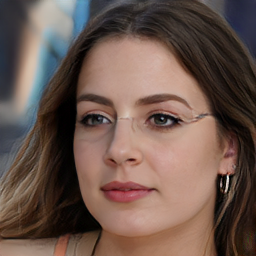}
        \caption{$0.4$}
    \end{subfigure}
    \begin{subfigure}[t]{0.092\textwidth}
        \includegraphics[width=\textwidth]{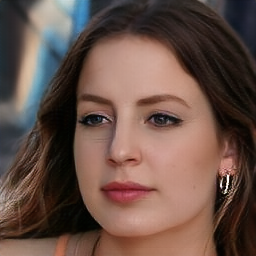}
        \vspace{-3.8mm}

        \includegraphics[width=\textwidth]{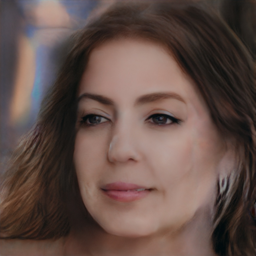}
        \vspace{-3.8mm}

        \includegraphics[width=\textwidth]{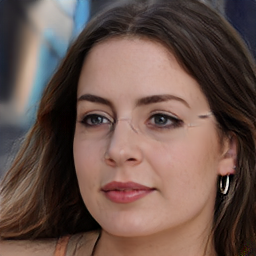}
        \vspace{-3.8mm}

        \includegraphics[width=\textwidth]{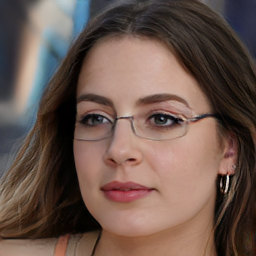}
        \caption{$0.6$}
    \end{subfigure}
    \begin{subfigure}[t]{0.092\textwidth}
        \includegraphics[width=\textwidth]{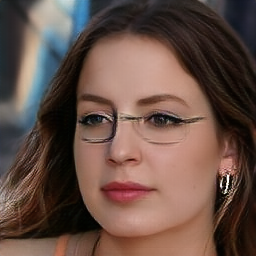}
        \vspace{-3.8mm}

        \includegraphics[width=\textwidth]{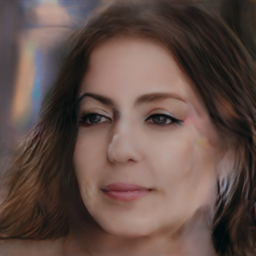}
        \vspace{-3.8mm}

        \includegraphics[width=\textwidth]{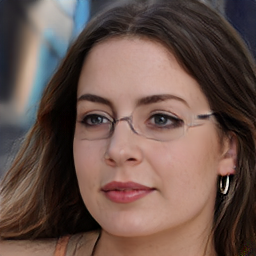}
        \vspace{-3.8mm}

        \includegraphics[width=\textwidth]{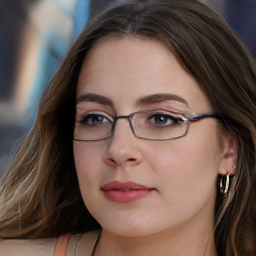}
        \caption{$0.8$}
    \end{subfigure}
    \begin{subfigure}[t]{0.092\textwidth}
        \includegraphics[width=\textwidth]{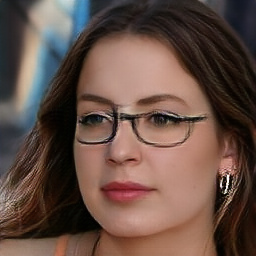}
        \vspace{-3.8mm}

        \includegraphics[width=\textwidth]{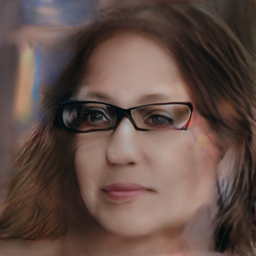}
        \vspace{-3.8mm}

        \includegraphics[width=\textwidth]{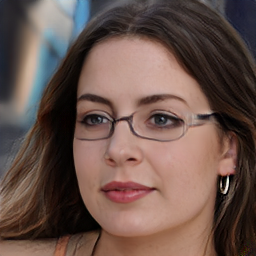}
        \vspace{-3.8mm}

        \includegraphics[width=\textwidth]{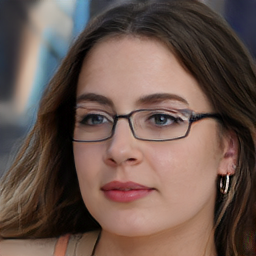}
        \caption{$1.0$}
    \end{subfigure}
    \begin{subfigure}[t]{0.092\textwidth}
        \includegraphics[width=\textwidth]{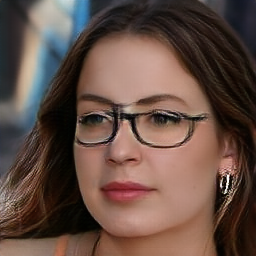}
        \vspace{-3.8mm}

        \includegraphics[width=\textwidth]{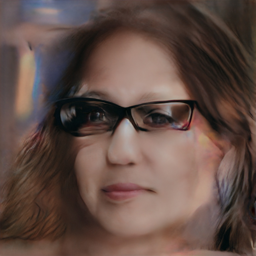}
        \vspace{-3.8mm}

        \includegraphics[width=\textwidth]{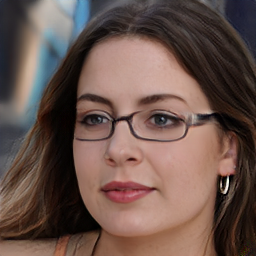}
        \vspace{-3.8mm}

        \includegraphics[width=\textwidth]{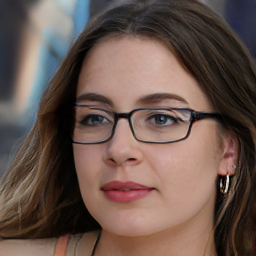}
        \caption{$1.2$}
    \end{subfigure}
    \begin{subfigure}[t]{0.092\textwidth}
        \includegraphics[width=\textwidth]{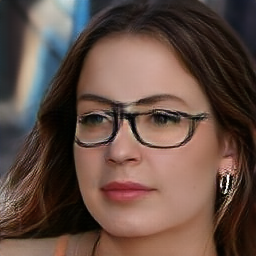}
        \vspace{-3.8mm}

        \includegraphics[width=\textwidth]{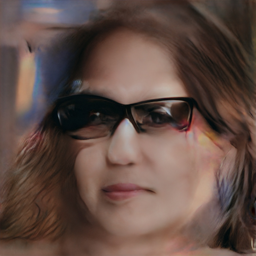}
        \vspace{-3.8mm}

        \includegraphics[width=\textwidth]{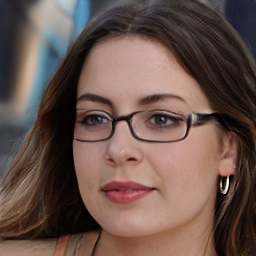}
        \vspace{-3.8mm}

        \includegraphics[width=\textwidth]{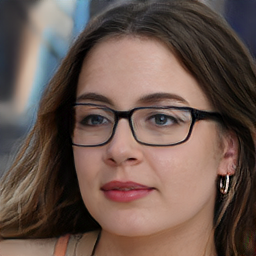}
        \caption{$1.4$}
    \end{subfigure}
    \begin{subfigure}[t]{0.092\textwidth}
        \includegraphics[width=\textwidth]{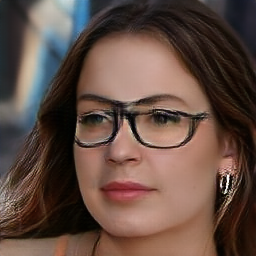}
        \vspace{-3.8mm}

        \includegraphics[width=\textwidth]{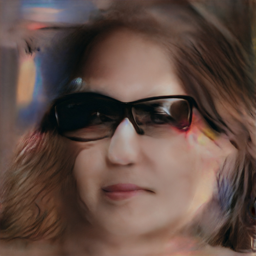}
        \vspace{-3.8mm}

        \includegraphics[width=\textwidth]{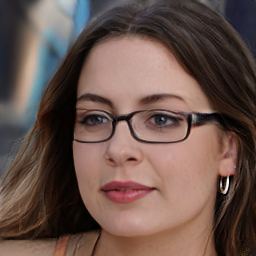}
        \vspace{-3.8mm}

        \includegraphics[width=\textwidth]{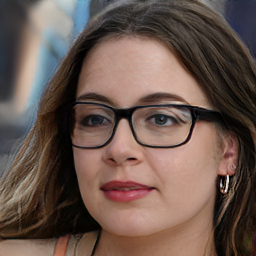}
        \caption{$1.6$}
    \end{subfigure}
    \begin{subfigure}[t]{0.092\textwidth}
        \includegraphics[width=\textwidth]{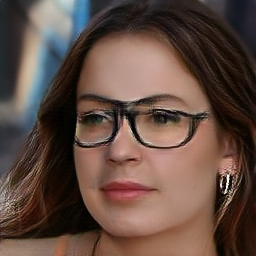}
        \vspace{-3.8mm}

        \includegraphics[width=\textwidth]{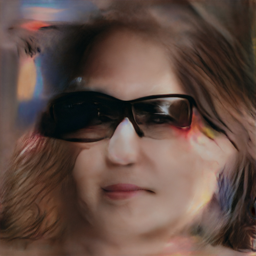}
        \vspace{-3.8mm}

        \includegraphics[width=\textwidth]{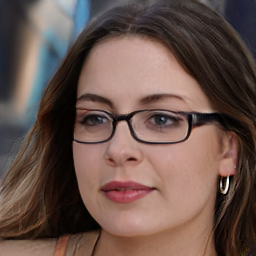}
        \vspace{-3.8mm}

        \includegraphics[width=\textwidth]{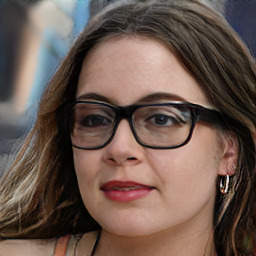}
        \caption{$1.8$}
    \end{subfigure}
    \begin{subfigure}[t]{0.092\textwidth}
        \includegraphics[width=\textwidth]{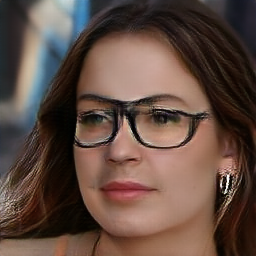}
        \vspace{-3.8mm}

        \includegraphics[width=\textwidth]{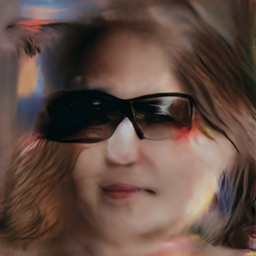}
        \vspace{-3.8mm}

        \includegraphics[width=\textwidth]{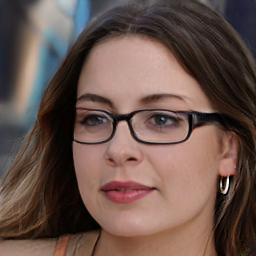}
        \vspace{-3.8mm}

        \includegraphics[width=\textwidth]{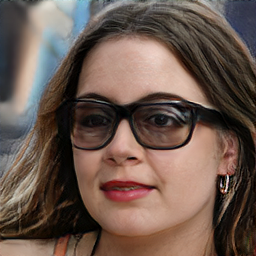}
        \caption{$2.0$}
    \end{subfigure}\hspace{0.4mm}
    \vspace{-3mm}
    \caption{Interpolation results on attribute ``eyeglasses'' obtained by RelGAN~\cite{Wu2019RelGANMI}, InterFaceGAN(IFGAN)~\cite{Shen2020InterFaceGANIT}, HifaFace without the $\mathcal{L}_{ar}$ and our HifaFace.}
    \label{fig:sm_arbitrary_comparison2}
    \vspace{-3mm}
\end{figure*}

\begin{figure*}[h]
    \captionsetup[subfigure]{labelformat=empty}
    \captionsetup[subfigure]{aboveskip=1pt} %
    \centering
    \begin{subfigure}[t]{0.11\textwidth}
        \includegraphics[width=\textwidth]{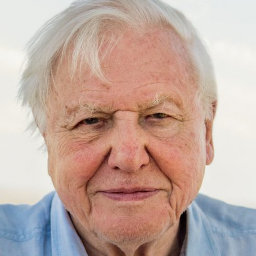}
        \vspace{-3.6mm}

        \includegraphics[width=\textwidth]{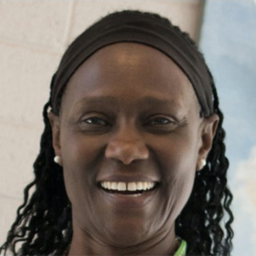}
        \caption{Input}
        \vspace{1mm}
        \includegraphics[width=\textwidth]{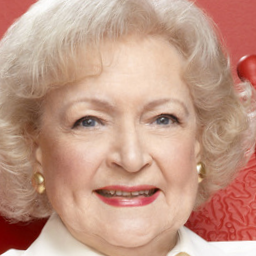}
        \vspace{-3.6mm}

        \includegraphics[width=\textwidth]{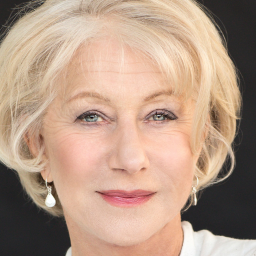}
        \caption{Input}
    \end{subfigure}\hspace{0.4mm}
    \begin{subfigure}[t]{0.11\textwidth}
        \includegraphics[width=\textwidth]{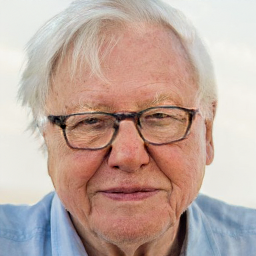}
        \vspace{-3.6mm}

        \includegraphics[width=\textwidth]{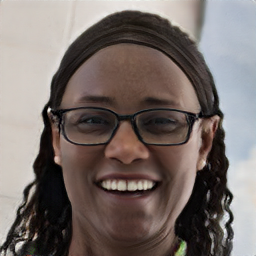}
        \caption{Eyeglasses}
        \vspace{1mm}
        \includegraphics[width=\textwidth]{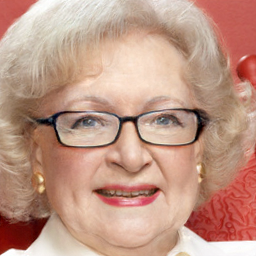}
        \vspace{-3.6mm}

        \includegraphics[width=\textwidth]{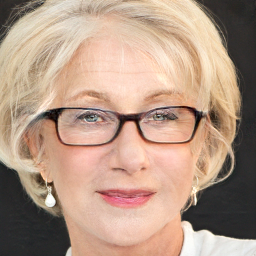}
        \caption{Eyeglasses}
    \end{subfigure}\hspace{0.4mm}
    \begin{subfigure}[t]{0.11\textwidth}
        \includegraphics[width=\textwidth]{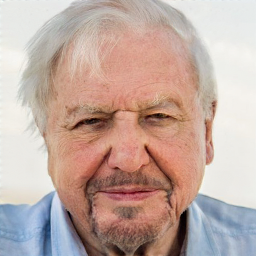}
        \vspace{-3.6mm}

        \includegraphics[width=\textwidth]{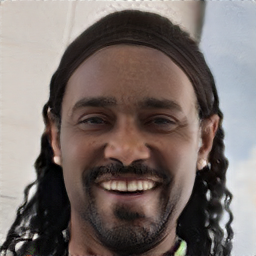}
        \caption{Mustache}
        \vspace{1mm}
        \includegraphics[width=\textwidth]{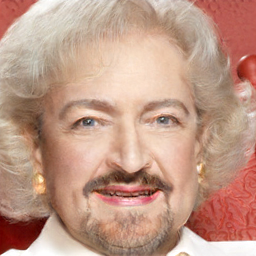}
        \vspace{-3.6mm}

        \includegraphics[width=\textwidth]{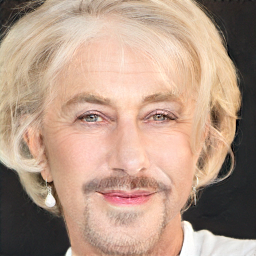}
        \caption{Mustache}
    \end{subfigure}\hspace{0.4mm}
    \begin{subfigure}[t]{0.11\textwidth}
        \includegraphics[width=\textwidth]{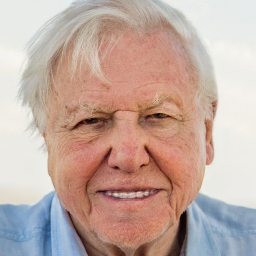}
        \vspace{-3.6mm}

        \includegraphics[width=\textwidth]{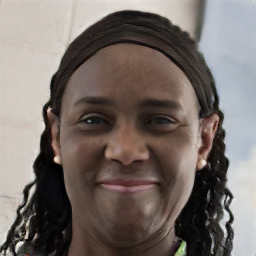}
        \caption{Expression}
        \vspace{1mm}
        \includegraphics[width=\textwidth]{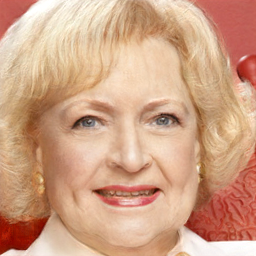}
        \vspace{-3.6mm}

        \includegraphics[width=\textwidth]{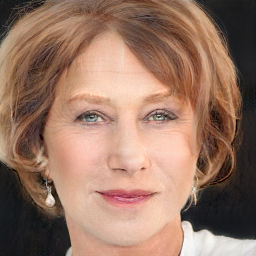}
        \caption{Hair color}
    \end{subfigure}\hspace{0.4mm}
    \begin{subfigure}[t]{0.11\textwidth}
        \includegraphics[width=\textwidth]{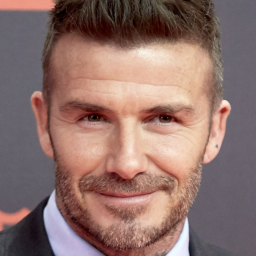}
        \vspace{-3.6mm}

        \includegraphics[width=\textwidth]{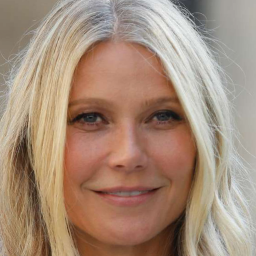}
        \caption{Input}
        \vspace{1mm}
        \includegraphics[width=\textwidth]{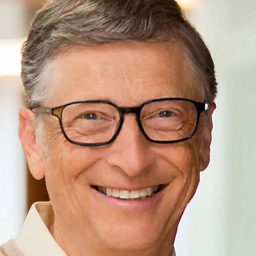}
        \vspace{-3.6mm}

        \includegraphics[width=\textwidth]{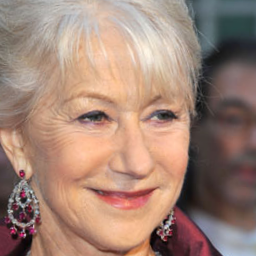}
        \caption{Input}
    \end{subfigure}\hspace{0.4mm}
    \begin{subfigure}[t]{0.11\textwidth}
        \includegraphics[width=\textwidth]{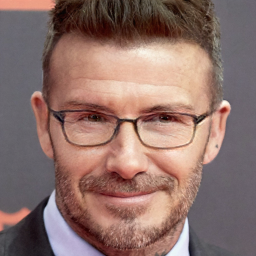}
        \vspace{-3.6mm}

        \includegraphics[width=\textwidth]{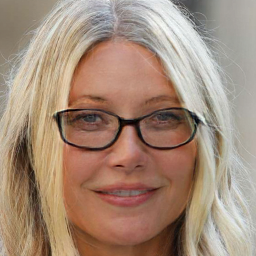}
        \caption{Eyeglasses}
        \vspace{1mm}
        \includegraphics[width=\textwidth]{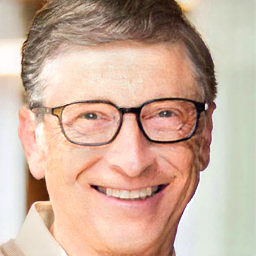}
        \vspace{-3.6mm}

        \includegraphics[width=\textwidth]{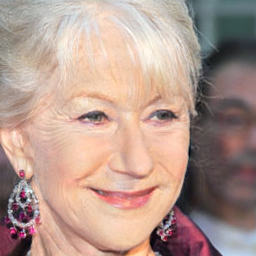}
        \caption{Illumination}
    \end{subfigure}\hspace{0.4mm}
    \begin{subfigure}[t]{0.11\textwidth}
        \includegraphics[width=\textwidth]{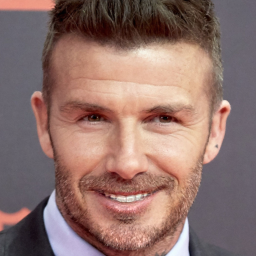}
        \vspace{-3.6mm}

        \includegraphics[width=\textwidth]{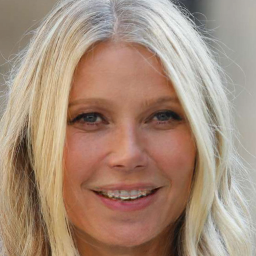}
        \caption{Expression}
        \vspace{1mm}
        \includegraphics[width=\textwidth]{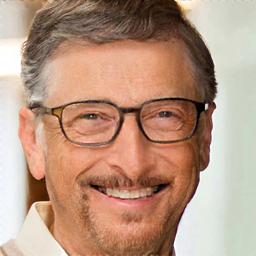}
        \vspace{-3.6mm}

        \includegraphics[width=\textwidth]{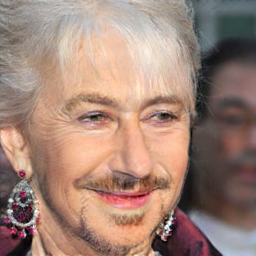}
        \caption{Mustache}
    \end{subfigure}\hspace{0.4mm}
    \begin{subfigure}[t]{0.11\textwidth}
        \includegraphics[width=\textwidth]{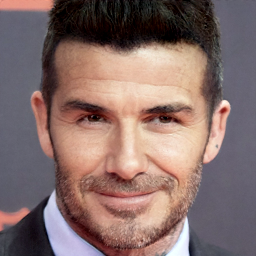}
        \vspace{-3.6mm}

        \includegraphics[width=\textwidth]{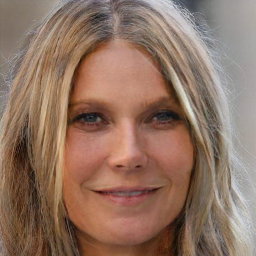}
        \caption{Hair color}
        \vspace{1mm}
        \includegraphics[width=\textwidth]{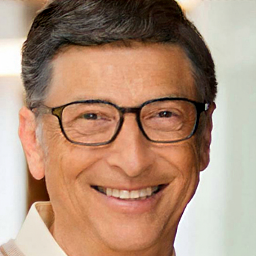}
        \vspace{-3.6mm}

        \includegraphics[width=\textwidth]{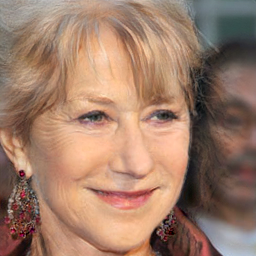}
        \caption{Hair color}
    \end{subfigure}
    \vspace{-3mm}
    \caption{Face editing results obtained by our \textbf{HifaFace} on wild images.}
    \label{fig:sm_comparison_wild}
    \vspace{-5mm}
\end{figure*}

\section{Additional Visual Results}
In this section, more visual results are provided to demonstrate the superiority of our model.
The figures to be presented and their corresponding subjects are listed as follows:
\begin{itemize}
    \item In Figure~\ref{fig:sm_comparison_sota}, we present the comparison of attribute-based face editing results obtained by our model and other existing methods, including GANimation~\cite{pumarola2018ganimation}, STGAN~\cite{Liu2019STGANAU}, RelGAN~\cite{Wu2019RelGANMI}, InterFaceGAN~\cite{Shen2020InterFaceGANIT} and StyleFlow~\cite{Abdal2020StyleFlowAE}. We also provide the results by an industrial app, FaceApp~\cite{faceapp}.
    \item In Figure~\ref{fig:sm_arbitrary_comparison1} and Figure~\ref{fig:sm_arbitrary_comparison2}, we show the comparison of arbitrary face editing results obtained by our HifaFace, our model without the attribute regression loss $\mathcal{L}_{ar}$, RelGAN~\cite{Wu2019RelGANMI} and InterFaceGAN~\cite{Shen2020InterFaceGANIT}.
    \item In Figure~\ref{fig:sm_comparison_wild}, we demonstrate that our method HifaFace can handle face images under various poses, races and expressions.
\end{itemize}

\end{document}